\documentclass[12pt]{article}
\usepackage[a4paper,left=3mm,right=3mm,top=15mm,bottom=15mm]{geometry}
\usepackage{caption,subcaption}
\usepackage{amsfonts}
\usepackage{amsmath}
\usepackage{amsthm}
\usepackage{amssymb,bm}
\usepackage[mathscr]{eucal}
\usepackage{graphicx}
\usepackage{color}
\usepackage{cite}
\usepackage{comment}
\usepackage{geometry}
\usepackage{multirow}
\usepackage{subcaption}
\usepackage{float}
\usepackage{multicol}
\usepackage{enumerate}
\usepackage{booktabs}
\usepackage{threeparttable}
\usepackage{algorithm}
\usepackage{algorithmic}
\newtheorem{definition}{Definition}
\newtheorem{theorem}{Theorem}

\usepackage{authblk} 

\title{\bf Anisotropic View Distance Metric for High-Dimensional Data: Theory, Geometry, and Fast Computation}
\author[1]{Yiqun Zhang}
\author[1]{Hou-biao Li\thanks{Corresponding author: Hou-biao Li, E-mail: \texttt{lhb0189@uestc.edu.cn}}}
\affil[1]{School of Mathematical Sciences, University of Electronic Science and Technology of China, Chengdu 611731, China}

\date{}

\begin{document}
	\captionsetup{font={normalsize}}
	\maketitle
	
	\abstract{\indent Euclidean distance is an efficient and interpretable similarity measurement, but its effectiveness may deteriorate in sample spaces with anisotropic structures, redundant features, or complex feature interactions. In this paper, we propose a novel distance metric called View distance. Inspired by orthographic projection, the proposed metric projects the sample space onto $n(n-1)/2$ two-dimensional planes and defines the final distance as the sum of the Euclidean distances across the projected planes. Theoretical derivations verify that the View distance strictly satisfies the metric axioms and norm constraints. Beyond that, the View distance achieves feature coupling through projection and not only enables constant features to indirectly participate in distance calculation, but also suppresses interference from redundant features while exhibiting anisotropic geometric properties. Furthermore, to address the high computational complexity and poor scalability of full-projection View distance, we propose a two-dimensional projection plane selection strategy based on iterative Maximum Weight Matching, which reduces the computational complexity of distance calculation from $\mathcal{O}(n^2)$ to $\mathcal{O}(k)$. Extensive experiments on 12 diverse datasets demonstrate that View distance and the deterministic selection strategy provide competitive or superior performance compared with Euclidean distance and other $L_{p}$ metrics while maintaining strong interpretability and computational efficiency. The View distance provides a new perspective and option for similarity measurements.}


	\section{Introduction}\label{sec1}	
	Metric learning is a fundamental research branch of machine learning that aims to learn effective distance metrics or feature embedding spaces to accurately distinguish similar and dissimilar samples. Its core principle is to minimize intra-class distances and maximize inter-class distances to improve feature discrimination. Existing metric learning methods can be divided into two categories: linear and nonlinear.
	
	Traditional linear metric learning is built on classic distance metrics, including Euclidean distance, Manhattan distance, cosine similarity, Mahalanobis distance, and others. It optimizes feature representation by learning a linear transformation matrix to map the original features into a new space that satisfies task-specific distance constraints\cite{JGoldberger, Bar-hillel, HoiSCH, Weinberger}. Nevertheless, linear methods can only model linearly separable relationships and exhibit limited performance on complex nonlinear data distributions. To address this limitation, kernel-based methods\cite{CorinnaCortes, CJCBurges} and Deep Metric Learning (DML)\cite{DuanYueqi, Kaya} have been developed for nonlinear feature modeling. Kernel-based nonlinear metric learning projects the original data into high-dimensional feature spaces via kernel functions (e.g., polynomial, Gaussian, and Laplacian kernels) to capture complex data structures. Deep metric learning employs deep neural networks, such as Siamese\cite{SumitChopra} and Triplet\cite{EladHoffer} networks, to learn discriminative embeddings end-to-end. It further integrates probability-based distance metrics, including Bregman divergence\cite{JVDavis}, KL divergence, JS divergence, and Wasserstein distance\cite{MartinArjovsky}, to enhance similarity measurement. Nonlinear metric learning methods have achieved remarkable performance across a range of challenging tasks, including image recognition, recommendation systems, and natural language processing.
    
    However, nonlinear metric learning still suffers from inherent limitations. First, it relies on large-scale labeled datasets and is prone to overfitting on small-scale or noisy datasets, leading to performance degradation. Second, whether based on kernel methods or deep learning frameworks, these approaches are essentially black-box models that lack interpretability for similarity judgments\cite{YannLeCun}. These shortcomings limit their use in fields such as biology and medicine, where sample sizes are limited, and feature differences are subtle. Consequently, linear metric methods remain available in these scenarios.
			
	As the most classic metric, Euclidean distance is widely adopted for its simplicity and low computational cost. Nevertheless, Euclidean distance has the following limitations in the sample space. First, it only characterizes straight-line geometric distance and cannot capture the data's manifold structure, resulting in insufficient spatial modeling capability\cite{JoshuaBTenenbaum, SamTRoweis, vandermaaten}. More critically, it merely aggregates univariate discrepancies without introducing cross-dimensional coupling terms, which is equivalent to the Mahalanobis distance under the assumption that the feature covariance matrix is the identity matrix (i.e., each dimension is independent and identically distributed). Therefore, it fails to characterize the correlations among different features. Furthermore, it suffers from distance concentration\cite{KevinBeyer, MiloRadovanovi, MichaelEHoule, ArthurZimek} in high-dimensional spaces, where the difference between the maximum and minimum distances diminishes sharply, degrading sample distinguishability. 
    
    \begin{figure}[H]
    	\centering
    	\includegraphics[width=0.32\textwidth]{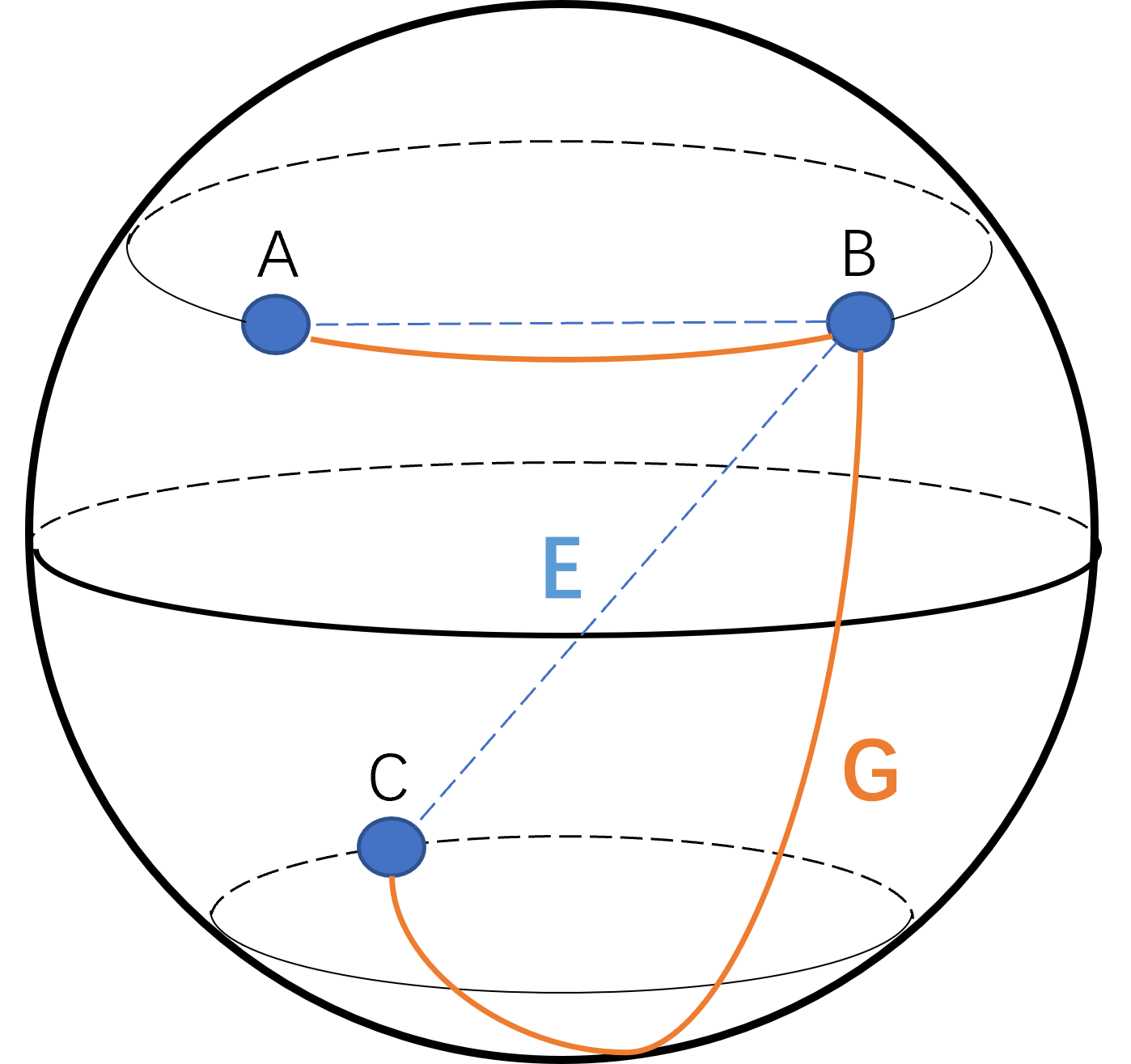}
    	\captionsetup{width=0.90\textwidth, font=footnotesize}
    	\caption{The sketch map shows geodesic and Euclidean distances on a three-dimensional sphere. The yellow and blue lines represent the geodesic distance and the Euclidean distance between two points.}
    \end{figure}

    Moreover, although derivative algorithms based on projection or view have been widely studied and applied\cite{ZhiwenYu}, research specifically focusing on projection‑based or view-based distance metrics remains noticeably scarce. This research gap reveals a clear need for fundamental mathematical tools to guide the research and application of such algorithms.
    
    In this paper, we propose a novel View distance metric inspired by orthographic projection. The proposed method projects two points in $n$-dimensional space onto $n(n-1)/2$ non-overlapping two-dimensional planes. It defines the final distance as the sum of Euclidean distances of all projected point pairs. The main contributions are summarized as follows.
    \begin{enumerate}[(1)]
    	\item Comprehensive theoretical property analysis. Theoretical proofs demonstrate that the View distance satisfies non-negativity, positivity, and the triangle inequality, which conforms to the strict definition of a valid distance metric. Its corresponding norm satisfies non-negativity, homogeneity, and the triangle inequality and is a legitimate norm defined on $\mathbb{R}^n$. Constant features can make an effective contribution to the calculation of the View distance. Moreover, the View distance exhibits anisotropic spatial characteristics.	
        \item An efficient iterative optimization strategy. To improve the scalability and computational efficiency of the View distance, this paper further proposes a two-dimensional projection plane selection strategy based on iterative maximum weight matching. By quantifying the importance score of each two-dimensional projection plane, this strategy selects $k$ planes that balance maximum discriminability and feature diversity, thereby reducing the computational complexity of distance calculation from $\mathcal{O}(n^2)$ to $\mathcal{O}(k)$. 
        \item Sufficient experimental validation. We integrate View distance into both the DBSCAN\cite{MartinEster} clustering algorithm and the $k$-Nearest Neighbors (KNN)\cite{TCover} classification algorithm to evaluate their performance on clustering and classification tasks. Furthermore, we adopt the conventional KNN algorithm as the baseline to verify the effectiveness of the proposed two-dimensional projection plane selection strategy. Extensive experiments across 2 synthetic datasets and 12 small-scale real-world datasets with diverse feature characteristics demonstrate that the View distance consistently achieves optimal or suboptimal overall performance. Specifically, the proposed two-dimensional selection strategy retains more than $98.9\%$ of the full-projection scheme's performance by selecting only $n/2$ projection planes. It even outperforms the full-projection model in certain cases. This provides a reliable and effective approach to applying and extending the View distance in high-dimensional scenarios. 
    \end{enumerate}
	
	\section{Related work}\label{sec2}	
	\subsection{Multi-view metric learning}\label{sec2_1}
	Multi-view learning is an emerging field that aims to enhance learning performance by leveraging multiple views or sources of data across various domains. By integrating information from diverse perspectives, multi-view learning methods effectively enhance accuracy, robustness, and generalization capabilities\cite{ZhiwenYu}. Wang et al.\cite{HuibingWang} proposed a KL divergence-based MML method that constructs consensus distance metrics and employs large-margin criteria to improve class separability. Hu et al.\cite{JunlinHu} presented the MvML method to learn both view-specific independent metrics and shared latent representations, and further extended it to the deep MvDML model to capture nonlinear data structures. Yu et al.\cite{YifanYu} developed the MVML-ISFS model, which learns large-margin independent metrics and establishes cross-view correlations via sparse representation to achieve superior performance across multiple classification tasks. Wang et al.\cite{ZhiWang} proposed a multi-view classification method integrating reweighted regression and $L_{1}$ sparse regularization, which adaptively weights different views to enhance classification performance. Tang et al.\cite{JingjingTang} proposed a novel MVMIML framework for image classification, which extracts multi-view multi-instance features and designs a new bag distance metric to learn view-dependent weights and improve label consistency. To address heterogeneous clustering problems, Zhang et al.\cite{YiqunZhang} presented the HARR paradigm. It unifies heterogeneous attribute spaces through projection reconstruction and learns adaptive metrics to optimize clustering performance.
	
	\subsection{Feature matching strategies}\label{sec2_2}
    Numerous studies have explored advanced feature matching algorithms. Park et al.\cite{JuhyeonPark} verified that mapping traditional Euclidean distances into structured metric spaces, such as Mahalanobis distance and large-margin metric spaces, can effectively improve feature discriminability, providing a fundamental basis for feature matching modeling in this work. Ji et al.\cite{GangJi} investigated nonlinear feature interaction scoring and subset filtering strategies to eliminate feature redundancy. Lu et al.\cite{YifanLu} proposed a feature matching method named GC-LAC. This method reformulates the feature matching task as a graph clustering problem and incorporates a geometric solver (MCDG) to extract local geometric information that is beneficial for graph construction. Accordingly, robust feature correspondences are built, and latent visual patterns are identified. Xia et al.\cite{YifanXia} proposed a novel outlier rejection framework termed LOGO. By integrating locality-guided fixed-point progressive optimization and a global affinity matrix, this framework achieves robust and efficient feature matching under complex non-rigid transformations and high outlier ratios. Hashemi et al.\cite{AminHashemi} proposed a bipartite graph matching-based feature selection method for multi-label learning (BMFS). By constructing a feature-label graph and applying the Hungarian algorithm to achieve optimal matching, this method selects discriminative features while jointly accounting for feature redundancy and the correlations between features and class labels.
      
    \subsection{Adaptive distance metrics for KNN classification}\label{sec2_3}
    The performance of the KNN classifier is fundamentally constrained by the chosen distance metric. Due to the aforementioned limitations of the Euclidean distance, adopting non-Euclidean, adaptive distance metrics has emerged as a core strategy to enhance classification accuracy and model robustness. Alfeilat et al.\cite{AbuAlfeilat} comprehensively evaluated the effects of 54 distance metrics on the performance of the KNN classifier. The experimental results reveal that the Hassanat distance (HasD) achieves the best performance in terms of accuracy, precision, and recall on both noisy and noise-free datasets, demonstrating its strong robustness against noise. Xiao et al.\cite{XinpingXiao} improved the Mahalanobis distance by introducing regularization and smoothing techniques. Combining the mRMR algorithm and the signal-to-noise ratio, they proposed a two-stage feature selection method to optimize the Mahalanobis-Taguchi system and enhance its robustness and classification performance for high-dimensional small-sample data.
    
    To improve the robustness of KNN against outliers, Gou et al.\cite{JianpingGou} introduced the Generalized Mean Distance KNN (GMDKNN), which applies a generalized mean operator to the distances of sample neighbors. Rodrigues et al.\cite{EORodrigues} proposed a novel metric combining Minkowski distance and Chebyshev distance via weights $\omega_{1}$ and $\omega_{2}$. This metric achieves efficient neighbourhood iteration speed and favorable accuracy when applied to KNN classification. Ghosh et al.\cite{AnneshaGhosh} proposed classifiers based on global and local Mahalanobis distances (MD and LMD). Using these distances as features in a generalized additive model to estimate posterior probabilities, the classifiers can effectively handle classification tasks involving both elliptical and non-elliptical distributions and achieve favorable performance in high-dimensional and small-sample scenarios. Palias et al.\cite{EfstratiosPalias} proposed a compressed Mahalanobis distance metric learning method, which adopts random projection to allow the error bound to adapt to the intrinsic dimensionality of data instead of the original high dimensionality. Zaidi et al.\cite{NayyarAbbasZaidi} proposed an adaptive metric learning algorithm named BoostML. Inspired by Boosting, the algorithm learns the optimal metric for query points using a feature-correlation measure, aiming to reduce bias in high-dimensional spaces and thereby improve the performance of the KNN classifier. Zhang et al.\cite{HuanxingZhang} proposed a joint-norm two-dimensional principal component analysis method (2DPCA), which combines the 2-norm and $L_{p}$-norm, simultaneously optimizes the maximum projection distance and minimum reconstruction error for robust dimensionality reduction, and tackles the limitations of existing single-norm methods regarding reconstruction error and rotation invariance. Wang et al.\cite{XiaofengWang} proposed a cosine two-dimensional principal component analysis method (Cosine 2DPCA). By introducing a weighted cosine objective function and the 2-norm distance metric, this method effectively improves the performance of 2DPCA for robust feature extraction, reconstruction, correlation analysis, and classification, while reducing computational complexity and reconstruction error. Song et al.\cite{KunSong} proposed the 2DLMNN method. This method defines the matrix‑based Mahalanobis distance via left and right projection matrices to reduce computational complexity and the risk of overfitting, and achieves efficient classification through alternating optimization.
    	
	\section{View distance}\label{sec3}
	\subsection{Definition of View distance}\label{sec3_1}
	Considering the mathematical concept of projection, two points in an $n$-dimensional(abbreviated as $nD$) space are projected onto the $(n-1)D$ hyperplane, yielding $n$ such $(n-1)D$ hyperplanes. These $n$ hyperplanes are then projected onto $(n-2)D$ hyperplane, resulting in $n\times(n-1)$ such $(n-2)D$ hyperplanes. By iteratively repeating this projection process until $2D$ hyperplanes are reached, a total of $n!/2$ $2D$ hyperplanes are obtained, among which $n(n-1)/2$ are non-overlapping.
	
\begin{figure}[H]
		\centering
		\includegraphics[width=0.44\textwidth]{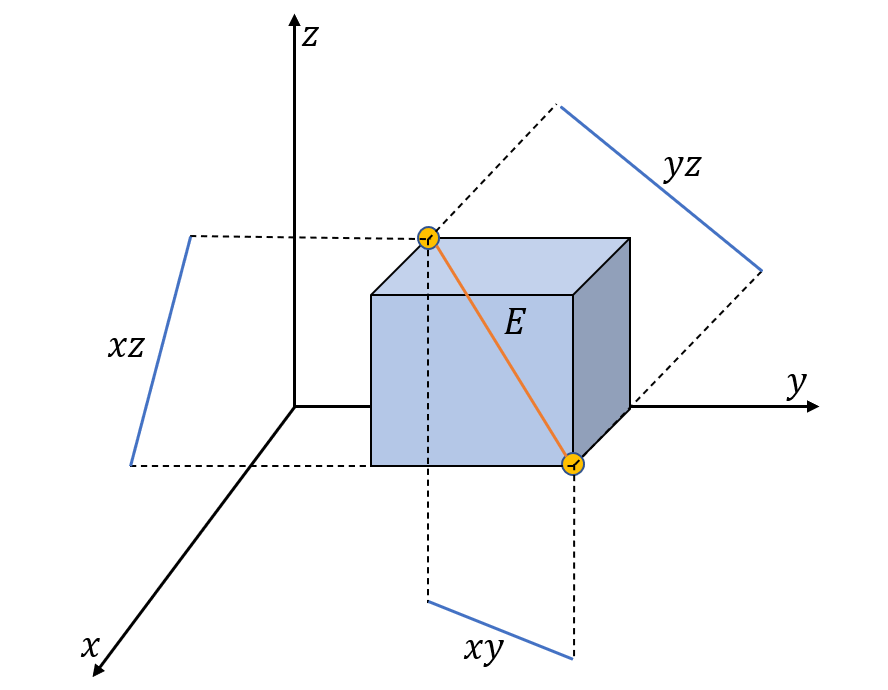}
		\captionsetup{width=0.90\textwidth, font=footnotesize}
		\caption{The sketch map of the View and Euclidean distance in $3D$ space. E represents the Euclidean distance between two points. $xy$, $xz$, $yz$ represent the Euclidean distance between two points projected on three $2D$ planes.}
		\label{figure2}
	\end{figure}

	The distance between two points in the $nD$ space is represented as the sum of the Euclidean distances of their projections onto these $n!/2$ $2D$ hyperplanes. Analogous to the concept of orthographic projection, this distance is termed View distance. Its schematic diagram is shown in Figure~\ref{figure2}, and the formal definition is provided below.
	
	For any $\boldsymbol{x}, \boldsymbol{y}, \boldsymbol{z} \in {\mathbb{R}^n}$, let $\boldsymbol{x} = (x_1, x_2, ..., x_n)$, $\boldsymbol{y} =$ $(y_1, y_2, \ldots, y_n)$, $\boldsymbol{z} = (z_1, z_2, \ldots, z_n)$, the Euclidean distance\cite{JC, Gower, IvanDokmani} between $\boldsymbol{x}$ and $\boldsymbol{y}$ is
	\begin{equation*}
		d_E(\boldsymbol{x}, \boldsymbol{y}) = \sqrt{\sum_{i=1}^{n} (x_i - y_i)^{2}}.
	\end{equation*}
	\begin{definition}
		The distance between $\boldsymbol{x}$ and $\boldsymbol{y}$ is defined by
		\begin{equation*}
			d_V(\boldsymbol{x}, \boldsymbol{y}) = (n-2)! \sum_{1\leq{i}<j\leq{n}}^{n} \sqrt{(x_i - y_i)^{2} + (x_j - y_j)^2},
		\end{equation*}
		where $n\ge{2}$, abbreviated as View distance.
	\end{definition}
	\indent Since the Euclidean distance satisfies the axioms of non-negativity, identity, symmetry, and the triangle inequality, it follows from the formula that the View distance also satisfies three axioms.
	\begin{proof}
        Let $\boldsymbol{x}_{ij}=(x_i, x_j)$, $\boldsymbol{y}_{ij}=(y_i, y_j)$, $\boldsymbol{z}_{ij}=(z_i, z_j)$,
        \begin{gather*}
	        d_E(\boldsymbol{x}_{ij}, \boldsymbol{y}_{ij}) = \sqrt{(x_i - y_i)^{2} + (x_j - y_j)^2},
        \end{gather*}
        represents the Euclidean distance between $(x_i,x_j)$ and $(y_i,y_j)$.
		
		(1) Non-negativity and identity.
		
		As the Euclidean distance obeys the Non-negativity and identity, that is, for any indices $i,j$ with $1 \leq i < j \leq n$, $d_E(\boldsymbol{x}_{ij}, \boldsymbol{y}_{ij}) \geq 0$ and  $d_E(\boldsymbol{x}_{ij}, \boldsymbol{y}_{ij})=0$ if and only if $\boldsymbol{x}_{ij}=\boldsymbol{y}_{ij}$.
		
		By summing over $i$ and $j$, we can obtain
		\begin{equation*}
			d_V = (n-2)! \sum_{1 \leq i < j \leq n}^{n} d_E(\boldsymbol{x}_{ij}, \boldsymbol{y}_{ij}) \geq 0,
		\end{equation*}
	    and
	    \begin{equation*}
            d_V = (n-2)! \sum_{1 \leq i < j \leq n}^{n} d_E(\boldsymbol{x}_{ij}, \boldsymbol{y}_{ij}) = 0,
	    \end{equation*}
	    if and only if $\boldsymbol{x}_{ij}=\boldsymbol{y}_{ij}$ for all integers $i$ and $j$ that satisfy the condition $1 \leq i < j \leq n$, which implies that $\boldsymbol{x} = \boldsymbol{y}$. We thus conclude that $d_V(\boldsymbol{x}, \boldsymbol{y})$ satisfies non-negativity and identity.
	    
		(2) Symmetry.
		
		Due to the symmetry of the Euclidean distance, that is, 
		
		\begin{equation*}
			d_E(\boldsymbol{x}_{ij}, \boldsymbol{y}_{ij}) = d_E(\boldsymbol{y}_{ij}, \boldsymbol{x}_{ij}),
		\end{equation*}
		
		summing over $i$ and $j$ yields
		
		\begin{equation*}
			\sum_{1 \leq i < j \leq n}^{n} d_E(\boldsymbol{x}_{ij}, \boldsymbol{y}_{ij}) = \sum_{1 \leq i < j \leq n}^{n} d_E(\boldsymbol{y}_{ij}, \boldsymbol{x}_{ij}),
		\end{equation*}			
		multiplying both sides by $(n-2)!$, the equality still holds, that is, $d_V(\boldsymbol{x}, \boldsymbol{y}) = d_V(\boldsymbol{y}, \boldsymbol{x})$, which means that $d_V(\boldsymbol{x}, \boldsymbol{y})$ satisfies symmetry.
		
		(3) Triangle inequality.
		
		Since the Euclidean distance satisfies the triangle inequality, that is, for any $\boldsymbol{x}_{ij}, \boldsymbol{y}_{ij}, \boldsymbol{z}_{ij}$, always holds
		\begin{equation*}
            d_E(\boldsymbol{x}_{ij}, \boldsymbol{y}_{ij}) \leq d_E(\boldsymbol{x}_{ij}, \boldsymbol{z}_{ij}) + d_E(\boldsymbol{z}_{ij}, \boldsymbol{y}_{ij}).
		\end{equation*}
		
		By summing over all indices $i$ and $j$ and multiplying both sides by $(n-2)!$, the inequality still holds, i.e. 
		\begin{equation*}
            d_V(\boldsymbol{x}, \boldsymbol{y}) \leq d_V(\boldsymbol{x}, \boldsymbol{z}) + d_V(\boldsymbol{z}, \boldsymbol{y}),
	    \end{equation*}
		where 
		\begin{equation*}
			\begin{gathered}
				d_V(\boldsymbol{x}, \boldsymbol{z}) = (n-2)! \sum_{1 \leq i < j \leq n} ^{n} d_E(\boldsymbol{x}_{ij}, \boldsymbol{z}_{ij}), \\
				d_V(\boldsymbol{z}, \boldsymbol{y}) = (n-2)! \sum_{1 \leq i < j \leq n} ^{n} d_E(\boldsymbol{z}_{ij}, \boldsymbol{y}_{ij}).
            \end{gathered}
        \end{equation*}
		Therefore, $d_V(\boldsymbol{x}, \boldsymbol{y})$ satisfies the triangle inequality.
			
		In summary, $d_V(\boldsymbol{x}, \boldsymbol{y})$ is a distance metric.
	\end{proof}
			
	In the sense of the View distance metric, a $3D$ space is composed of three $2D$ planes. A $4D$ space is composed of four $3D$ spaces or twelve $2D$ planes (including six non-overlapping $2D$ planes). A $5D$ space is composed of five $4D$ spaces, or twenty $3D$ spaces, or sixty $2D$ planes (including ten non-overlapping $2D$ planes). Although the formulation does not specify how these $n!/2$ $2D$ planes constitute the $nD$ space, it still provides a novel perspective and mathematical tool for analyzing metric spaces.
		
	\noindent \textbf{Remark 1.} According to the formula, within the high-dimensional space, the View distance between any pair of points incorporates a constant factor of $(n-2)!$. For the purpose of streamlining numerical computations, the formula (1) presented below will be utilized for all subsequent derivations and analyses. Notably, the revised metric, obtained by eliminating the constant term $(n-2)!$, retains compliance with the three axioms without any loss of validity.
	
	\begin{equation}\label{formula_1}
		d_V(\boldsymbol{x}, \boldsymbol{y}) = \sum_{1\leq{i}<j\leq{n}}^{n} \sqrt{(x_i - y_i)^{2} + (x_j - y_j)^2}.
	\end{equation}
		
	\noindent \textbf{Remark 2.} Based on formula (1), for any $\boldsymbol{x}, \boldsymbol{y} \in {\mathbb{R}^n}$, $d_V(\boldsymbol{x}, \boldsymbol{y}) \geq d_E(\boldsymbol{x}, \boldsymbol{y})$, where $n\ge{2}$. More generally, for $n \geq 3$, the View distance between any two non-zero points $\boldsymbol{x}$ and $\boldsymbol{y}$ is greater than the $L_{p}$ distance.
	
	\begin{figure}[H]
		\centering
		\includegraphics[width=0.5\textwidth]{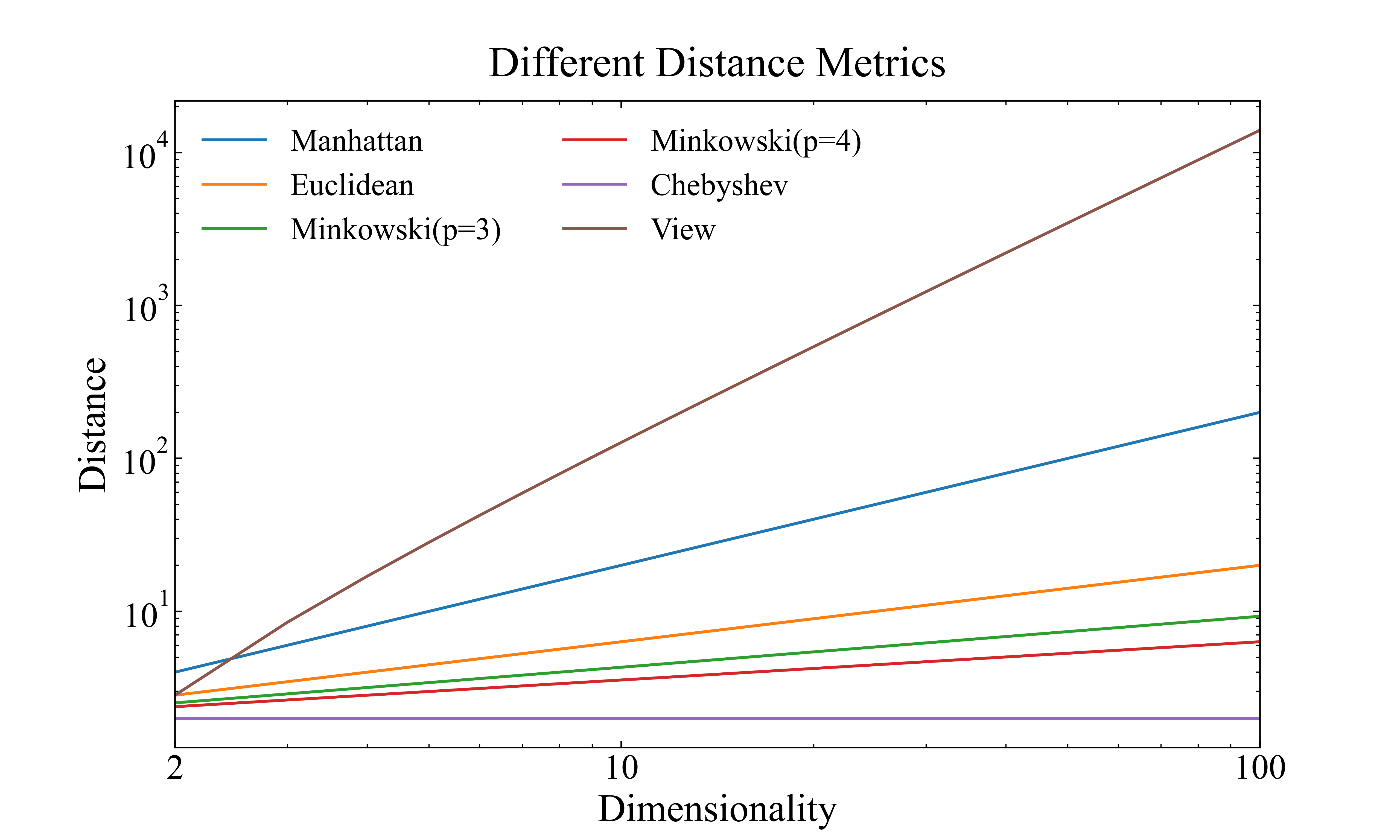}
		\captionsetup{width=0.90\textwidth, font=footnotesize}
		\caption{The distances of two points $-\mathbf{1}$ and $\mathbf{1}$ under different distance metrics in spaces of different dimensions, where $\mathbf{1}=(1,1,...,1)$. As the dimension increases, the View distance between two points becomes much larger than other distance metrics. The graph uses a logarithmic scale for both the $x$ and $y$ coordinate axes.}
	\end{figure}
		
	\noindent \textbf{Remark 3.} Suppose the $k$-th dimension is a constant, i.e., $x_{k}=y_{k}=c$ for any $\boldsymbol{x}, \boldsymbol{y} \in {\mathbb{R}^n}$. For the constant dimension $k$, the square of dimensional difference satisfies
	\begin{equation*}
		(x_k-y_k)^2=(c-c)^2=0.
	\end{equation*}

    The View distance is defined as the sum of Euclidean distances over all pairwise $2D$ projections of dimensions. The sum of projection terms for all dimension pairs $(k,j)$ satisfies
	\begin{equation*}
        \displaystyle \sum_{\substack{j=1 \\ j \ne k}}^{n} \sqrt{(x_{k} - y_{k})^{2} + (x_{j} - y_{j})^{2}}= \displaystyle \sum_{\substack{j=1 \\ j \ne k}}^{n} \sqrt{0^{2} + (x_{j} - y_{j})^{2}}= \displaystyle \sum_{\substack{j=1 \\ j \ne k}}^{n} |x_{j} - y_{j}|.
	\end{equation*}
	    
    For conventional $L_p$ metrics, including Euclidean distance, constant dimensions are ignored in the distance calculation since their differences are zero. In contrast, for the View distance, any $2D$ projection pair $(k,j)$ that includes a constant dimension $k$ will compute the absolute difference $|x_{j} - y_{j}|$ for the non-constant dimension $j$ and then sum it up.
		
    \subsection{$L_{v}$-norm}\label{sec3_2}
	At the end of this section, we present the definition of the $L_{v}$-norm and then prove that it satisfies the three axioms of a vector norm, namely the conditions of non-negativity and identity, homogeneity, the triangle inequality.
	\begin{definition}\label{norm}
		Given a vector $\boldsymbol{x}=(x_1,x_2,...,x_n)\in{\mathbb{R}^{n}}$, $L_{v}$-norm is defined by
		\begin{equation}
			\lVert{\boldsymbol{x}}\rVert_v=\sum_{1\leq{i}<j\leq{n}}^{n}\sqrt{x_i^2+x_j^2}.
		\end{equation}
	\end{definition}
	\begin{proof}
		For any $\boldsymbol{x}, \boldsymbol{y} \in {\mathbb{R}^n}$, let $\boldsymbol{x} = (x_1, x_2, ..., x_n)$, $\boldsymbol{y} = (y_1, y_2, \ldots, y_n)$.
			
		(1) Non-negativity and identity.
			
		For any indices $i,j$ with $1 \leq i < j \leq n$,
		\begin{equation*}
			\sqrt{x_i^2+x_j^2} \geq 0,
		\end{equation*}
		summing over $i$ and $j$ yields
		\begin{equation*}
			\lVert{\boldsymbol{x}}\rVert_v = \sum_{1\leq{i}<j\leq{n}}^{n}\sqrt{x_i^2+x_j^2} \geq 0.
		\end{equation*}
		\indent The equality 
		\begin{equation*}
			\lVert{\boldsymbol{x}}\rVert_v=0
		\end{equation*} holds if and only if 
		\begin{equation*}
			x_i = 0, x_j = 0
		\end{equation*}
		for all pairs of indices $i,j$ with $1 \leq i < j \leq n$, which implies that $x_i=0$, i.e., $\boldsymbol{x} = \boldsymbol{0}$. It follows that $L_{v}$-norm fulfills the non-negativity and identity axiom as required.
			
		(2) Homogeneity.
			
		For any $\alpha \in \mathbb{R}$, 
		\begin{equation*}
			\begin{gathered}
				\lVert{\alpha \boldsymbol{x}}\rVert_v = 	\sum_{1\leq{i}<j\leq{n}}^{n}\sqrt{(\alpha x_i)^{2}+(\alpha x_j)^{2}} =\lvert \alpha \rvert  \sum_{1\leq{i}<j\leq{n}}^{n}\sqrt{ x_i^{2}+x_j^{2}} = \lvert \alpha \rvert \lVert{\boldsymbol{x}}\rVert_v,
			\end{gathered}
		\end{equation*}		
		which means that $\lVert{\alpha \boldsymbol{x}}\rVert_v = \lvert \alpha \rvert \lVert{\boldsymbol{x}}\rVert_v$, i.e., $L_{v}$-norm satisfies homogeneity.
			
		(3) Triangle inequality.
			
		From the formulas of
		\begin{equation*}
			\begin{gathered}
				\lVert{\boldsymbol{x} + \boldsymbol{y}}\rVert_v = \sum_{1\leq{i}<j\leq{n}}^{n}\sqrt{(x_i + y_i)^{2} + (x_j + y_j)^{2}}, \\
				\lVert{\boldsymbol{x}}\rVert_v + \lVert{\boldsymbol{y}}\rVert_v = \sum_{1\leq{i}<j\leq{n}}^{n}\sqrt{x_i^2 + x_j^2} + \sum_{1\leq{i}<j\leq{n}}^{n}\sqrt{y_i^2 + y_j^2},
			\end{gathered}
		\end{equation*}
		we reduce the problem to comparing
		\begin{equation*}
			\begin{gathered}
			    \lVert{\boldsymbol{x}_{ij} + \boldsymbol{y}_{ij}}\rVert_{2} = \sqrt{(x_i + y_i)^{2} + (x_j + y_j)^{2}}, \\
			    \lVert{\boldsymbol{x}_{ij}}\rVert_{2} + \lVert{\boldsymbol{y}_{ij}}\rVert_{2} = \sqrt{x_i^2 + x_j^2} + \sqrt{y_i^2 + y_j^2}.
		    \end{gathered}
		\end{equation*}
		
		According to the Minkowski inequality, we have
		\begin{equation*}
			\lVert{\boldsymbol{x}_{ij} + \boldsymbol{y}_{ij}}\rVert_{2} \leq \lVert{\boldsymbol{x}_{ij}}\rVert_{2} + \lVert{\boldsymbol{y}_{ij}}\rVert_{2}.
		\end{equation*}		

		\indent Summing over all $i,j$ with $1 \leq i < j \leq n$, the inequality still holds, i.e.,
		\begin{gather*}
			\lVert{\boldsymbol{x} + \boldsymbol{y}}\rVert_v \leq \lVert{\boldsymbol{x}}\rVert_v + \lVert{\boldsymbol{y}}\rVert_v.
		\end{gather*}
		\indent Therefore, $L_v$-norm satisfies the triangle inequality.			
	\end{proof}
    In summary, the $L_v$-norm is a vector norm defined on $\mathbb{R}^n$. However, it does not satisfy the parallelogram law. Consequently, it cannot induce an inner product, and the associated normed linear space does not constitute a Hilbert space. Geometrically, this means that while we can rigorously define the length of vectors within this space, the concept of angles and orthogonality remains undefined.
    
    \section{Probabilistic interpretation of View distance}\label{sec4}
    \subsection{Statistical  characteristics of $L_{v}$-norm}\label{sec4_1}
    In this section, we examine the statistical properties of $L_{v}$-norm, aiming to characterize its distributional pattern and its variation with dimensionality in the sample space.
        
    Let $\boldsymbol{\xi}=(\xi_1, \xi_2, \dots, \xi_n)$ is an $nD$ random variable (where each component $\xi_i$ is a random variable), the $L_v$-norm is a non-negative random variable
    \begin{equation*}
       	\lVert{\boldsymbol{\xi}}\rVert_v=\sum_{1\leq{i}<j\leq{n}}^{n}\sqrt{\xi_i^2+\xi_j^2}.
    \end{equation*}
        
    Let the joint distribution function of $(\xi_i, \xi_j)$ be $F(x_i, x_j)$, the probability density function is $f(\xi_i, \xi_j)$. Let $\eta_{ij} = \sqrt{\xi_i^2 + \xi_j^2}$, the expectation of $\lVert{\boldsymbol{\xi}}\rVert_v$ is
    \begin{equation}
   		\begin{gathered}
       		E\left( \lVert{\boldsymbol{\xi}}\rVert_v \right) = E\left( \sum_{1 \leq i < j \leq n}^{n} \eta_{ij} \right) = \sum_{1 \leq i < j \leq n}^{n} E\left( \eta_{ij} \right) \\
       	    = \sum_{1 \leq i < j \leq n}^{n} \int_{-\infty}^{+\infty} \int_{-\infty}^{+\infty} \sqrt{x_i^2 + x_j^2} dF(x_i, x_j) = \sum_{1 \leq i < j \leq n}^{n} \int_{-\infty}^{+\infty} \int_{-\infty}^{+\infty} \sqrt{x_i^2 + x_j^2} f(x_i, x_j) dx_i dx_j.
       	\end{gathered}
    \end{equation}
    \indent The variance of $\lVert{\boldsymbol{\xi}}\rVert_v$ is
    \begin{equation}
        D\left( \lVert{\boldsymbol{\xi}}\rVert_v \right) = D\left( \sum_{1 \leq i < j \leq n}^{n} \eta_{ij} \right) = \sum_{1 \leq i < j \leq n}^{n} D(\eta_{ij}) + \sum_{\substack{1 \leq i < j \leq n \\ 1 \leq k < l \leq n \\ (i,j)\neq (k,l)}}^{n} \mathrm{cov}(\eta_{ij}, \eta_{kl}).
    \end{equation}
        
    Particularly, if $\xi_1, \xi_2, \dots, \xi_n$ are independent and identically distributed (i.i.d.), then
    \begin{equation*}
       	E(\eta_{ij}) = E(\eta_{kl}) \ \text{and} \  D(\eta_{ij}) = D(\eta_{kl}).
    \end{equation*}
    \indent Moreover, when $\eta_{ij}$ and $\eta_{kl}$ satisfy $i \neq k$ and $j \neq l$, they are mutually independent. In this case,
    \begin{gather*}
        \mathrm{cov}(\eta_{ij}, \eta_{kl}) = E\left( \left[ \eta_{ij} - E(\eta_{ij}) \right] \left[ \eta_{kl} - E(\eta_{kl}) \right] \right) = 0.
    \end{gather*}
    \indent The expectation of $\lVert{\boldsymbol{\xi}}\rVert_v$ is
    \begin{equation}
        E\left( \lVert{\boldsymbol{\xi}}\rVert_v \right) = \sum_{1 \leq i < j \leq n}^{n} E\left( \eta_{ij} \right) = \frac{n(n-1)}{2} E\left( \eta_{ij} \right).
    \end{equation}
    \indent By the symmetry of covariance, 
    \begin{gather*}
       	\mathrm{cov}(\eta_{ij}, \eta_{kl}) = \mathrm{cov}(\eta_{kl}, \eta_{ij}),
    \end{gather*}
    the variance of $\lVert{\boldsymbol{\xi}}\rVert_v$ is
    \begin{equation}        	
       	D\left( \lVert{\boldsymbol{\xi}}\rVert_v \right) = D\left( \sum_{1 \leq i < j \leq n}^{n} \eta_{ij} \right) = \frac{n(n-1)}{2} D(\eta_{ij}) + 2 \cdot \frac{n(n-1)(n-2)}{2} \mathrm{cov}(\eta_{ij}, \eta_{kl}),
    \end{equation} 		
    where exactly one element is equal between $(i,j)$ and $(k,l)$.
    
    Under the assumption that the components $\xi_i$ are i.i.d., although $\eta_{ij}$ and $\eta_{kl}$ are not independent, they follow the same distribution and thus have equal second moments. By the Cauchy-Schwarz inequality
    \begin{gather*}
       	\left( E(\xi_1 \cdot \xi_2) \right)^2 \leq E(\xi_1^2) \cdot E(\xi_2^2),
    \end{gather*}
    we can obtain
    \begin{gather*}
        \left\lvert E\left( \eta_{ij} \cdot \eta_{kl} \right) \right\rvert \leq \sqrt{E\left( \eta_{ij}^2 \right) \cdot E\left( \eta_{kl}^2 \right)} = \left\lvert E\left( \eta_{ij}^{2} \right) \right\rvert,
    \end{gather*}
    when both sides of the inequality are non-negative,       
    \begin{gather*}
       	\mathrm{cov}(\eta_{ij}, \eta_{kl}) = E\left( \left[ \eta_{ij} - E(\eta_{ij}) \right] \left[ \eta_{kl} - E(\eta_{kl}) \right] \right) \\
       	= E\left( \eta_{ij} \cdot \eta_{kl} \right) - \left( E\left( \eta_{ij} \right) \right)^{2} \leq E\left( \eta_{ij}^{2} \right) - \left( E\left( \eta_{ij} \right) \right)^{2},
    \end{gather*}
    i.e.,
    \begin{gather*}
        \mathrm{cov}(\eta_{ij}, \eta_{kl}) \leq  D\left( \eta_{ij} \right).
    \end{gather*}
    \indent The upper bound of the variance is
    \begin{gather*}
        D\left( \lVert{\boldsymbol{\xi}}\rVert_v \right) \leq \frac{2n^3 - 5n^2 + 3n}{2} D(\eta_{ij}),
    \end{gather*}
    where
    \begin{equation*}
        D(\eta_{ij}) = E(\eta_{ij}^{2}) - \left( E\left(\eta_{ij} \right) \right)^{2} = 2 E\left( \xi_{i}^{2} \right) - \left( E\left( \sqrt{\xi_i^2 + \xi_j^2} \right) \right)^{2}.
    \end{equation*}
            
    \indent Based on the formulas of expectation and variance, the View distance exhibits distinctive statistical properties that differ from traditional $L_p$ metrics.
    
    \begin{itemize}
       	\item The expectation of View distance is linearly superposed by the independent expectations of each $2D$ projection distance without any cross-coupling terms.
       	\item The variance of View distances is a composite term. It includes not only the individual variance terms for the $n(n-1)/2$ $2D$ projection distances, but also the pairwise covariance terms between any two distinct $2D$ projection distances, which capture the statistical correlation between different projection components.
    \end{itemize}
    
    Specifically, the View distance is obtained by summing the distances of $n(n-1)/2$ $2D$ projections. Each feature can form $n-1$ cross-projection pairs with the other features, namely $(x_1,x_i), \dots,(x_{i-1},x_i),(x_i,x_{i+1}),\dots,$ $(x_i,x_n)$. Even under the assumption of i.i.d., different $2D$ projection components remain statistically dependent. This dependence introduces non-zero covariance terms in the total variance calculation, which ultimately prevents the overall variance of the View distance from being decomposed into a sum of independent one-dimensional variances. We conduct numerical simulation experiments based on the $nD$ standard uniform distribution (denoted as $U(0,1)$ for short) and the $nD$ standard normal distribution (denoted as $\mathcal{N}(0,1)$ for short). We perform visualization analysis on the expectation and variance of the $L_{v}$-norm under the i.i.d. assumption, thereby verifying the accuracy and validity of the theoretically derived formulas.
    
    The experimental results show that the theoretical calculation values for the $L_{v}$-norm expectation are highly consistent with the numerical simulation results, and the actual variances are all lower than the theoretically derived upper bounds.
    
    \begin{figure}[H]
    	\centering
    	\begin{minipage}{0.3\linewidth}
    		\centering
    		\includegraphics[width=\linewidth]{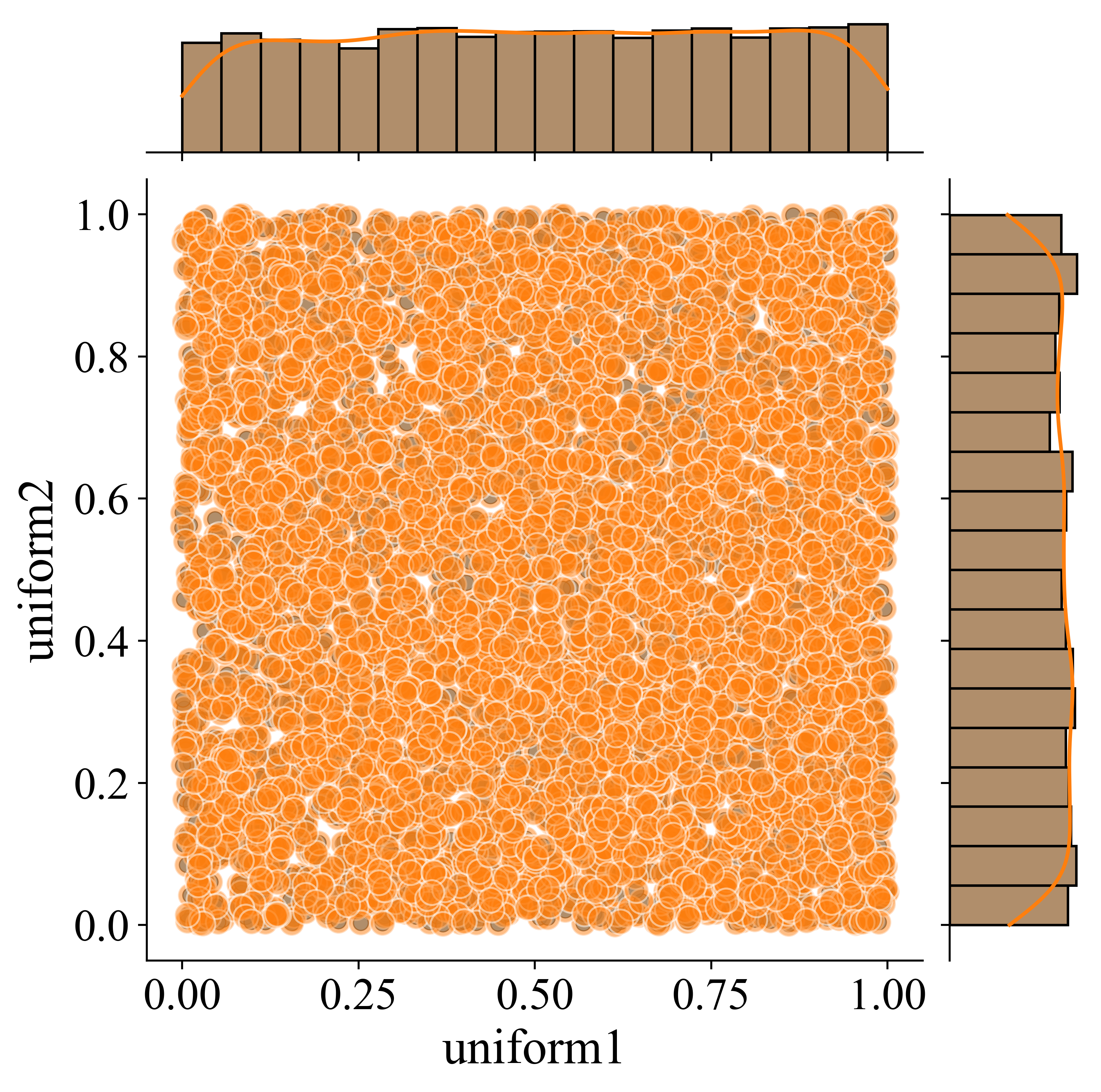}
    	\end{minipage}
    	\hspace{0.02\linewidth}
    	\begin{minipage}{0.3\linewidth}
    		\centering
    		\includegraphics[width=\linewidth]{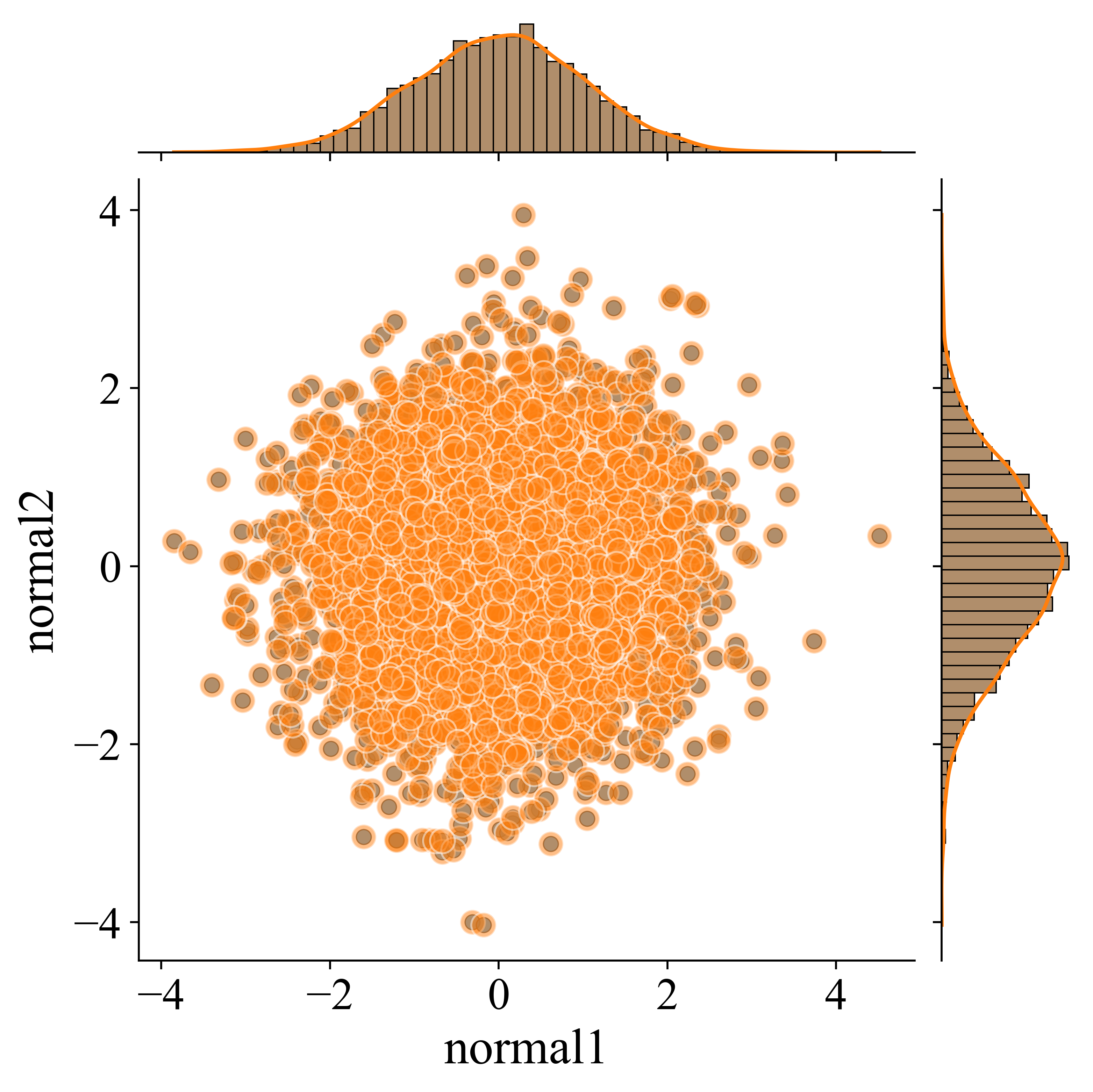}
    	\end{minipage}
    	\captionsetup{width=0.90\linewidth,font=footnotesize}
    	\caption{Schematic diagrams of the $2D$ $U(0,1)$ and $2D$ $\mathcal{N}(0,1)$.}
    \end{figure}
    
    \noindent \textbf{Mode 1.} Let the $nD$ random variable $\boldsymbol{\xi} = (\xi_1, \xi_2,$ $\dots, \xi_n)^{\top}$, where each component $\xi_i \sim U(0,1)$ are i.i.d., and let $\eta_{ij} = \sqrt{\xi_i^2 + \xi_j^2}$, then
    \begin{gather*}
		E\left( \lVert{\boldsymbol{\xi}}\rVert_v \right) = \frac{n(n-1)}{2}   \cdot E\left( \eta_{ij} \right), \\
		D\left( \lVert{\boldsymbol{\xi}}\rVert_v \right) \leq \frac{2n^3 - 5n^2  + 3n}{2} \cdot D(\eta_{ij}),
	\end{gather*}
    where
    \begin{gather*}
       	E\left( \eta_{ij} \right) = \frac{\sqrt{2} + \ln(1 + \sqrt{2})}{3}, \\
       	D(\eta_{ij}) = \frac{2}{3} - \left( \frac{\sqrt{2} + \ln(1 + \sqrt{2})}{3} \right)^2.
    \end{gather*}
            
    \noindent \textbf{Mode 2.} Let the $nD$ random variable $\boldsymbol{\xi} = (\xi_1, \xi_2,$ $\dots, \xi_n)^{\top}$, where each component $\xi_i \sim \mathcal{N}(0,1)$ are i.i.d., and let $\eta_{ij} = \sqrt{\xi_i^2 + \xi_j^2}$ (i.e., the Rayleigh distribution with scale parameter $\sigma = 1$), then
    \begin{gather*}
       	E\left( \lVert{\boldsymbol{\xi}}\rVert_v \right) = \frac{n(n-1)}{2} \cdot E\left( \eta_{ij} \right), \\
       	D\left( \lVert{\boldsymbol{\xi}}\rVert_v \right) \leq \frac{2n^3 - 5n^2 + 3n}{2} \cdot D(\eta_{ij}),
    \end{gather*}
    where
    \begin{gather*}
       	E\left( \eta_{ij} \right) = \sqrt{\frac{\pi}{2}}, \\
       	D(\eta_{ij}) = 2 - \frac{\pi}{2}.
    \end{gather*}
    
    \begin{figure}[H]
    	\centering
    	\begin{minipage}{0.41\linewidth}
    		\centering
    		\includegraphics[width=\linewidth]{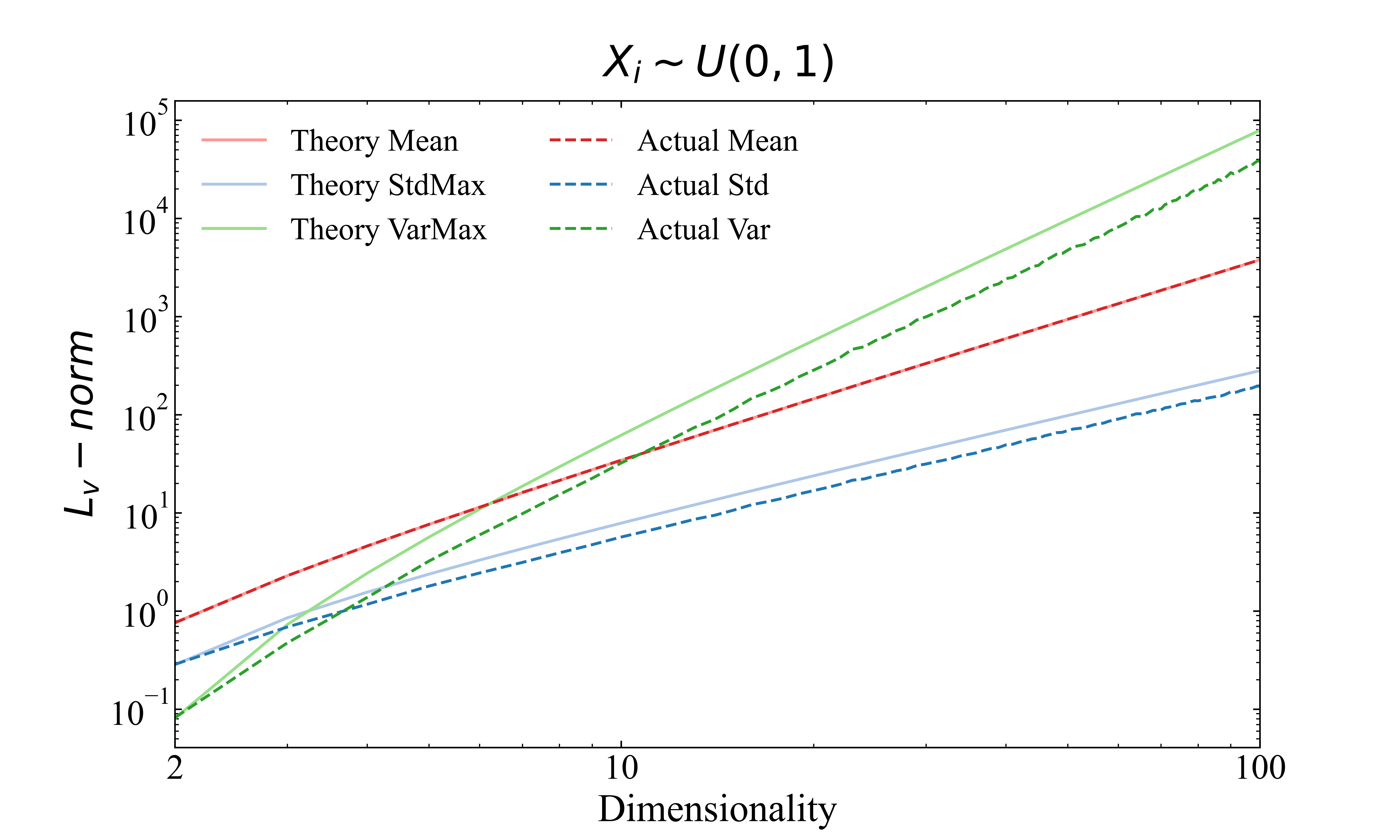}
    	\end{minipage}
    	\hspace{0.01\linewidth}
    	\begin{minipage}{0.41\linewidth}
    		\centering
    		\includegraphics[width=\linewidth]{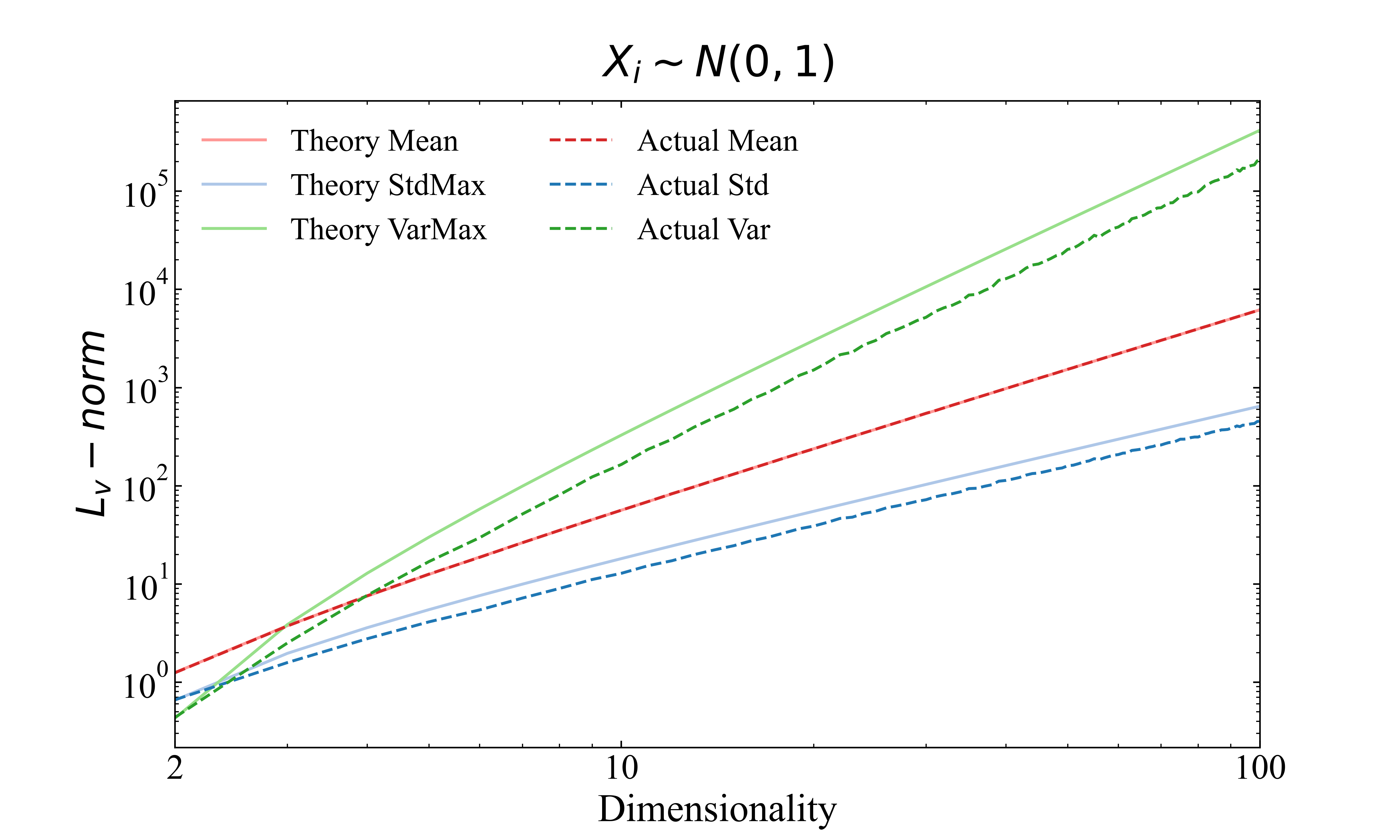}
    	\end{minipage}
    	\captionsetup{width=0.90\linewidth,font=footnotesize}
    	\caption{The theoretical values of the expectation and maximum variance of the $L_{v}$-norm for the $nD$ $U(0,1)$ and $\mathcal{N}(0,1)$, as well as the empirical values simulated from 5000 data points. The graph uses a logarithmic scale for both the $x$ and $y$ coordinate axes.}
    \end{figure}
    
    \subsection{Distance concentration}
    Subject to the i.i.d. assumption, the expectation of the View distance distribution exhibits a rapid growth trend at a rate of $n^2$, whereas the upper bound of its standard deviation scales at a rate of $n^{3/2}$. The ratio of the upper bound of the standard deviation to the expectation is given by $n^{-1/2}$. When the dimensionality $n$ tends to infinity, $n^{-1/2}$ approaches 0 asymptotically. This implies that in the high-dimensional regime, the View distance will tend to concentrate progressively around the central point as the dimensionality increases.
  
    In the $nD$ unit hypercube, the maximum distance from a vector to the origin under the $L_v$-norm is given by
    \begin{equation*}
        \frac{n(n-1)}{2} \cdot \sqrt{2}.
    \end{equation*}
    
    As the dimension increases, the average distance converges to its expected value. For vectors following the $U(0,1)$ within the $nD$ unit hypercube, after normalization by the maximum distance, the average distance converges to
    \begin{equation*}
    	\frac{\sqrt{2} + \ln(1 + \sqrt{2})}{3\sqrt{2}} \approx 0.541,
    \end{equation*}
    while the corresponding average distance for the $L_2$-norm converges to $1/\sqrt{3} \approx 0.577$.
      
    For vectors following the $\mathcal{N}(0,1)$, after normalization by the maximum distance, the average distance converges to
    \begin{equation*}
    	\sqrt{\frac{\pi}{4}} \approx 0.886,
    \end{equation*}
    and the corresponding average distance for the $L_2$-norm converges to $1$.
      
    \begin{figure}[H]
    	\centering
    	\begin{minipage}{0.40\linewidth}
    		\centering
    		\includegraphics[width=\linewidth]{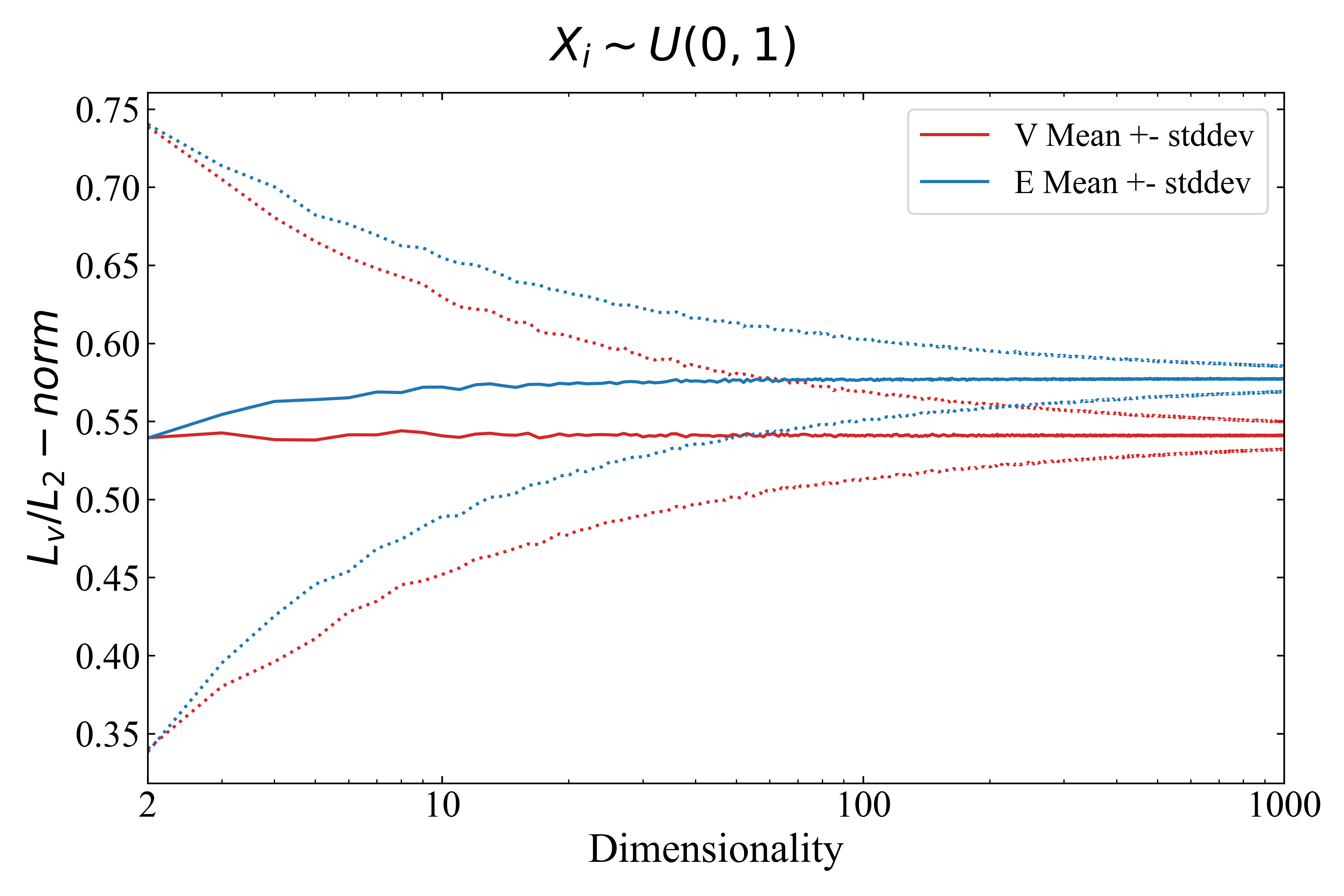}
    	\end{minipage}
    	\hspace{0.05\linewidth}
    	\begin{minipage}{0.40\linewidth}
    		\centering
    		\includegraphics[width=\linewidth]{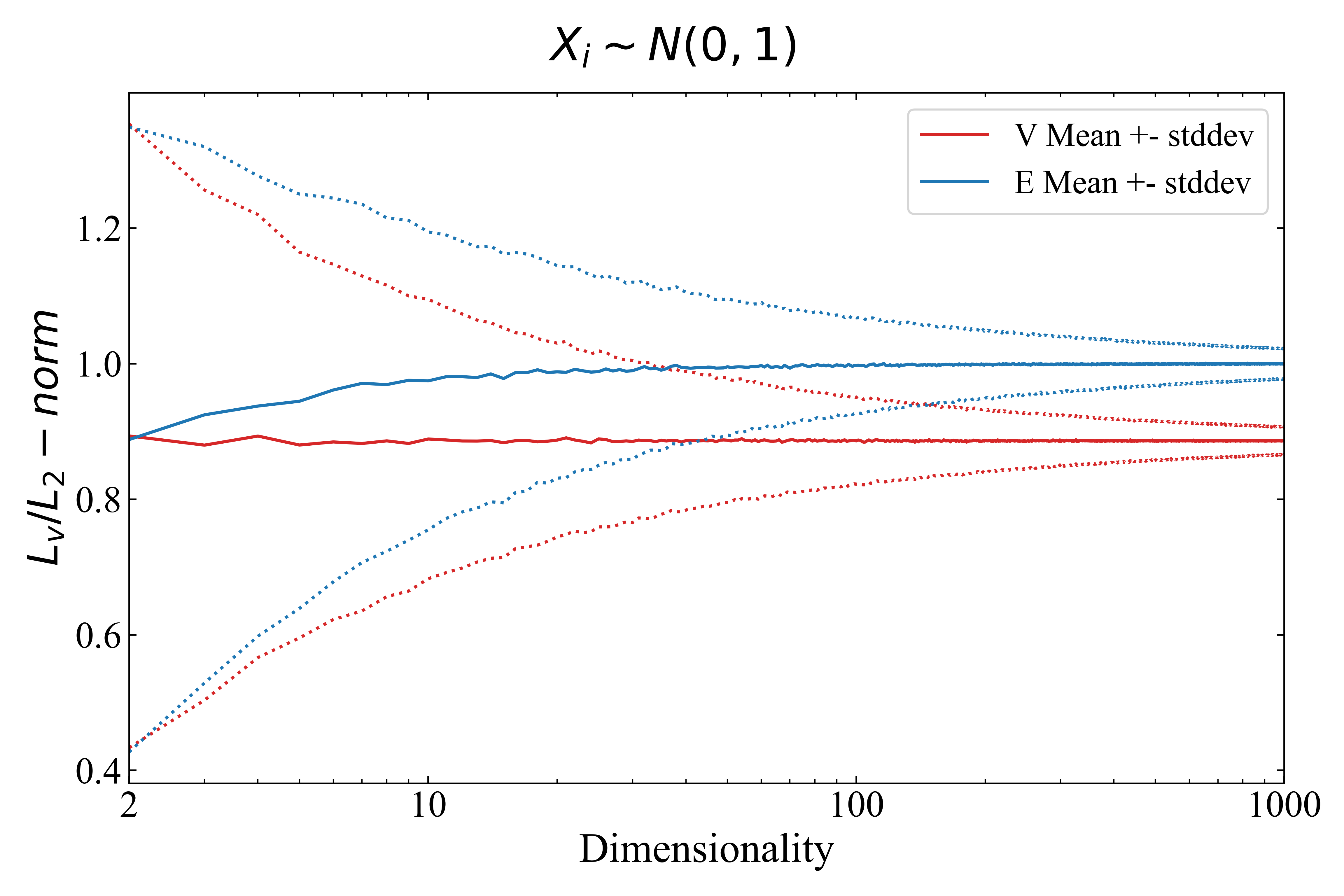}
    	\end{minipage}
    	\captionsetup{width=0.90\linewidth,font=footnotesize}
    	\caption{Variation trends of the mean values of the $L_{v}$-norm and the $L_{2}$-norm with dimensionality after maximum normalization under $nD$ $U(0,1)$ and $\mathcal{N}(0,1)$. Both the $x$‑axis and the $y$‑axis are plotted on a logarithmic scale.}
    	\label{NormConcentrationMean}
    \end{figure}

    \begin{figure}[H]
    	\centering
    	\begin{minipage}{0.40\linewidth}
    		\centering
    		\includegraphics[width=\linewidth]{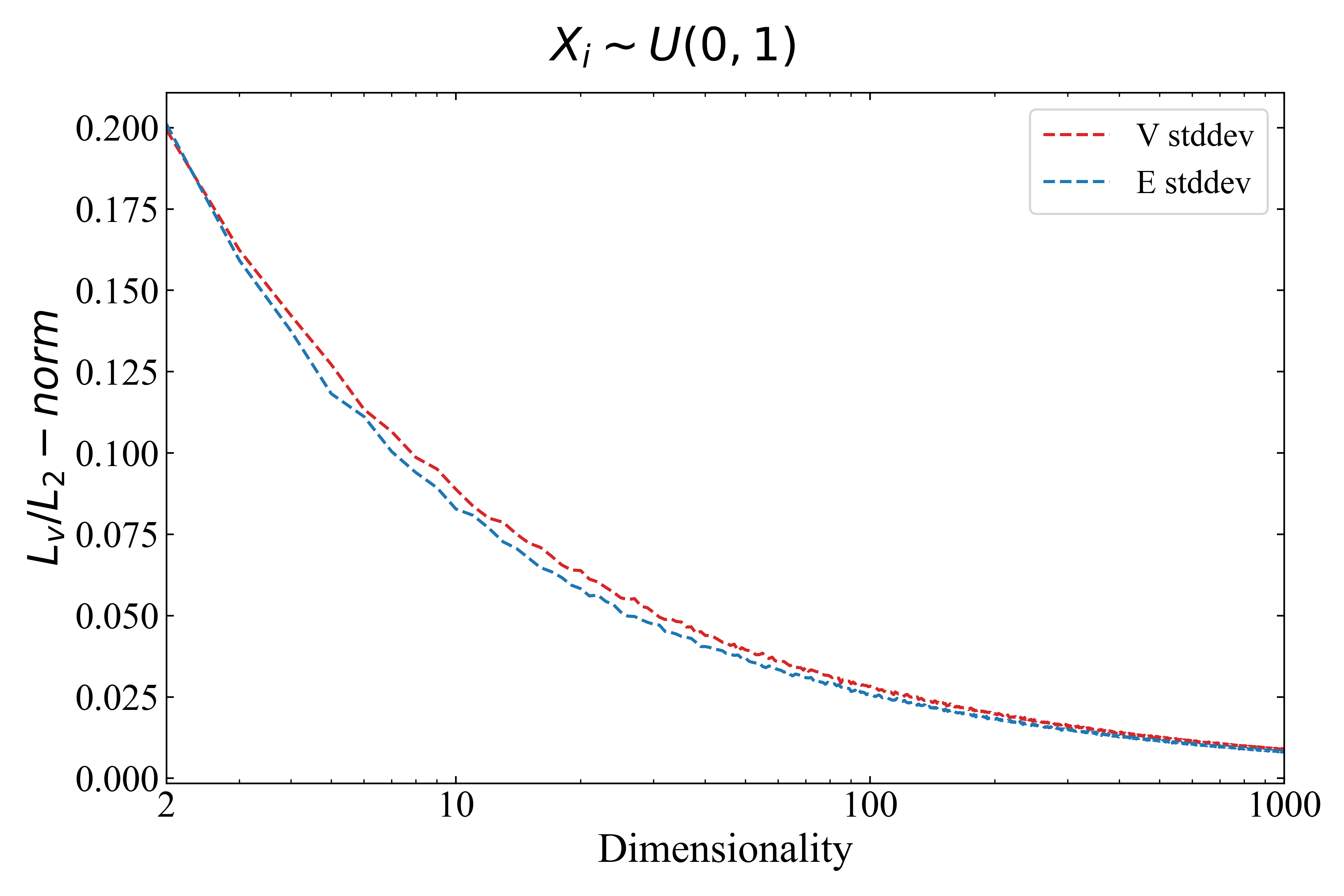}
    	\end{minipage}
    	\hspace{0.05\linewidth}
    	\begin{minipage}{0.40\linewidth}
    		\centering
    		\includegraphics[width=\linewidth]{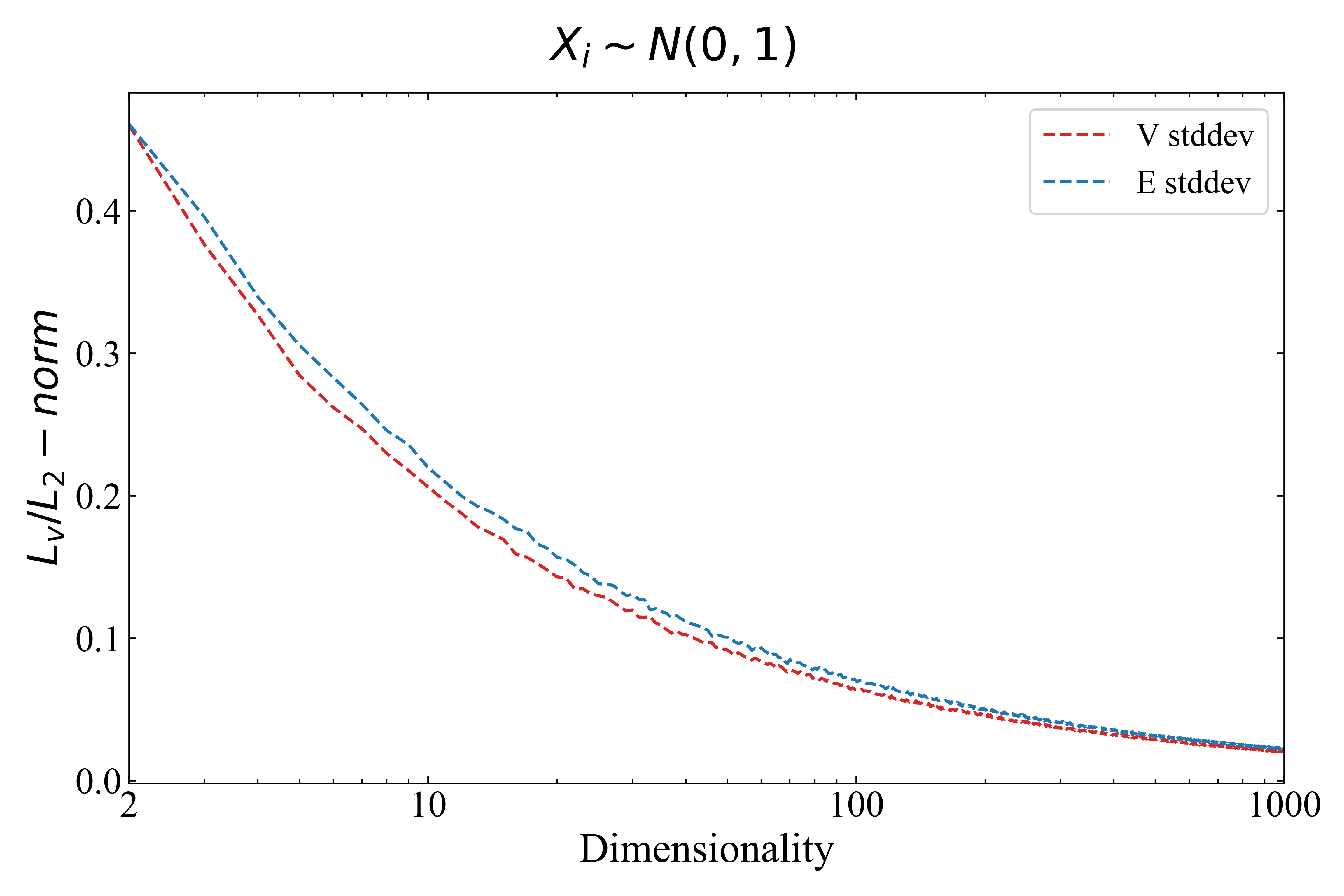}
    	\end{minipage}
    	\captionsetup{width=0.90\linewidth,font=footnotesize}
    	\caption{Variation trends of the standard deviations of the $L_{v}$-norm and the $L_{2}$-norm with dimensionality after maximum normalization under $nD$ $U(0,1)$ and $\mathcal{N}(0,1)$. Both the $x$‑axis and the $y$‑axis are plotted on a logarithmic scale.}
    	\label{NormConcentrationStd}
    \end{figure}
    
    Under the $U(0,1)$ and $\mathcal{N}(0,1)$, we study the variation of Euclidean distance and View distance with dimensionality after maximum normalization, as illustrated in Figures~\ref{NormConcentrationMean}-\ref{NormConcentrationStd}. Under both distributions, the mean value of the View distance remains stable with increasing dimensionality and is consistently lower than that of the Euclidean distance. Specifically, the standard deviation of the View distance is always larger than that of the Euclidean distance for uniformly distributed data, while it is consistently smaller for normally distributed data. This indicates that the View distance exhibits a weaker concentration effect than the Euclidean distance under uniform distribution.
    
    \subsection{Application of the $L_{v}$-norm in synthetic data}\label{sec4_2}
    To further explore the differences between View distance and Euclidean distance, we apply the $L_{v}$-norm to three categories of synthetic datasets as specified below and perform a comprehensive comparative analysis with the conventional $L_2$-norm\cite{ArthurZimek}.
        
    \textbf{Mode 1 and Mode 2.} All features follow an i.i.d. pattern.
    
    Based on the aforementioned \textbf{mode 1} and \textbf{mode 2}, we compare the vector lengths (i.e., distances from the origin) and pairwise distance of the $L_v$-norm and $L_2$-norm, and observe the following results.
      
    \begin{itemize}
    	\item The vector length is a form of specific simplification. However, identical results are obtained when examining the pairwise distances between any two points in the synthetic datasets.
    	\item The results derived from the $nD$ standard uniform distribution and standard normal distribution are nearly identical.
    \end{itemize}
    
    \begin{figure}[H]
    	\centering
    	\begin{minipage}{0.42\linewidth}
    		\centering
    		\includegraphics[width=\linewidth]{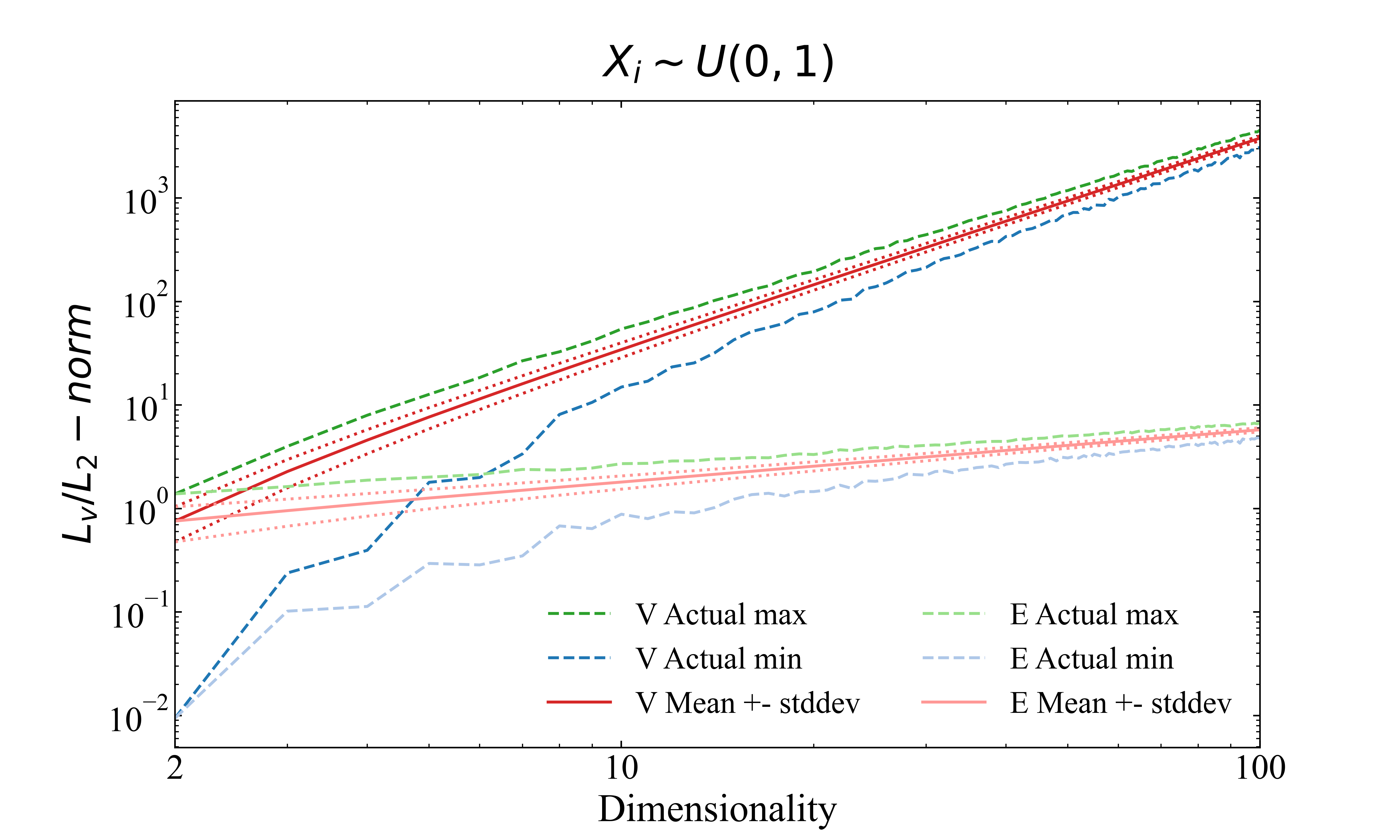}
    	\end{minipage}
    	\hspace{0.01\linewidth}
    	\begin{minipage}{0.42\linewidth}
    		\centering
    		\includegraphics[width=\linewidth]{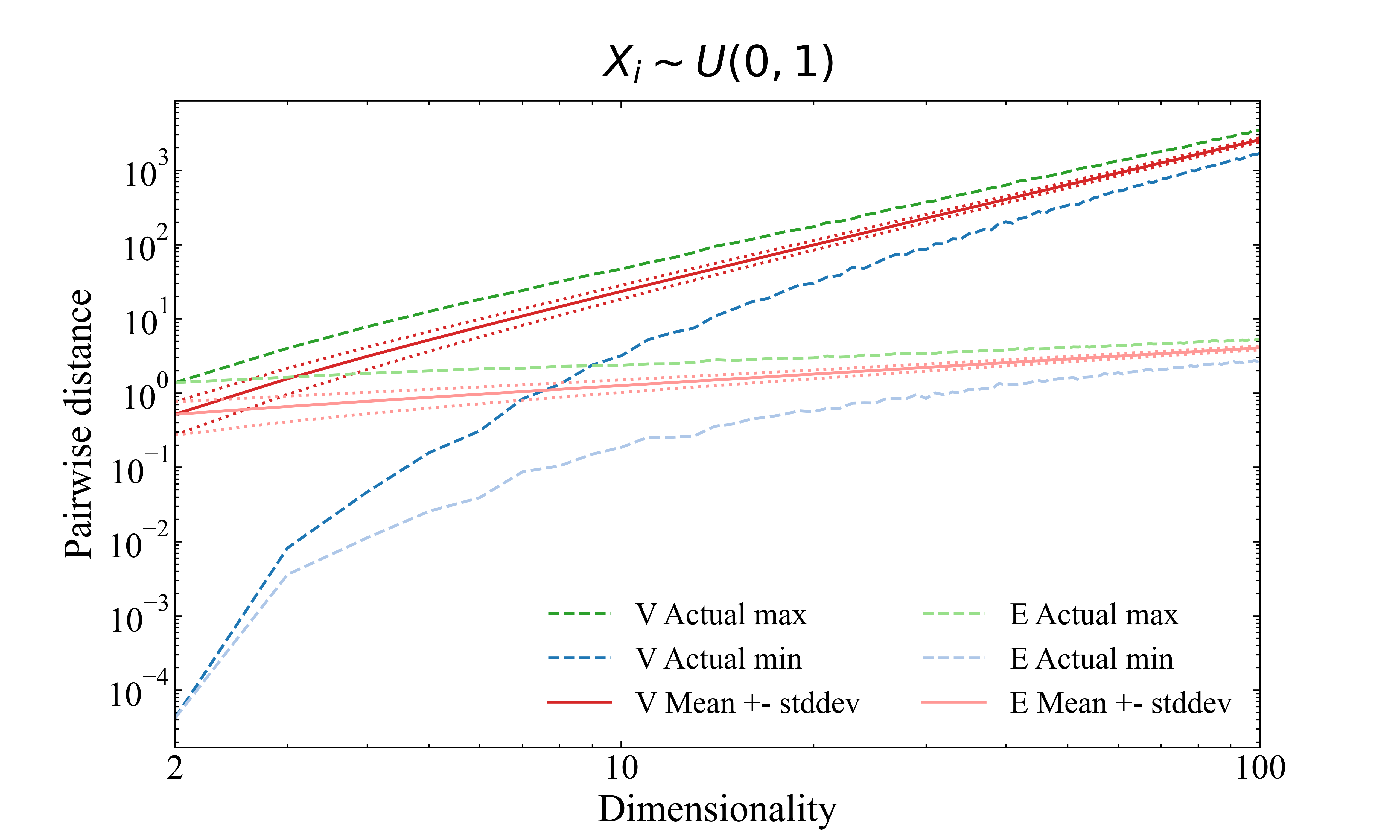}
    	\end{minipage}
    	\captionsetup{width=0.90\linewidth,font=footnotesize}
    	\caption{Lengths of the $L_{v}$-norm and the $L_{2}$-norm, pairwise View distance and Euclidean distance, where $X_{i} \sim U(0,1)$. Both the $x$‑axis and the $y$‑axis are plotted on a logarithmic scale.}
    \end{figure}
    
    \begin{figure}[H]
    	\centering
    	\begin{minipage}{0.42\linewidth}
    		\centering
    		\includegraphics[width=\linewidth]{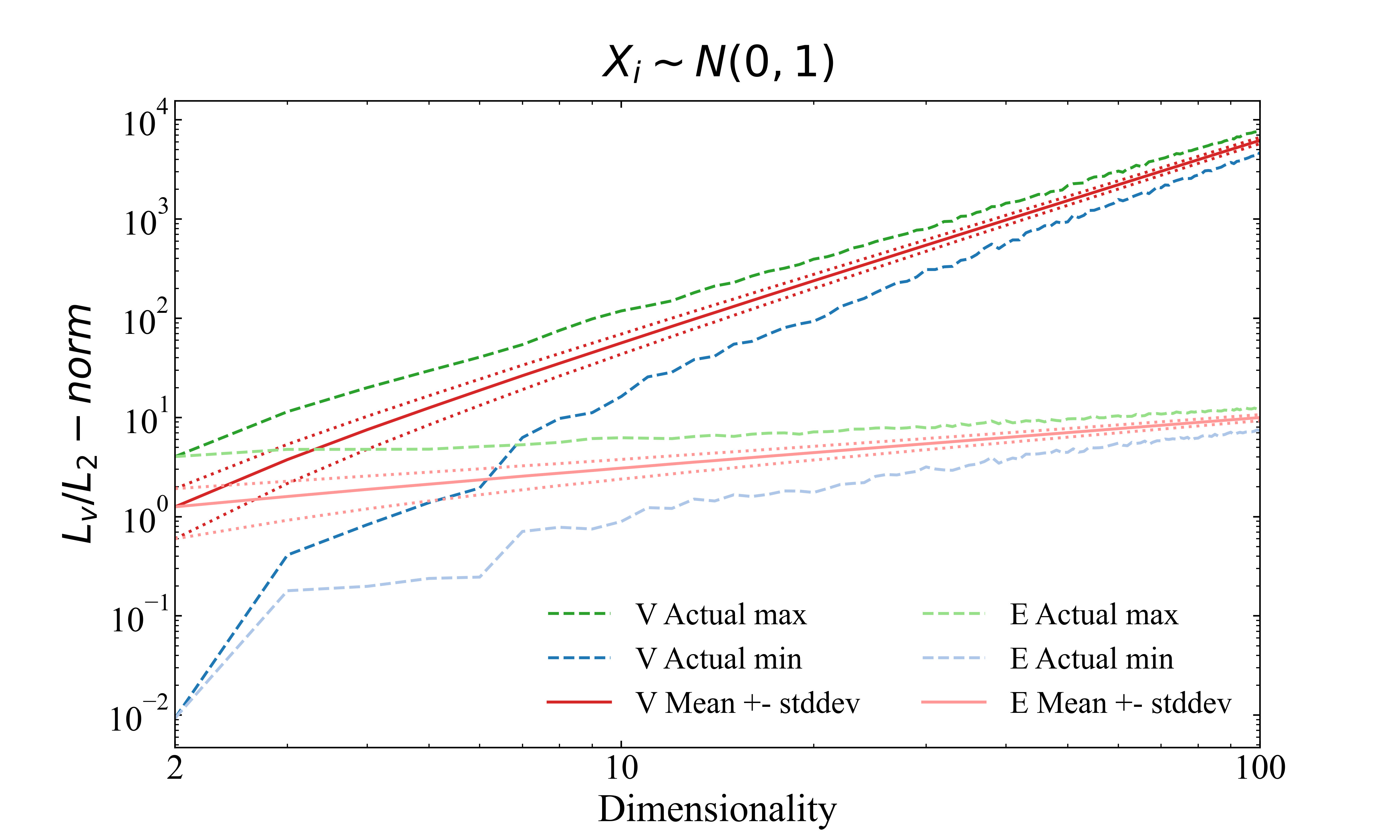}
    	\end{minipage}
    	\hspace{0.01\linewidth}
    	\begin{minipage}{0.42\linewidth}
    		\centering
    		\includegraphics[width=\linewidth]{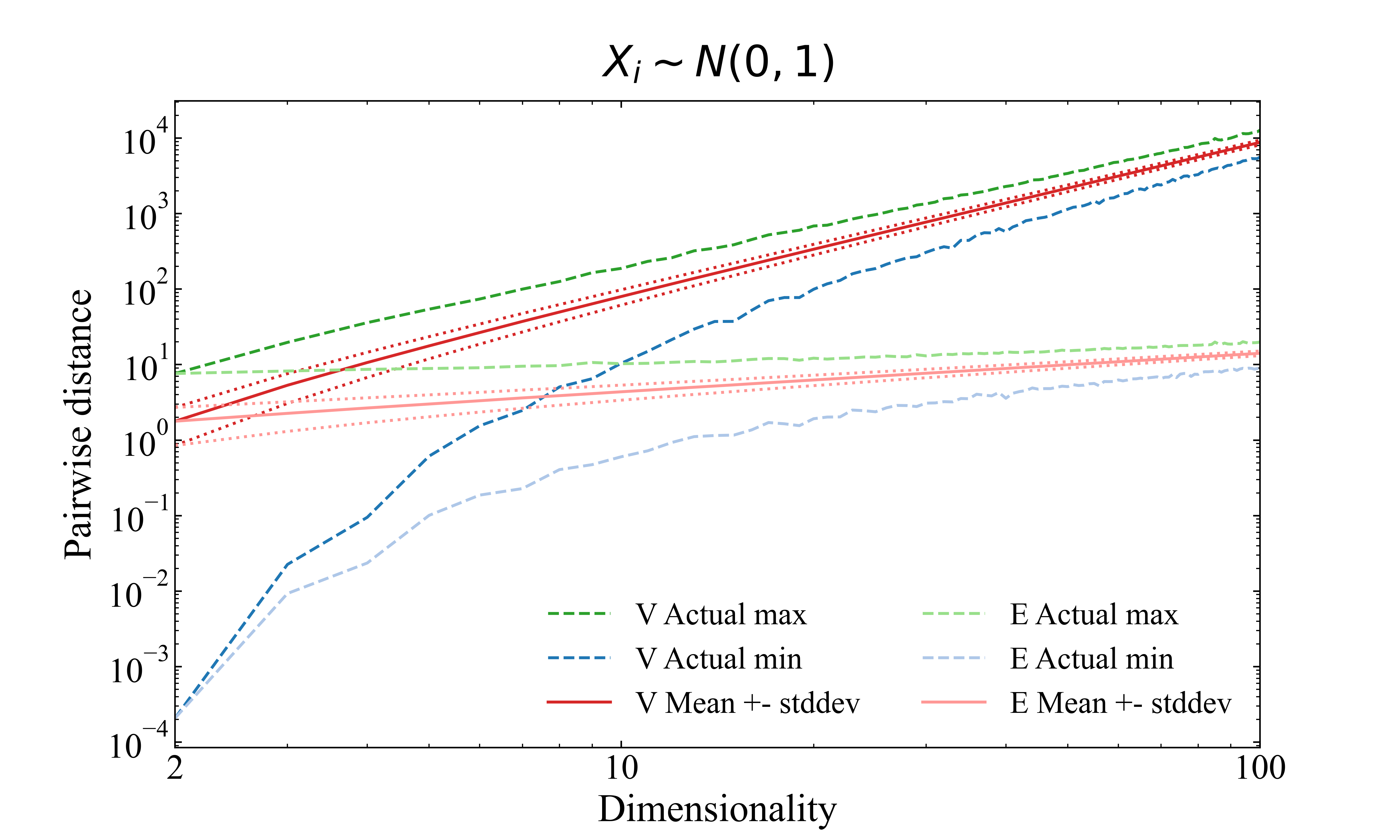}
    	\end{minipage}
    	\captionsetup{width=0.90\linewidth,font=footnotesize}
    	\caption{Lengths of the $L_{v}$-norm and the $L_{2}$-norm, pairwise View distance and Euclidean distance, where $X_{i} \sim \mathcal{N}(0,1)$. Both the $x$‑axis and the $y$‑axis are plotted on a logarithmic scale.}
    \end{figure}
    
    In the subsequent example, we will relax the i.i.d. assumption to explore whether there are differences between the two norms beyond numerical magnitude in scenarios where all features are intercorrelated.
    
    \textbf{Mode 3.} All features are mutually correlated.
        
    In this example, all features exhibit mutual correlations, with the variances of newly added features increasing progressively. The following is a description of the problem\cite{KevinBeyer}.
        
    We generate a dataset consisting of 5000 data points, with each data point $\boldsymbol{X_m} = (X_1, \dots, X_n)$ conforming to the following constraints:
    \begin{itemize}
       	\item First, we take independent random variables $U_1, \dots, U_n$ such that $U_i \sim U(0, \sqrt{i})$.
       	\item We define $X_1 = U_1$.
       	\item For all $2 \leq i \leq n$, define $X_i = U_i + (X_{i-1}/2)$.
    \end{itemize}
        
    The results demonstrate that, in scenarios where all features are intercorrelated, and the variance of newly added features gradually diverges as the dimension increases, both norms still converge toward their respective means with identical convergence trends and rates, without exhibiting significant differences.
    
    Finally, we discuss whether there are discrepancies between the two norms in the scenario where the variance of newly added features gradually approaches zero as the dimension increases.
    
    \begin{figure}[H]
    	\centering
    	\begin{minipage}{0.42\linewidth}
    		\centering
    		\includegraphics[width=\linewidth]{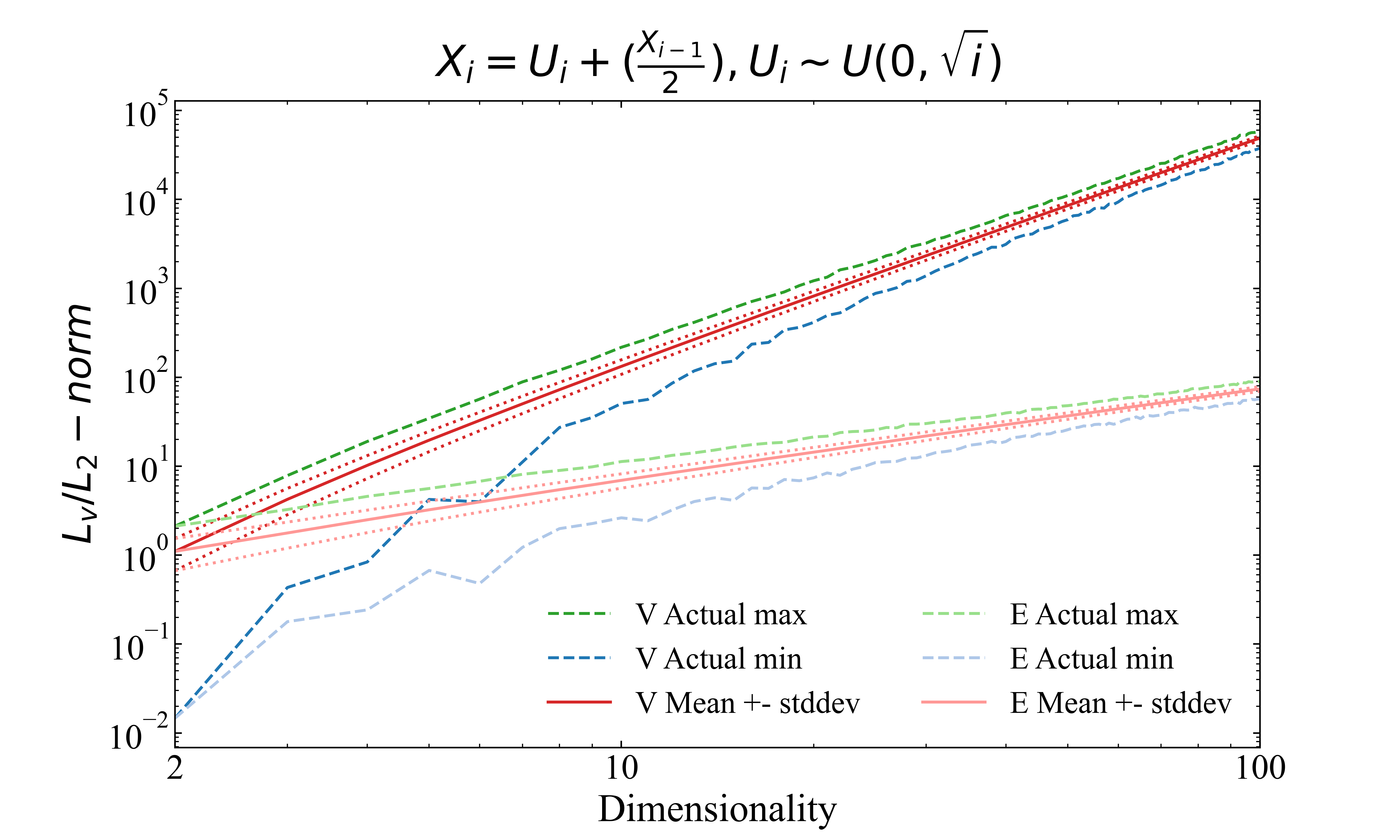}
    	\end{minipage}
    	\hspace{0.01\linewidth}
    	\begin{minipage}{0.42\linewidth}
    		\centering
    		\includegraphics[width=\linewidth]{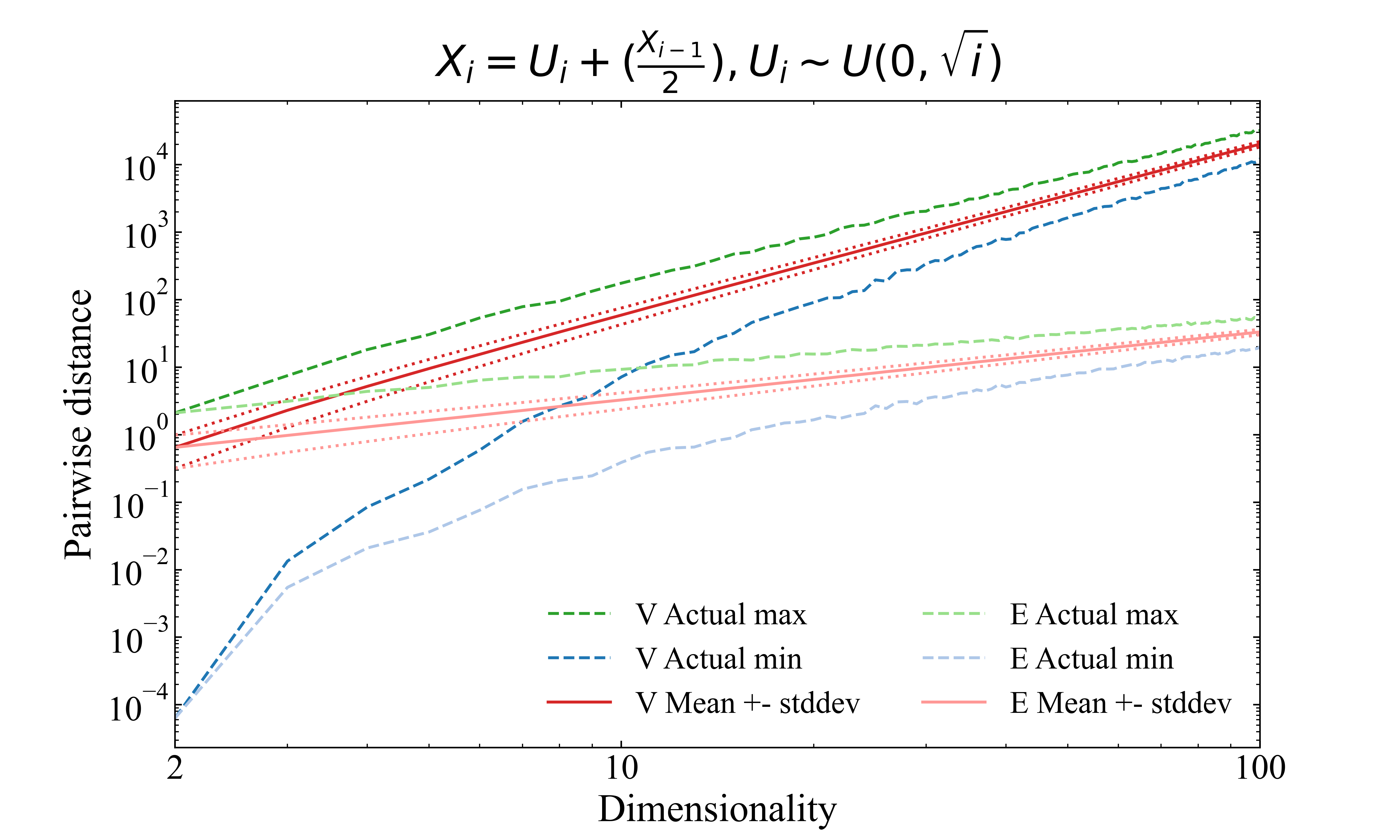}
    	\end{minipage}
    	\captionsetup{width=0.90\linewidth,font=footnotesize}
    	\caption{Lengths of the $L_{v}$-norm and the $L_{2}$-norm, pairwise View distance and Euclidean distance, where $X_{i}=U_{i}+(\frac{X_{i-1}}{2})$, $U_{i} \sim U(0,\sqrt{i})$. Both the $x$‑axis and the $y$‑axis are plotted on a logarithmic scale.}
    \end{figure}
                       
    \textbf{Mode 4.} Variance Converging to 0.
        
    The variances of the newly added features approach zero, indicating that the values of these features exhibit minimal fluctuations or even remain constant.
    
    We generate 5000 data points, where each point $\boldsymbol{X_m} = (X_1,X_2,\dots,X_n)$ satisfies the condition that each component $X_i$ is mutually independent and follows the normal distribution $X_i \sim \mathcal{N}(0, 1/i)$\cite{KevinBeyer}.
    
    In this case, a distinct discrepancy emerges between the two norms: the mean and standard deviation of the vector lengths under the $L_v$-norm continue to increase with dimension and gradually converge toward the mean. In contrast, the mean and standard deviation of the vector lengths under the $L_2$-norm stabilize gradually and remain unchanged.
    
    \begin{figure}[H]
    	\centering
    	\begin{minipage}{0.42\linewidth}
    		\centering
    		\includegraphics[width=\linewidth]{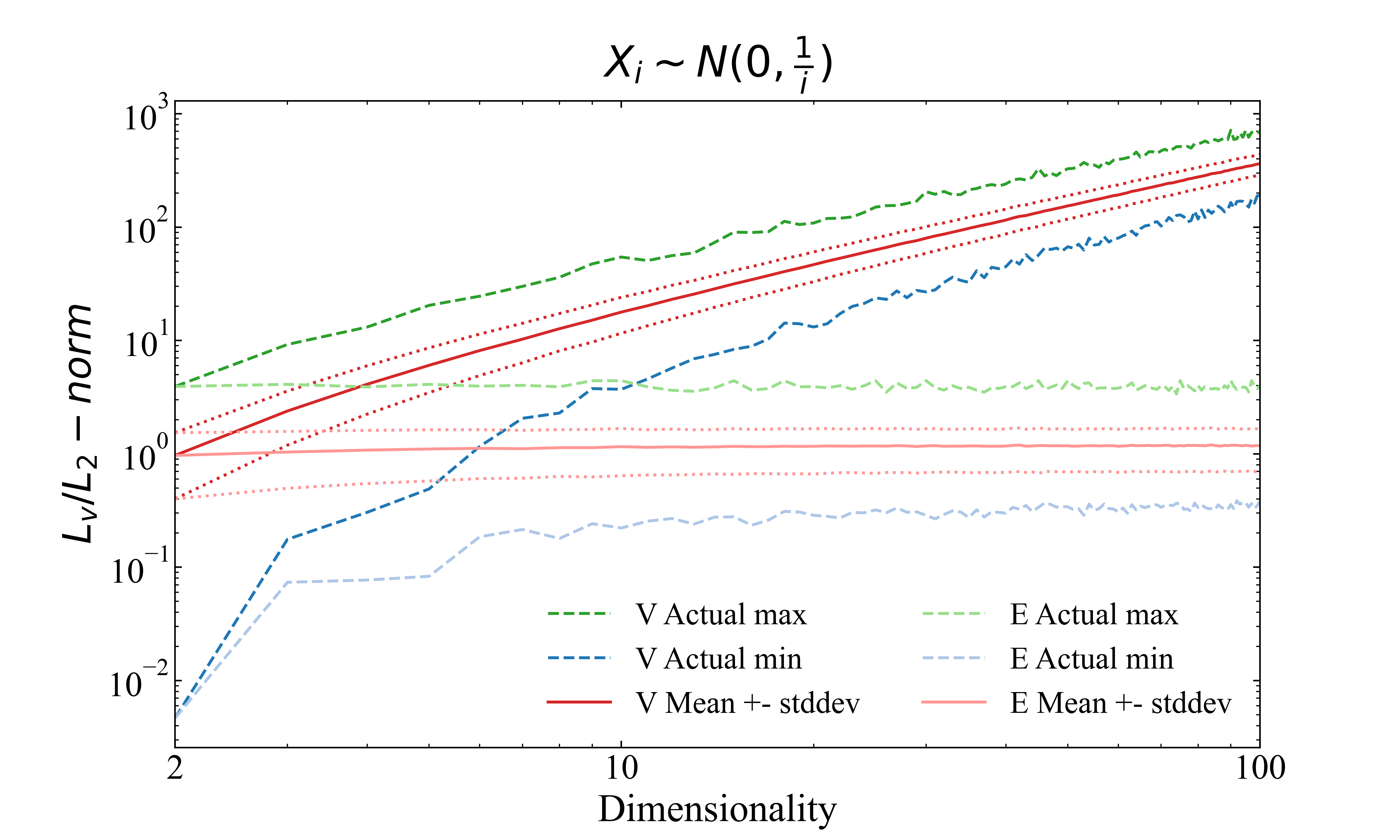}
    	\end{minipage}
    	\hspace{0.01\linewidth}
    	\begin{minipage}{0.42\linewidth}
    		\centering
    		\includegraphics[width=\linewidth]{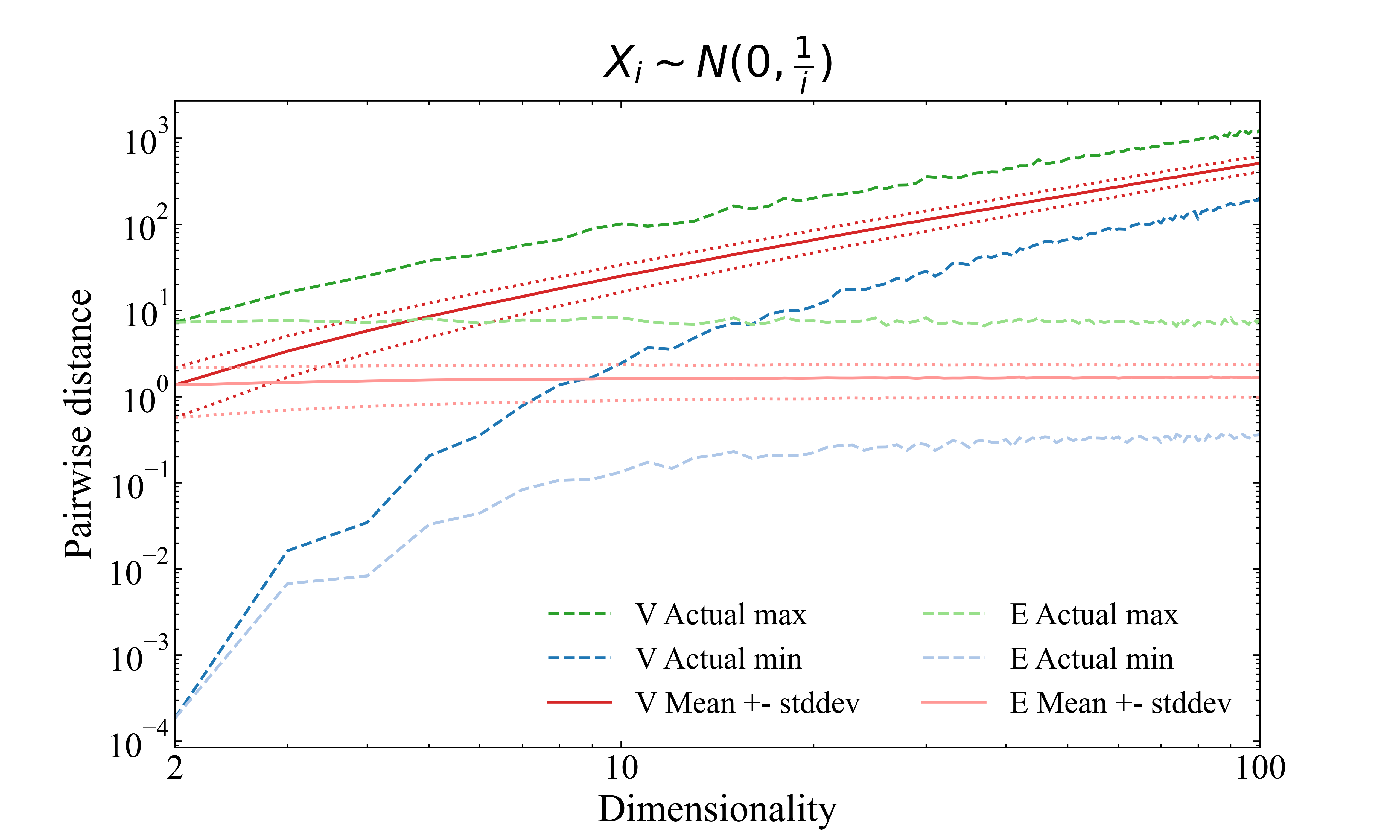}
    	\end{minipage}
    	\captionsetup{width=0.90\linewidth,font=footnotesize}
    	\caption{Lengths of the $L_{v}$-norm and the $L_{2}$-norm, pairwise View distance and Euclidean distance, where $X_{i} \sim N(0,\frac{1}{i})$. Both the $x$‑axis and the $y$‑axis are plotted on a logarithmic scale.}
    \end{figure}
                    
    According to Remark 3, in the context of the Euclidean distance, constant features (i.e., features with zero variance) contribute nothing to the distance calculation. As a result, such features cannot assist in distinguishing between two data points and are considered to carry no valid discriminative information. The same holds for the $L_p$ distance. Although constant features do not contribute to the discriminability of $L_p$ distances between data points, they are still integral to the feature space. They explicitly delineate the positions of data points in the high-dimensional space and constrain the scope of the data distribution. In this case, the data distribution corresponds to a low-dimensional subspace distribution within the high-dimensional space, which is fundamentally different from the distribution in the space that excludes these constant features. 
        
    In contrast, in the computation of the View distance, although the univariate difference of a constant feature is invariably zero, it can form multiple $2D$ projection pairs with non-constant features. Accumulating the absolute differences of the non-constant features can contribute to distance calculation and alter the distance distribution among samples. Specifically, in scenarios involving constant or low-variance features, the View distance constitutes a hybrid metric structure composed of the $L_1$ distances of non-constant features and the $L_2$ distances of their $2D$ projection pairs. This provides a novel metric paradigm for characterizing the underlying distribution or intrinsic structure of the data.
             	
    \section{Geometric interpretation of View distance}\label{sec5}
	\subsection{Unit sphere}\label{sec5_1}
	In this section, we analyze its geometry through the unit sphere of the View distance function, i.e., the region satisfying $L_v(\boldsymbol{x}) \leq 1$. Based on the definition of the $L_v$-norm in Section~\ref{sec3_2}, we denote its expression in functional form. For any point $\boldsymbol{x} = (x_1, x_2, \dots, x_n)$ in the $nD$ space $\mathbb{R}^n$, we define
	\begin{equation*}
		L_v(\boldsymbol{x}) = \sum_{1\le i<j\le n}^{n} \sqrt{x_i^2 + x_j^2}.
	\end{equation*}
	
	Meanwhile, we denote the functional forms of the $L_1$-norm\cite{RobertTibshirani}, $L_2$-norm, and Elastic Net\cite{HuiZou} (a combination of $L_1$-norm and $L_2$-norm) as follows
	\begin{equation*}
		L_1(\boldsymbol{x}) = \sum_{i=1}^n |x_i|,
	\end{equation*}
	\begin{equation*}
		L_2(\boldsymbol{x}) = \sqrt{\sum_{i=1}^n x_i^2},
	\end{equation*}
	\begin{equation*}
		L_{1+2}(\boldsymbol{x}) = \lambda R_1(x) + (1 - \lambda) L^{2}_{2}(x).
	\end{equation*}

	\noindent \textbf{Remark 4.} For the $3D$ space $(x,y,z)$ with parameter $t>0$, the View distance $L_v(x,y,z)$ simplifies along the coordinate‑axis direction as
	\begin{equation*}
		L_v(t,0,0) = 2t.
	\end{equation*}
	Along the space‑diagonal direction, we obtain
	\begin{equation*}
		L_v(t,t,t) = 3\sqrt{2}\,t.
	\end{equation*}
	
	\begin{table}[htbp]
		\centering
		\captionsetup{width=0.9\textwidth, font={footnotesize}}
		\caption{Distance propagation along coordinate-axis and space‑diagonal directions in $3D$ space.}
		\label{distancePropagationTab}
		\begin{tabular}{c|cccc|c}
			\toprule
			Direction & $L_{1}$ & $L_{2}$ & $L_{1+2} (\lambda=0.5)$ & $L_{v}$ & $L_{v}/L_{2}$ \\
			\midrule
			Axis & $t$ & $t$ & $t$ & $2t$ & 2\\
			Diagonal & $3t$ & $\sqrt{3}\,t$ & $\dfrac{3+\sqrt{3}}{2}t$ & $3\sqrt{2}\,t$ & $\sqrt{6}$ \\
			\bottomrule
		\end{tabular}
	\end{table}
	
	From Table~\ref{distancePropagationTab}, we observe that the View distance $L_v$ yields faster rates of change along coordinate‑axis and space‑diagonal directions than $L_1$ and $L_2$, revealing its heightened sensitivity to small‑magnitude feature perturbations. The ratio of $L_v$ to $L_2$ equals $2$ along the coordinate-axis direction and $\sqrt{6}$ along the diagonal direction. It reveals that the View distance is not a simple isotropically scaled variant of the Euclidean distance, but an inherently anisotropic metric.
			
    \begin{figure}[H]
		\centering
		\includegraphics[width=0.7\textwidth]{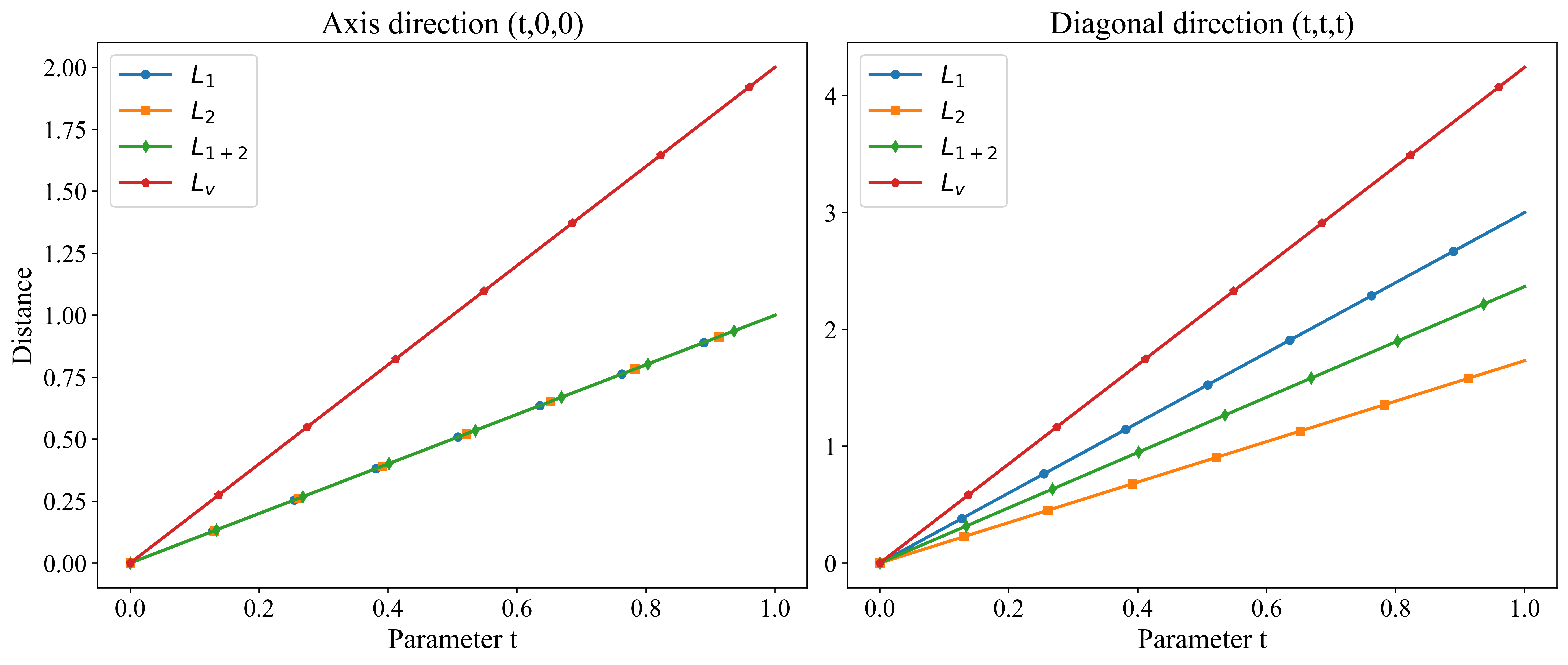}
		\captionsetup{font={scriptsize}}
		\caption{The distance propagation curves of $L_1$, $L_2$, $L_{1+2} (\lambda=0.5)$, and $L_v$ in $3D$ space.}
		\label{distancePropagation}
	\end{figure}
	
	\begin{theorem}
		\label{L1_norm_equivalent}
		In finite-dimensional spaces $\mathbb{R}^n$, the $L_v$-norm and the $L_{1}$-norm are equivalent, satisfying
		\begin{equation*}
			\frac{n-1}{\sqrt{2}}\|\boldsymbol{x}\|_1 \le \|\boldsymbol{x}\|_v \le (n-1)\|\boldsymbol{x}\|_1.
		\end{equation*}
	\end{theorem}
	
	\begin{proof}
		In the $\mathbb{R}^2$ space, from the equivalence relationship between the $L_{2}$-norm and $L_{1}$-norm, we can get 
		\begin{equation*}
			\frac{1}{\sqrt{2}}(|x_{i}|+|x_{j}|) \le \sqrt{x_{i}^{2}+x_{j}^{2}} \le |x_{i}|+|x_{j}|.
		\end{equation*}
		
		Summing over all $i,j$ with $1 \leq i < j \leq n$, the upper bound is
		\begin{equation*}
			\sum_{1 \le i<j \le n}^{n} (|x_i|+|x_j|) = (n-1)\sum_{i=1}^{n} |x_i| = (n-1)\|\boldsymbol{x}\|_1,
		\end{equation*}
		
		the lower bound is
		\begin{equation*}
			\sum_{1 \le i<j \le n}^{n} \frac{1}{\sqrt{2}} (|x_i|+|x_j|) = \frac{n-1}{\sqrt{2}}\sum_{i=1}^{n} |x_i| = \frac{n-1}{\sqrt{2}}\|\boldsymbol{x}\|_1,
		\end{equation*}
		
		the inequality still holds, i.e.,
		\begin{equation*}
			\frac{n-1}{\sqrt{2}}\|\boldsymbol{x}\|_1 \;\le\; \|\boldsymbol{x}\|_v \;\le\; (n-1)\|\boldsymbol{x}\|_1.
		\end{equation*}
	\end{proof}
	
	\begin{theorem}
		\label{L2_norm_equivalent}
		In finite-dimensional spaces $\mathbb{R}^n$, the $L_v$-norm and the $L_{2}$-norm are equivalent, satisfying
		\begin{equation*}
			\frac{n-1}{\sqrt{2}}\|\boldsymbol{x}\|_2 \le \|\boldsymbol{x}\|_v \le (n-1)\sqrt{n}\|\boldsymbol{x}\|_2.
		\end{equation*}
	\end{theorem}
	
	\begin{proof}
		In the $\mathbb{R}^2$ space, from the equivalence relationship between the $L_{2}$-norm and $L_{1}$-norm, we can get 
		\begin{equation*}
			\|x\|_2 \le \|x\|_1 \le \sqrt{n}\|x\|_2
		\end{equation*}
		
		According to Theorem~\ref{L1_norm_equivalent}, the inequality holds, i.e.,
		
		\begin{equation*}
			\frac{n-1}{\sqrt{2}}\|\boldsymbol{x}\|_2 \le \|\boldsymbol{x}\|_v \le (n-1)\sqrt{n}\|\boldsymbol{x}\|_2.
		\end{equation*}
	\end{proof}
	
 	\begin{figure}[H]
 		\centering
 		\begin{minipage}{0.4\linewidth}
 			\centering
 			\includegraphics[width=\linewidth]{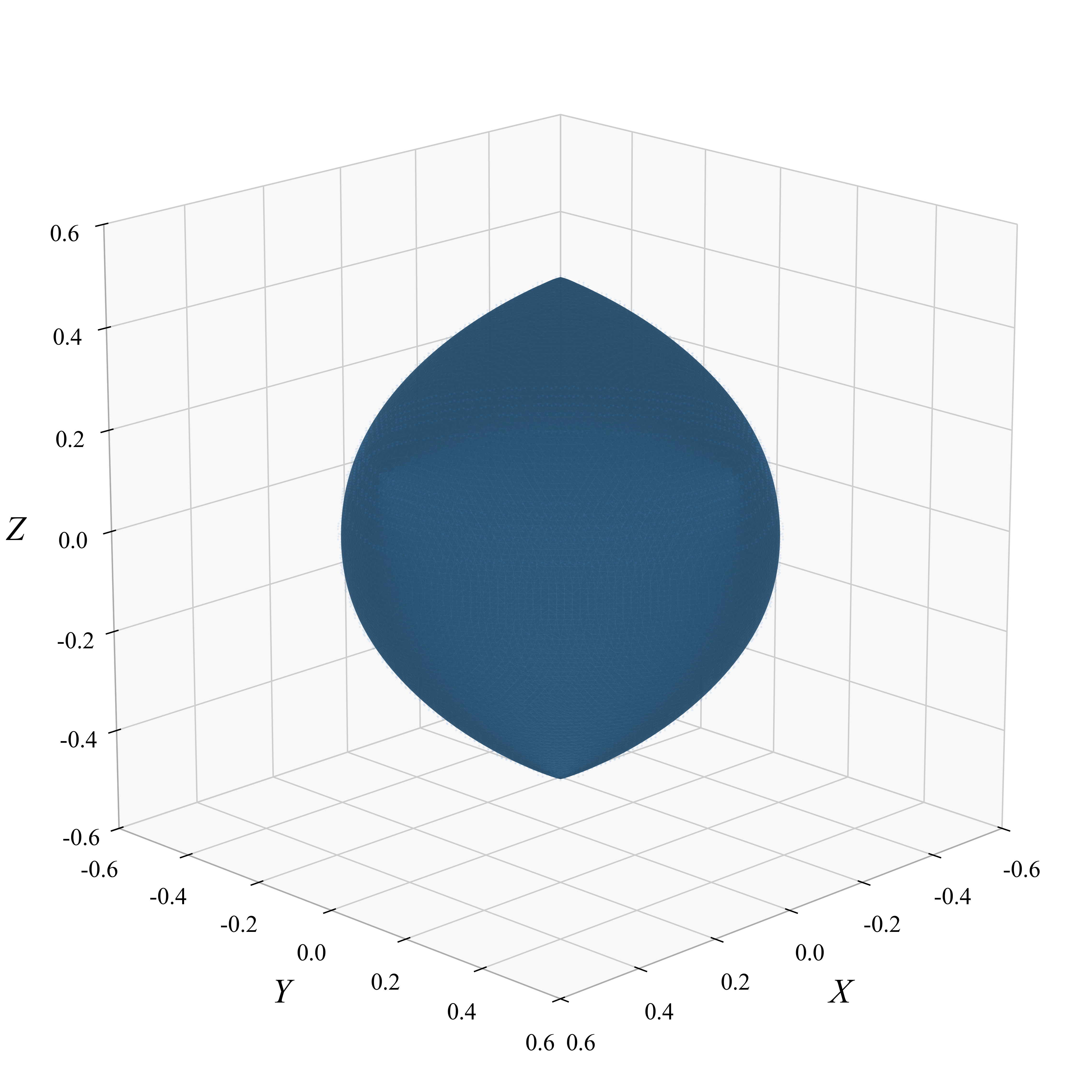}
 		\end{minipage}
 		\hspace{0.01\linewidth}
 		\begin{minipage}{0.4\linewidth}
 			\centering
 			\includegraphics[width=\linewidth]{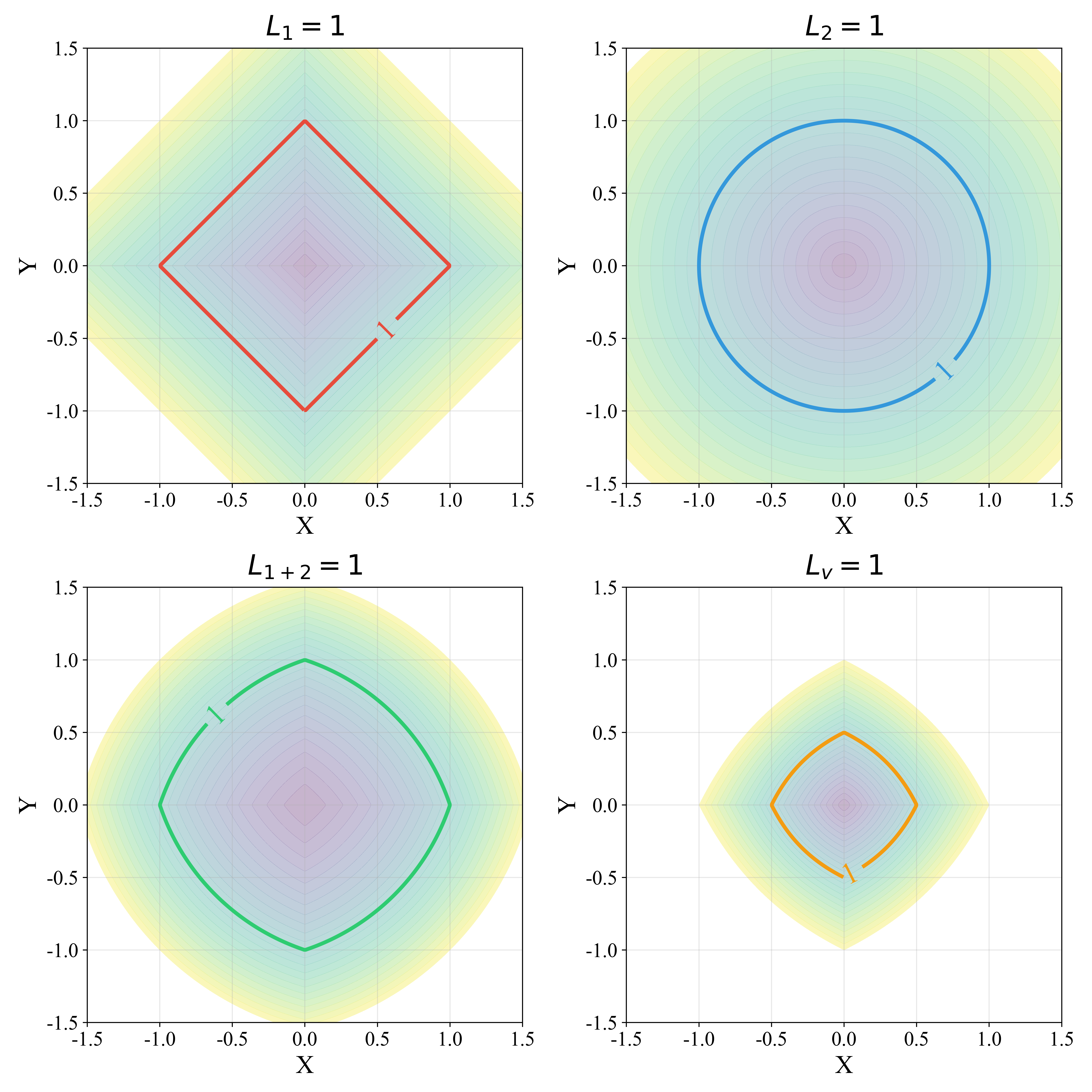}
 		\end{minipage}
 		\captionsetup{width=0.90\linewidth,font=footnotesize}
 		\caption{The Unit sphere of the View distance function in $3D$ space (left). The projections of $L_1(\boldsymbol{x})$, $L_2(\boldsymbol{x})$, $L_{1+2}(\boldsymbol{x})$ ($\lambda=0.5$), and $L_v(\boldsymbol{x})$ onto the XY plane (right).}
 		\label{UnitSphere}
 	\end{figure}
 	
 	Theorem~\ref{L1_norm_equivalent} and Theorem~\ref{L2_norm_equivalent} respectively establish that the ratio of the $L_v$-norm to the $L_1$-norm lies in $[(n-1)/\sqrt{2},\, n-1]$ and that the ratio of the $L_v$-norm to the $L_2$-norm is bounded by $[(n-1)/\sqrt{2},\, (n-1)\sqrt{n}]$. The unit sphere of the View distance has a centrally symmetric ellipsoidal form with anisotropic characteristics, lying between the hypersphere of Euclidean distance and the hypercube of Manhattan distance. Figure~\ref{UnitSphere} illustrates the $2D$ cross-sections of the $L_1(\boldsymbol{x})$, $L_2(\boldsymbol{x})$, $L_{1+2}(\boldsymbol{x})$ ($\lambda=0.5$), and View distance $L_v(\boldsymbol{x})$ metrics projected onto the XY plane.
 	
 	Compared to the circular boundary of the Euclidean distance, the cross-section of the View distance protrudes along the coordinate axes. In contrast to the diamond-shaped boundary of the Manhattan distance, it presents smoother edges, bearing a striking resemblance to the Elastic Net cross-section overall. Notably, the isolines for View distance are more tightly spaced than those for the other three metrics. The boundary of the coordinate axes within its unit constraint region ($L_v(x) \le 1$) is confined to $[-0.5, 0.5]$, which is half of the interval length of the boundaries ($[-1, 1]$) defined by the Euclidean, Manhattan, and Elastic Net. This compact unit area and contracted boundary range also indicate that the View distance changes faster, making it more sensitive to small variations.
 	
 	From a regularization perspective, View distance imposes multiple overlapping constraints by summing the $L_2$ constraints across all parameter pairs. Consequently, the View distance can effectively compress the range of parameter changes and apply stronger shrinkage on the parameters, which helps limit the model's complexity and reduce overfitting during training.
 	 		   		
   	\subsection{Hessian matrix}\label{sec5_2}
   	This section aims to derive the analytical expression for the Hessian matrix $\mathbf{H}(\mathbf{x}) = \nabla^{2} L_{v}(\boldsymbol{x})$ of the View distance function, and to analyze the local curvature information contained within this matrix.

   	For any pair $(i,j)$ with $1 \le i < j \le n$ , define
   	\begin{equation*}
   		L_{ij}(\boldsymbol{x}) = \sqrt{x_i^2 + x_j^2}.
   	\end{equation*}
	
	Fix $i$ and $j$ $(1 \leq i < j \leq n)$, the gradient of $L_{ij}$ is given by
	\begin{equation*} 
		\frac{\partial L_{ij}}{\partial x_i} = \dfrac{x_i}{L_{ij}}, \quad \frac{\partial L_{ij}}{\partial x_j} = \dfrac{x_j}{L_{ij}}, \quad \frac{\partial L_{ij}}{\partial x_k} = 0, k \neq i,j.
	\end{equation*}
	
	Differentiating again yields the second-order partial derivatives
	\begin{equation*}
		\begin{gathered}
			\frac{\partial^2 L_{ij}}{\partial x_i^2} =  \frac{x_j^2}{L_{ij}^{3}}, \quad \frac{\partial^2 L_{ij}}{\partial x_j^2} = \frac{x_i^2}{L_{ij}^{3}},\\
			\frac{\partial^2 L_{ij}}{\partial x_i \partial x_j} = \frac{\partial^2 L_{ij}}{\partial x_j \partial x_i} = -\frac{x_i x_j}{L_{ij}^{3}}, \\
			\frac{\partial^2 L_{ij}}{\partial x_k \partial x_l} = 0, (k, l) \notin \{i,j\} \times \{i,j\}.
		\end{gathered}
	\end{equation*}
		
   	Note that $L_{ij}$ is non-differentiable only when both $x_i = 0$ and $x_j = 0$. 
   		
   	Summing over all $i,j$ with $1 \leq i < j \leq n$, the second-order partial derivatives of $L_v$ is
   	\begin{equation*}
   		\nabla^2 L_v(\mathbf{x}) = \sum_{1 \le i < j \le n}^{n} \nabla^2 L_{ij}(\mathbf{x}).
    \end{equation*}

	 We obtain the Hessian $H(L_v) = (h_{kl})_{n \times n} $ with entries  		
	\begin{equation}\label{Hessian} 	
		h_{k\ell} =
		\begin{cases}
			\displaystyle \sum_{\substack{j=1 \\ j \ne k}}^{n} \frac{x_j^2}{(x_k^2 + x_j^2)^{3/2}}, & k = \ell, \\
			\displaystyle -\frac{x_k x_\ell}{(x_k^2 + x_\ell^2)^{3/2}}, & k \ne \ell.
		\end{cases}
	\end{equation}

    Since each $L_{ij}$ is a convex function, and the sum of convex functions is convex, $L_{v}$ is convex on its domain. Consequently, $ \mathbf{H}(L_{v}) $ is positive semidefinite.
   	
   	The degree of curvature of the Viewing distance varies in different directions, confirming its anisotropy. When a component $x_{k}$ is large, the denominator $(x_{k}^{2} + x_{j}^{2})^{3/2}$ becomes larger, causing $h_{kl}$ to decrease, which means the surface near the coordinate axis is relatively flat. When $x_{k}$ approaches $x_j$, $h_{kl}$ becomes relatively large, meaning the surface is more curved along the diagonal. Specifically, when $x_{k}=0$, the diagonal curvature
	\begin{equation*} 	
        h_{kk} = \displaystyle \sum_{\substack{j=1 \\ j \ne k}}^{n} \frac{x_{j}^{2}}{(0 + x_{j}^{2})^{3/2}} = \displaystyle \sum_{\substack{j=1 \\ j \ne k}}^{n} \frac{1}{|x_{j}|}, 
	\end{equation*}
    the curvature of this dimension itself is determined only by other dimensions.
    
    \noindent \textbf{Remark 5.} To enable stable numerical computation of the Hessian matrix for gradient-dependent downstream algorithms, a smoothing factor is introduced to the original projection distance.
    
    We define the smoothed $2D$ projection distance as
    \begin{equation*}
        L_{ij}(\boldsymbol{x}) = \sqrt{x_i^2 + x_j^2 + \epsilon^2},
    \end{equation*}
    where $\epsilon > 0$ denotes a tiny positive constant (a typical choice is $10^{-8}$). This smoothed distance function is infinitely differentiable on the domain. By computing its second-order partial derivatives, we obtain the entries of the Hessian matrix corresponding to the smoothed distance
    
    \begin{equation}\label{Hessian_smooth} 	
    	h_{k\ell} =
    	\begin{cases}
    		\displaystyle \sum_{\substack{j=1 \\ j \ne k}}^{n} \frac{x_j^2 + \epsilon^2}{(x_k^2 + x_j^2 + \epsilon^2)^{3/2}}, & k = \ell, \\
    		\displaystyle -\frac{x_k x_j}{(x_k^2 + x_j^2 + \epsilon^2)^{3/2}}, & k \ne \ell.
    	\end{cases}
    \end{equation}
    
    With the smoothing term $\epsilon$ incorporated, the denominator has a universal lower bound $\epsilon^{3}$, which eliminates the numerical singularity (division by zero) emerging in Equation~\ref{Hessian} when $x_i \to 0$ and $x_j \to 0$.
    
    \section{$2D$ projection plane selection strategy}\label{sec6}  
    According to the formulation of the View distance, the computational complexity of evaluating the complete set of projection planes scales quadratically, i.e., $\mathcal{O}(n^2)$. When the feature dimension $n$ is sufficiently large, exhaustive calculation not only incurs prohibitive computational costs but also tends to yield extremely large numerical values. This can compromise numerical stability and adversely affect subsequent algorithmic operations. Furthermore, it is important to note that not all $2D$ projection planes contribute equally to the model. Certain planes may be redundant or dominated by noise, which can potentially degrade the overall performance. To address these challenges, this section investigates effective strategies for selecting a subset of $2D$ projection planes. The primary objective is to reduce computational complexity while eliminating redundant or noisy planes, thereby preserving or even enhancing model performance. Consequently, we propose two distinct approaches: random $2D$ projection plane selection and $2D$ projection plane selection based on iterative Maximum Weight Matching.
    	       
    \subsection{Random $2D$ projection plane selection}\label{sec6_1}
	To simplify the calculation of the View distance, we initially employ a random selection strategy to reduce computational complexity. Specifically, $k$ distinct $2D$ projection planes are randomly selected from the total $n(n-1)/2$ candidate planes. This approach reduces the computational complexity of distance calculation from quadratic scale $\mathcal{O}(n^2)$ to linear scale $\mathcal{O}(k)$, thereby significantly improving computational efficiency and avoiding the generation of excessively large numerical values. Furthermore, by investigating how model performance varies with $k$, we can obtain an unbiased baseline estimate of the model's potential without relying on prior domain knowledge.
	
	However, this random selection strategy exhibits inherent limitations. Since the selection process is purely random, it fails to account for the discriminative ability of individual projection planes. Consequently, the selected subset may inadvertently include redundant or noisy planes while overlooking highly valuable planes that are critical for distance discrimination. This lack of directionality implies that a larger $k$ is often required to achieve satisfactory performance, resulting in suboptimal efficiency. To overcome these drawbacks and further enhance model accuracy with fewer projections, it is essential to move beyond random selection towards a more principled criterion. This prompted us to develop a selection mechanism based on the importance of the $2D$ projection plane, which we explain in detail in the next section.
			     
    \begin{center}
	    \begin{minipage}{0.7\linewidth}	    
		    \begin{algorithm}[H]
		    	\footnotesize
		    	\captionsetup{font=footnotesize}
		    	\caption{Random $2D$ projection plane selection}
		    	\begin{algorithmic}[1]
		    		\REQUIRE Data space $X \in \mathbb{R}^{m \times n}$, number of planes to select $k$.
		    		\STATE Define the set of all possible $2D$ projection planes $S = \{(i, j) \mid 1 \le i < j \le n\}$.	    		
		    		\STATE Randomly sample $k$ distinct planes from $S$ to form the subset $S_p$.	    		
		    		\STATE For any vectors $\boldsymbol{x}, \boldsymbol{y} \in {X}$, compute the distance based on the selected planes:
		    		\begin{equation*}
		    			d(\boldsymbol{x}, \boldsymbol{y}) = \sum_{(i,j) \in S_p} \sqrt{(x_i - y_i)^2 + (x_j - y_j)^2}.
		    		\end{equation*}
		    		\RETURN The distance metric $d(\boldsymbol{x}, \boldsymbol{y})$.
		    	\end{algorithmic}
		    \end{algorithm}
	    \end{minipage}
    \end{center}
            
	\subsection{$2D$ projection plane selection based on iterative Maximum Weight Matching}\label{sec6_2}
    Building upon the random $2D$ projection plane selection strategy, we propose a quantitative criterion to evaluate the importance of each plane.
 
    On one hand, if features $i$ and $j$ are constant, their contribution to the View distance calculation is null. This indicates that the feature pair $(i,j)$ carries no effective information, rendering the corresponding $2D$ projection plane invalid. On the other hand, a larger variance in features $i$ and $j$ generally implies greater distances between data points on the corresponding projection plane, thereby yielding more discriminative information. Specifically, we define the variance of the $L_{2}$-norm on the projection plane $(i,j)$, denoted as $D(\eta_{ij})$, as the metric for its importance. This variance is derived from the second-order raw moments and the expectation of the $L_{2}$-norm. According to the formulation in Section~\ref{sec2_2}, the expression for the variance is given by
	    
	\begin{equation}\label{variance}
	    \begin{gathered}
	  	    D(\eta_{ij}) = E(\eta_{ij}^{2}) - \left( E\left(\eta_{ij} \right) \right)^{2} \\ 
	    	= E\left( \xi_{i}^{2} \right) + E\left( \xi_{j}^{2} \right) - \left( E\left( \sqrt{\xi_i^2 + \xi_j^2} \right) \right)^{2}. \\
	    \end{gathered}
	\end{equation}  
		
	Based on Equation~\ref{variance}, we propose an iterative mutually exclusive $2D$ projection plane selection method based on the variance $D(\eta_{ij})$. The core idea of this method is to prioritize selecting $2D$ projection planes that exhibit the greatest data dispersion and mutually exclusive feature pairs in each iteration. The selection method can be modeled equivalently as an iterative Maximum Weight Matching (abbreviated as MWM) problem. The complete modeling process is explained as follows.
	
	We construct a weighted undirected complete graph $G=(V,E,W)$.
	\begin{itemize}
		\item Vertex set $V$. Each vertex corresponds to a feature in the dataset, and the total number of vertices satisfies $|V|=n$.
		\item Edge set $E$. An undirected edge is assigned to each feature pair $(i,j)$, which represents a $2D$ feature projection plane. The size of the initial full edge set is $|E|=n(n-1)/2$.
		\item Edge weight mapping $W$. For any edge $e_{ij}\in E$, its weight is defined as $w_{ij}=D(\eta_{ij})$, where $D(\eta_{ij})$ denotes a predefined scoring function. The weight value quantifies the discriminative importance of the corresponding projection pair.
	\end{itemize}
	
	In graph theory, an edge subset $M\subseteq E$ is defined as a matching if no two edges inside $M$ share a common vertex. The optimization objective for every single iteration is to solve a matching $M\subseteq E$ on the current residual subgraph $G$, such that the number of edges in $M$ does not exceed $n/2$, and the total weight $\sum_{e\in M} w_e$ of all edges in $M$ is maximized. This constraint ensures that projection pairs selected in one round adopt distinct features without repetition, which guarantees feature diversity.
	
	After solving the matching for a single iteration, all edges contained in the matching $M$ are removed to generate a residual subgraph
	\begin{equation*}
		G \leftarrow (V,\ E \setminus M,\ W \setminus \{w_{e} | e\in M\}).
	\end{equation*}
	
	We repeatedly compute the MWM on the updated residual subgraph, and terminate the iteration when the cumulative count of selected $2D$ projection pairs reaches the predefined target $k$. Notably, the first iteration is formulated as an MWM problem on a weighted undirected complete graph. Subsequent iterations involve computing the MWM on a weighted undirected graph. Particularly, when $k$ does not divide $n/2$, the final iteration degenerates into an MWM problem with a finite edge constraint on a weighted undirected graph. When $k=n/2$, all features are grouped pairwise, which corresponds to a special case of Group Lasso\cite{MingYua, NoahSimon, TTonyCai} with $\lambda=1$ and $K$ being an identity matrix.
			    
	\begin{center}
	   	\begin{minipage}{0.7\linewidth}	    	    
		    \begin{algorithm}[H]
		    	\centering
		    	\footnotesize
		    	\captionsetup{font=footnotesize}
		    	\caption{$2D$ projection plane selection based on iterative MWM}
		    	\label{MWM_selection}
		    	\begin{algorithmic}[1]
		    		\REQUIRE Data space $X \in \mathbb{R}^{m \times n}$, number of planes to select $k$.
		    			    		
		    		\STATE Define the set of all possible $2D$ projection planes $\mathcal{S} = \{(i, j) \mid 1 \le i < j \le n\}$. Initialize an empty list $\mathcal{L}$.
	
		    		\FOR{each pair $(i, j) \in \mathcal{S}$}
		    		\STATE Extract columns $X_i$ and $X_j$ from $X$.
		    		\STATE Compute $E(X_i^2)$ and $E(X_j^2)$ by
		    	        \begin{equation*}
		    			    E(X^2) = D(X) + (E(X))^2.
		    		    \end{equation*}
		    		\STATE Compute the squared expectation of the Euclidean norm:
		    	        \begin{equation*}
		    			    \left( E\left(\sqrt{X_i^2 + X_j^2}\right) \right)^2.
		    		    \end{equation*}
		    		\STATE Calculate the score $s_{ij} = D(\eta_{ij})$ using Equation~\ref{variance}.
		    		\STATE Append tuple $(i, j, s_{ij})$ to $\mathcal{L}$.
		    		\ENDFOR	
		    		\STATE Sort $\mathcal{L}$ in descending order based on the score $s_{ij}$.
		    		
		    		\STATE Initialize $\mathcal{S}_p$.
	 			    
		    		\WHILE{$|\mathcal{S}_p| < k$}
		    		\STATE Initialize a set of used features $\mathcal{U}$.
		    		\STATE Initialize an empty list for remaining candidates $\mathcal{I}$.
		    		\FOR{each index $p$ and tuple $(i, j, s_{ij})$ in the sorted $\mathcal{L}$}
		    		\IF{$i \notin \mathcal{U}$ \AND $j \notin \mathcal{U}$}
		    		\STATE Add $(i, j)$ to $\mathcal{S}_p$.
		    		\STATE Update $\mathcal{U} \leftarrow \mathcal{U} \cup \{i, j\}$.
	                \IF{$|\mathcal{S}_p| = k$}
		    		\STATE break
		    		\ELSE
		    		\STATE Append $p$ to $\mathcal{I}$.
		    		\ENDIF
		    		\ENDIF
		    		\ENDFOR
		    		\STATE Update candidate set $\mathcal{L} \leftarrow \mathcal{L}_{\mathcal{I}}$.
		    		\ENDWHILE	    		
		    		\STATE For any $\boldsymbol{x}, \boldsymbol{y} \in \mathbb{R}^n$, compute the distance based on the selected planes:
		    		\begin{equation*}
		    			d(\boldsymbol{x}, \boldsymbol{y}) = \sum_{(i,j) \in \mathcal{S}_p} \sqrt{(x_i - y_i)^2 + (x_j - y_j)^2}.
		    		\end{equation*}
		    		\RETURN The distance metric $d(\boldsymbol{x}, \boldsymbol{y})$.
		    	\end{algorithmic}
		    \end{algorithm}
	    \end{minipage}
    \end{center}
	
	Projection pairs selected within a single iteration share no common features. This mutually exclusive selection mechanism is designed to prevent potential bias: if a specific feature $i$ has exceptionally high variance, all its associated projection planes $(i, *)$ might be selected when $k$ is small. This would inadvertently exclude other equally effective projection planes. Projection pairs selected in different iterative rounds are allowed to reuse identical features, thereby capturing the joint distribution information among features. In each iteration, this matching strategy preferentially selects projection combinations with the highest discriminative scores and non-overlapping features from the remaining candidates. Consequently, the final selected set balances the discriminative importance of each projection pair and the diversity of feature dimensions. 
	
	We use the following greedy algorithm to implement iteration. First, the variance score $D(\eta_{ij})$ is computed for each feature pair $(i, j)$ using Equation~\ref{variance}, serving as the importance metric for the corresponding projection plane. Subsequently, all candidate planes are sorted in descending order by score. Finally, a non-replacement selection strategy is employed to iteratively get the top $n/2$ disjoint $2D$ projection planes with the highest scores from the candidate set, until a total of $k$ planes are obtained, thereby forming the optimal subset $\mathcal{S}_{p}$.
	
	This deterministic selection strategy effectively overcomes the inherent instability of random sampling. By explicitly filtering out planes with low variance and redundancy, we ensure that every selected projection plane maximizes the separability between samples and the diversity of features. Although pre-computing the variances for all candidate planes incurs a computational overhead, this is strictly a one-time preprocessing cost. In contrast, it prevents performance fluctuations caused by the introduction of noisy planes during subsequent model training, thereby achieving a superior trade-off between computational efficiency and model accuracy. The detailed procedure is summarized in Algorithm~\ref{MWM_selection}.
	
	To provide a comprehensive evaluation of computational efficiency, we analyze the time complexity of the proposed method in two phases: the offline precomputation phase and the online distance calculation phase.
	
	\begin{itemize}
		\item \textbf{Offline precomputation cost}. As noted in Algorithm~\ref{MWM_selection}, calculating the importance scores for all possible feature pairs requires iterating through the set $\mathcal{S}$, which contains $n(n-1)/2$ pairs. Given a dataset with $m$ samples, computing the statistical expectations for each pair takes $\mathcal{O}(m)$ time. Thus, the total complexity for scoring all pairs is $\mathcal{O}(mn^{2})$. Additionally, sorting these scores requires $\mathcal{O}(n^{2} \log{n})$ (Lines 2--9). During the iterative greedy selection, the outer while loop runs at most $k$ times. In each iteration, the inner loop scans remaining candidate planes, which takes $\mathcal{O}(n^{2})$ in the worst case. The worst‑case complexity of this step is $\mathcal{O}(k n^{2})$ (Line 10--26). The total preprocessing time complexity is
		\begin{equation*}
		    T_{\text{preprocess}} = \mathcal{O}(mn^{2} + n^{2}\log {n} + k n^{2}).
		\end{equation*}
		Since $\log n$ grows slowly with feature dimension $n$, the sample size $m$ is much larger than $\log n$ for most real‑world datasets. Therefore, $\mathcal{O}(mn^2)$ serves as the dominant term, which comes from expectation and variance evaluation over all samples for each feature pair. The sorting term $\mathcal{O}(n^2\log n)$ can dominate only under extreme conditions with very few samples. 
		\item \textbf{Online distance calculation}. Once the optimal subset of planes $\mathcal{S}_p$ (with size $k$) is selected, the distance calculation between any two samples $\boldsymbol{x}$ and $\boldsymbol{y}$ only involves summing the Euclidean distances on these $k$ projection planes. Consequently, the online complexity is reduced to $\mathcal{O}(k)$, which is significantly lower than the standard $\mathcal{O}(n^2)$ costs of full-projection metrics.
	\end{itemize}  
    		    	    	    
	\section{Experiments and analysis}\label{sec7}
	To verify the effectiveness of the proposed View distance metric, this paper constructs baseline models based on the DBSCAN clustering algorithm and the KNN classification algorithm. The performance of the $L_{p}$ and View distance metrics is systematically compared on clustering and classification tasks. Furthermore, to rigorously evaluate the practical impact of different projection plane selection strategies, we conduct a series of experiments using the KNN algorithm as the baseline model. Only partial experimental results are presented in the main text. The complete results are provided in the Appendix.
		
	\subsection{Datasets and experimental setup}
	\subsubsection{Data preparation}
	\noindent \textbf{Clustering datasets}. Two types of synthetic Gaussian datasets, including isotropic and anisotropic Gaussian clusters, are constructed for clustering evaluation. Each dataset contains 3 clusters with 1000 samples per cluster, and the cluster centers are randomly generated within the range of $[-10, 10]$. For the isotropic Gaussian dataset, the standard deviation ($(\sqrt{n}+1)/2$) increases adaptively with the feature dimension. For the anisotropic Gaussian dataset, symmetric semi-positive definite covariance matrices ($AA^{T}$) are constructed via random matrices ($A$) to generate irregular cluster structures. 
		
	Since DBSCAN relies on spatial distance and density distribution, it is highly susceptible to the curse of dimensionality. In high-dimensional spaces, the discriminability of pairwise distances decreases significantly, making the neighborhood radius $eps$ difficult to determine. Therefore, the feature dimensions are set to $\{3, 5, 7, 9\}$ to conduct comparative experiments from low to medium dimensions.
    
    \begin{figure}[H]
    	\centering
    	\begin{minipage}{0.4\linewidth}
    		\centering
    		\includegraphics[width=\linewidth]{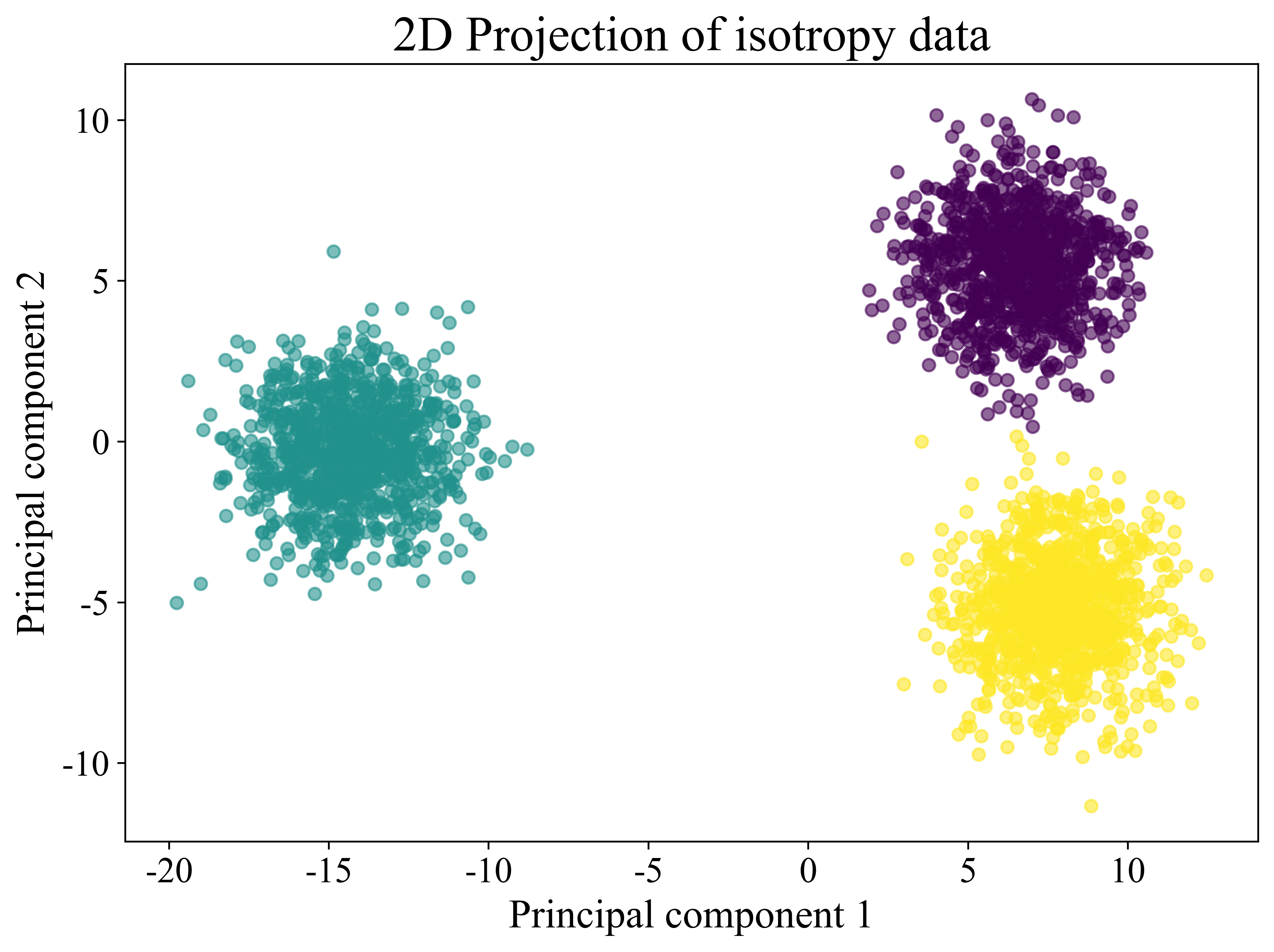}
    	\end{minipage}
    	\hspace{0.01\linewidth}
    	\begin{minipage}{0.4\linewidth}
    		\centering
    		\includegraphics[width=\linewidth]{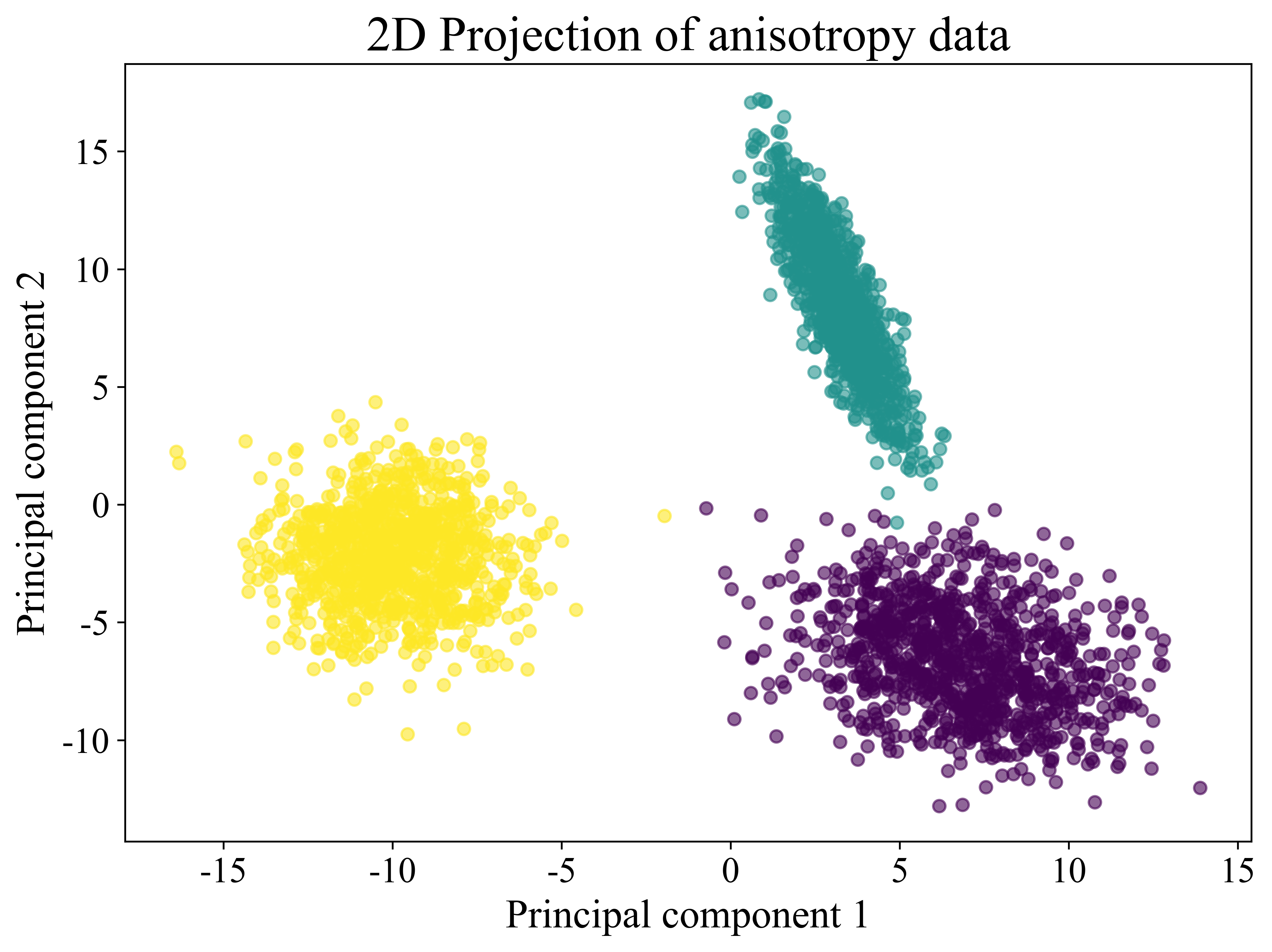}
    	\end{minipage}
    	\captionsetup{width=0.90\linewidth,font=footnotesize}
    	\caption{The $5D$ data is projected into a $2D$ plane via PCA for visualization. The isotropic dataset exhibits circular structures, while the anisotropic dataset shows elliptical structures.}
    \end{figure}
    
	\noindent \textbf{Classification datasets}. All datasets utilized in this study were sourced from the UCI Machine Learning Repository\cite{uci} and scikit-learn\cite{sklearn}. Specifically, we employed 12 representative datasets spanning diverse application domains: toxicity\cite{toxicity}, darwin\cite{darwin}, wine\cite{wine}, connectionist bench (sonar, mines vs. rocks)\cite{sonar}, seeds\cite{seeds}, heart disease\cite{heart}, ionosphere\cite{ionosphere}, musk (version 1)\cite{musk}, drug induced autoimmunity prediction\cite{drug}, breast cancer wisconsin (diagnostic)\cite{breast}, digits\cite{sklearn}, image segmentation\cite{image}. Table~\ref{classification_datasets} summarizes detailed statistics for these datasets, including the number of instances, features, and classes, as well as the distribution of instances across classes.
	
	\begin{table}[H]
		\centering
		\small
		\captionsetup{font={footnotesize}}
		\caption{Description of the classification datasets.}
		\label{classification_datasets}
		\begin{tabular}{ccccc}
			\hline
			Dataset & Instances & Features & Class & Distribution\\
			\hline
			toxicity & 171 & 1203 & 2 & 115/56 \\
			darwin & 174 & 451 & 2 & 89/85 \\
			wine & 178 & 13 & 3 & 71/59/48 \\
			sonar & 208 & 60 & 2 & 111/97 \\ 
			seeds & 210 & 7 & 3 & 70/70/70 \\
			heart & 297 & 13 & 2 & 160/137 \\
			ionosphere & 351 & 34 & 2 & 225/126\\
			musk & 476 & 168 & 2 & 269/207 \\
			drug & 477 & 196 & 2 & 359/118 \\
			wdbc & 569 & 30 & 2 & 357/212 \\
			digits & 1797 & 64 & 10 & 183/182/182/181/181/180/179/178/177/174 \\
			image & 2310 & 19 & 7 & 330/330/330/330/330/330/330 \\
			\hline
		\end{tabular}
	\end{table}
	
	To ensure feature scale consistency and eliminate dimensional discrepancies, all data instances were normalized using StandardScaler prior to model training.
	
	\subsubsection{Parameter settings}
	\noindent \textbf{Clustering parameters}. The DBSCAN minimum neighborhood size is fixed to 5. The neighborhood radius $eps$ is adaptively determined by taking the 0.9-quantile of all 5-th nearest-neighbor distances, ensuring objective and stable parameter selection.
	
    \noindent \textbf{Classification parameters}. To strictly control variables and ensure a fair comparison, the number of nearest neighbors for all KNN classifiers is uniformly set to the default value of $K=5$.

	\subsubsection{Experimental protocol}	
	To ensure the statistical robustness and reproducibility of the experimental results, we conducted comprehensive performance evaluations under a rigorous assessment protocol. Specifically, to mitigate potential biases introduced by random initialization, we configured 100 distinct random seeds and executed 100 independent experiments. Particularly, for classification tasks, a stratified 10-fold cross-validation strategy was employed for data partitioning and model evaluation in each experiment.
		
	\subsubsection{Evaluation metrics}
	\noindent \textbf{Clustering metrics}. The Adjusted Rand Index (ARI) and Silhouette Score (SIL) are adopted as core evaluation metrics. Both the ARI and Silhouette Score range from $-1$ to $1$, where values closer to $1$ indicate better clustering performance. The Silhouette Score should be computed using the corresponding distance metric, rather than uniformly adopting the Euclidean distance. We compute the mean and standard deviation across 100 repeated experiments and count invalid clustering cases (i.e., a single cluster or full noise distribution) to comprehensively evaluate the clustering performance and robustness of different distance metrics.
	
	\noindent \textbf{Classification metrics}. Classification accuracy and Macro-F1 score are used to evaluate classification performance. We report the mean and standard deviation across 100 cross-validation trials to quantitatively compare the accuracy and stability of different distance metrics.
    	
    \subsection{Importance analysis of $2D$ projection planes}
\begin{figure}[H]
    	\centering
    	\includegraphics[width=0.60\textwidth]{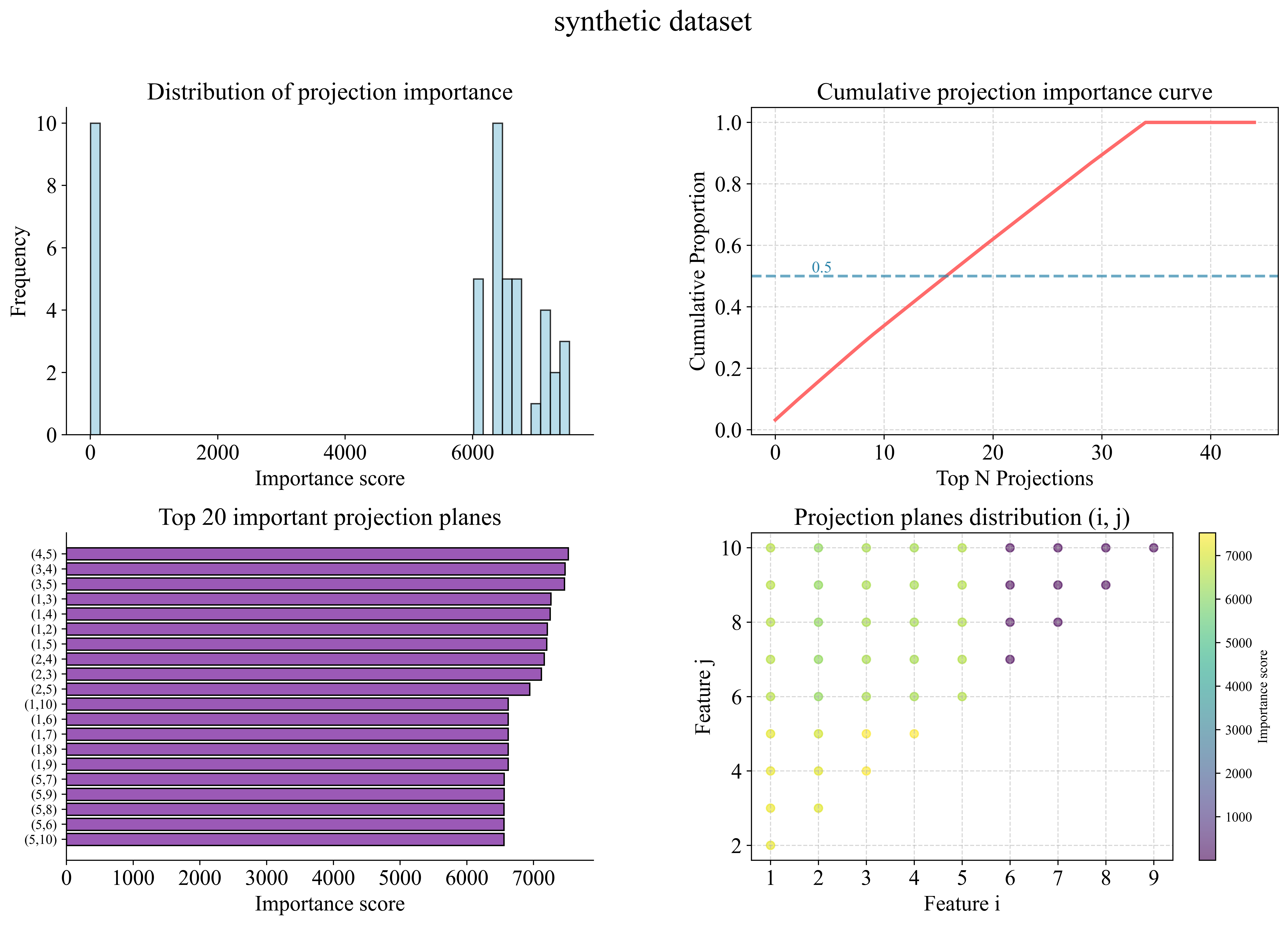}
    	\captionsetup{width=0.9\textwidth, font={footnotesize}}
    	\caption{Distribution of importance scores for $2D$ projection planes for the synthetic dataset: histogram, cumulative trend, top-20 ranked planes, and their spatial distribution.}
    	\label{scoresSynthetic}
    \end{figure}

\begin{figure}[H]
    	\centering
    	\begin{minipage}{0.45\linewidth}
    		\centerline{\includegraphics[width=\linewidth]{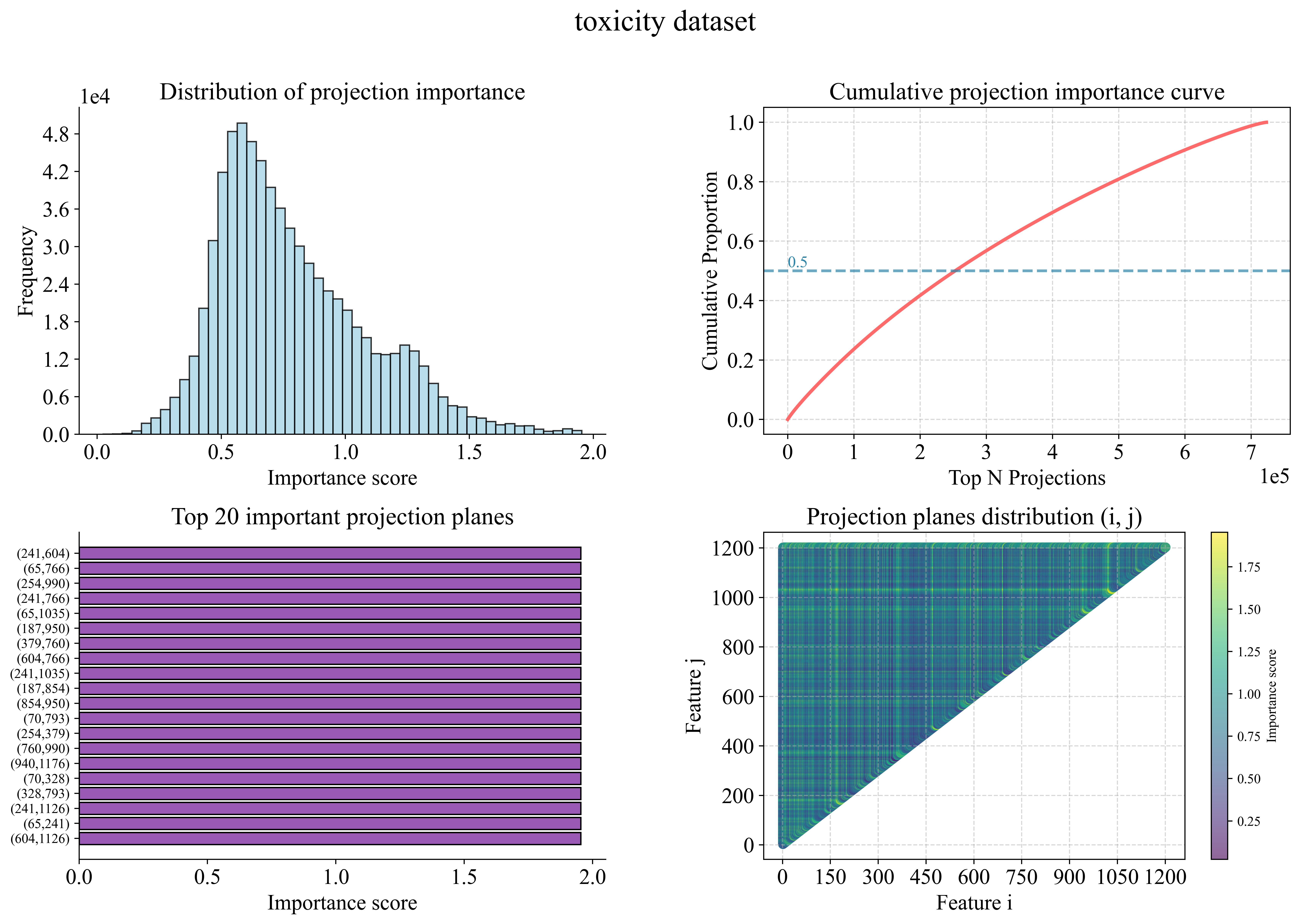}}
    	\end{minipage}
    	\vspace{3pt}
    	\begin{minipage}{0.45\linewidth}
    		\centerline{\includegraphics[width=\linewidth]{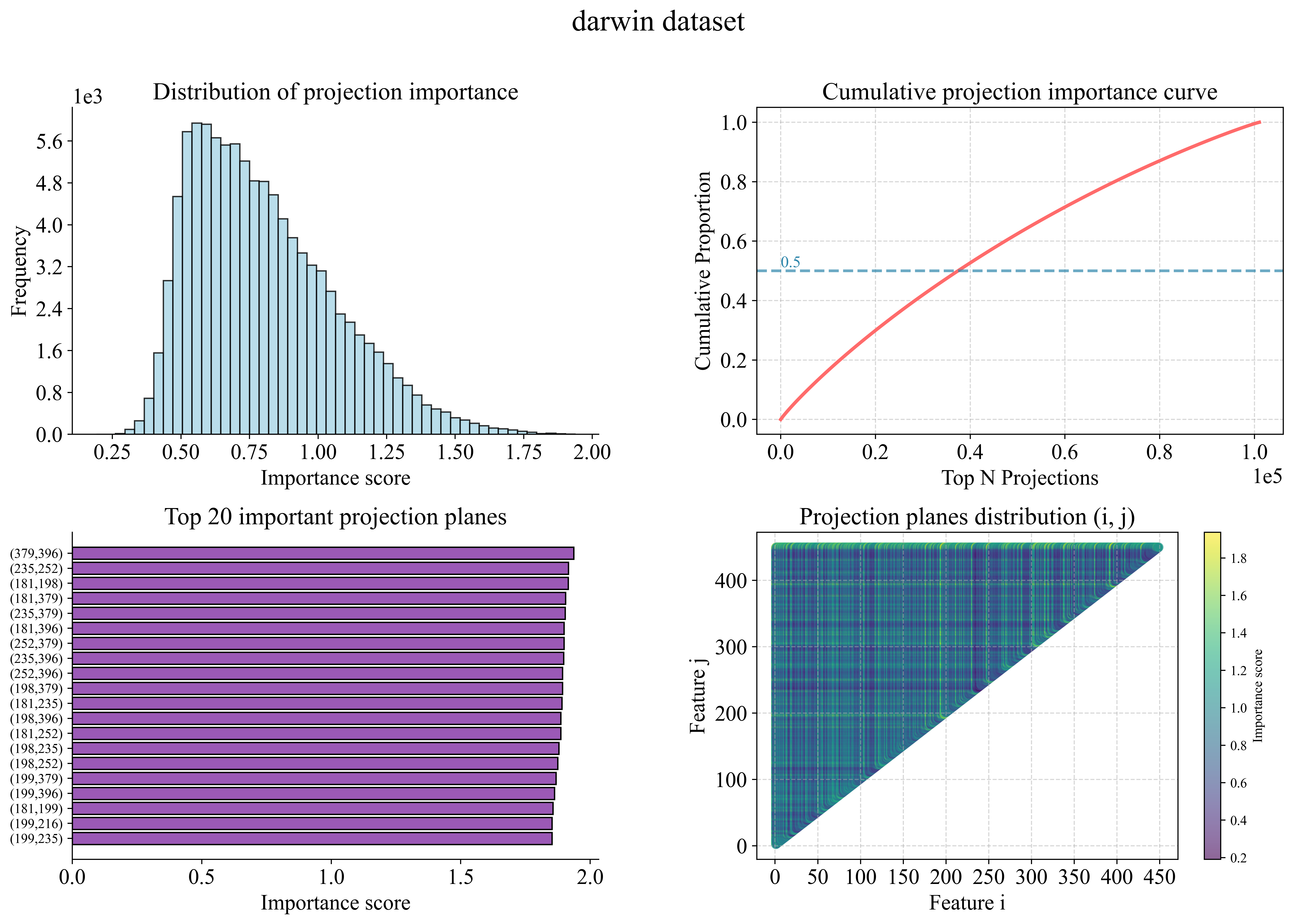}}
    	\end{minipage}
    	\captionsetup{width=0.9\textwidth, font={footnotesize}}
    	\caption{Distribution of importance scores for $2D$ projection planes for the toxicity and darwin datasets: histogram, cumulative trend, top-20 ranked planes, and their spatial distribution.}
    	\label{scoresHigh}
    \end{figure}

\begin{figure}[H]
    	\centering
    	\begin{minipage}{0.45\linewidth}
    		\centerline{\includegraphics[width=\linewidth]{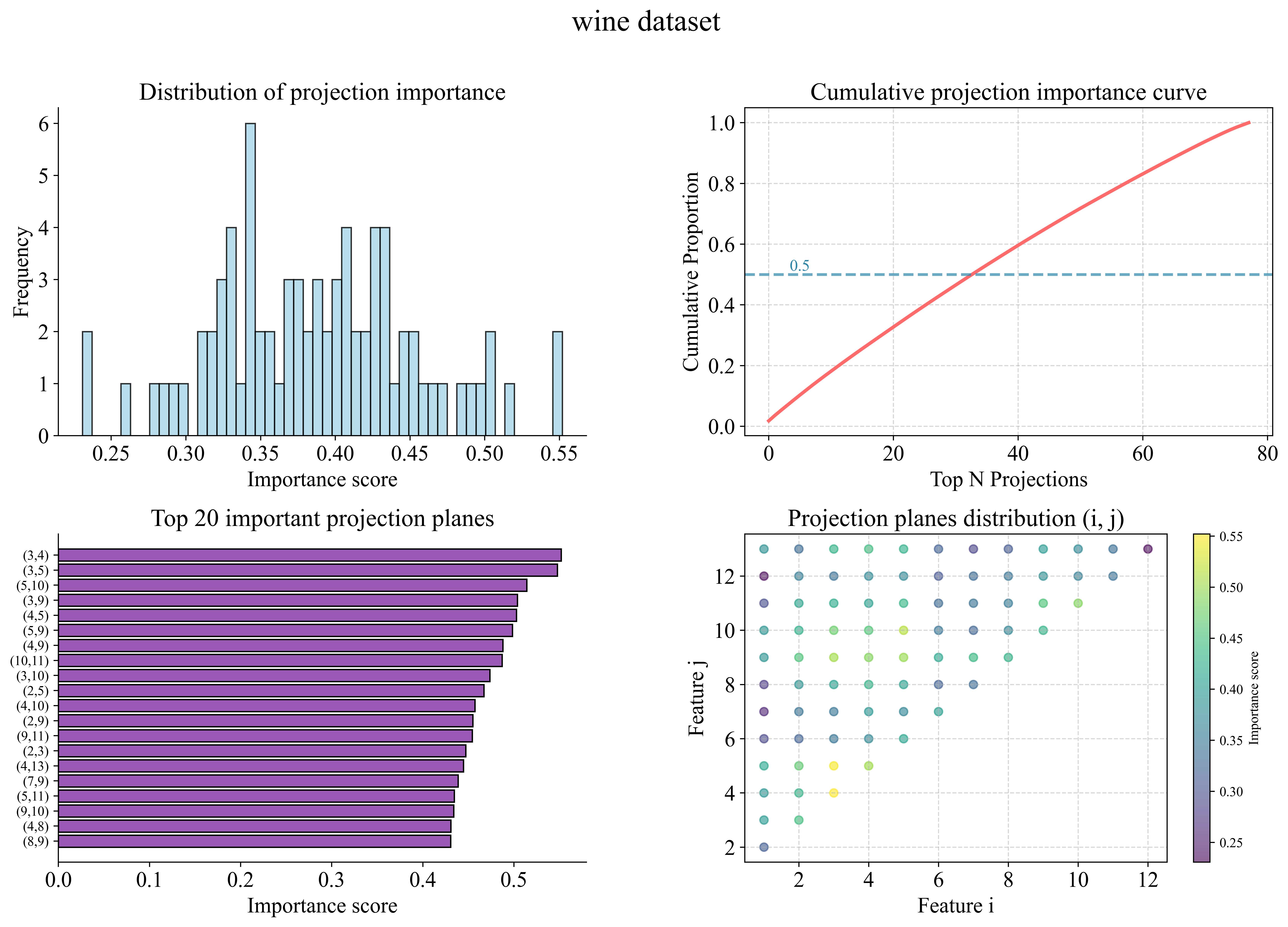}}
    	\end{minipage}
    	\vspace{3pt}
    	\begin{minipage}{0.45\linewidth}
    		\centerline{\includegraphics[width=\linewidth]{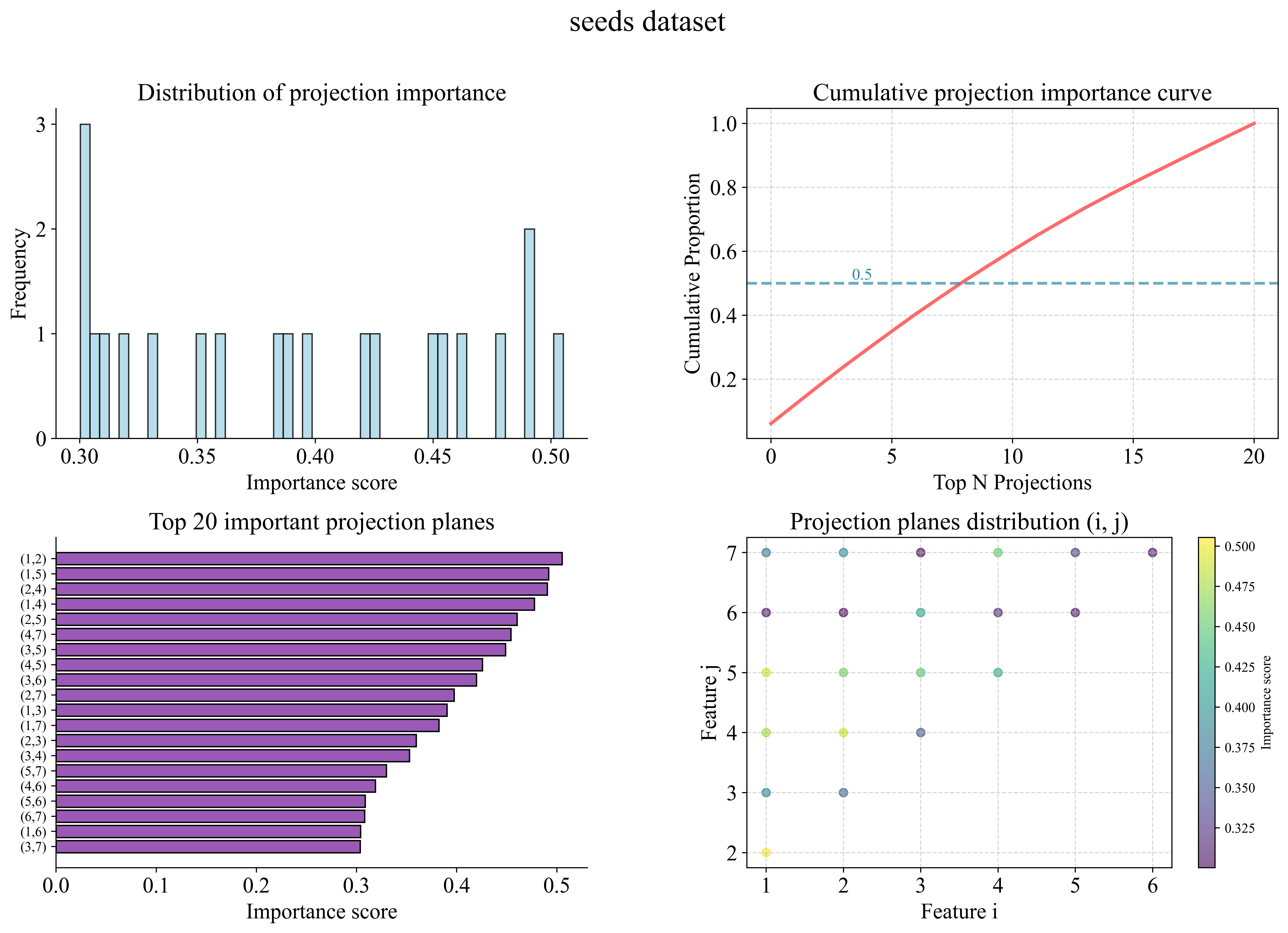}}
    	\end{minipage}
    	\captionsetup{width=0.9\textwidth, font={footnotesize}}
    	\caption{Distribution of importance scores for $2D$ projection planes for the wine and seeds datasets: histogram, cumulative trend, top-20 ranked planes, and their spatial distribution.}
    	\label{scoresLow}
    \end{figure}

\begin{figure}[H]
    	\centering
    	\vspace{3pt}
    	\begin{minipage}{0.45\linewidth}
    		\centerline{\includegraphics[width=\linewidth]{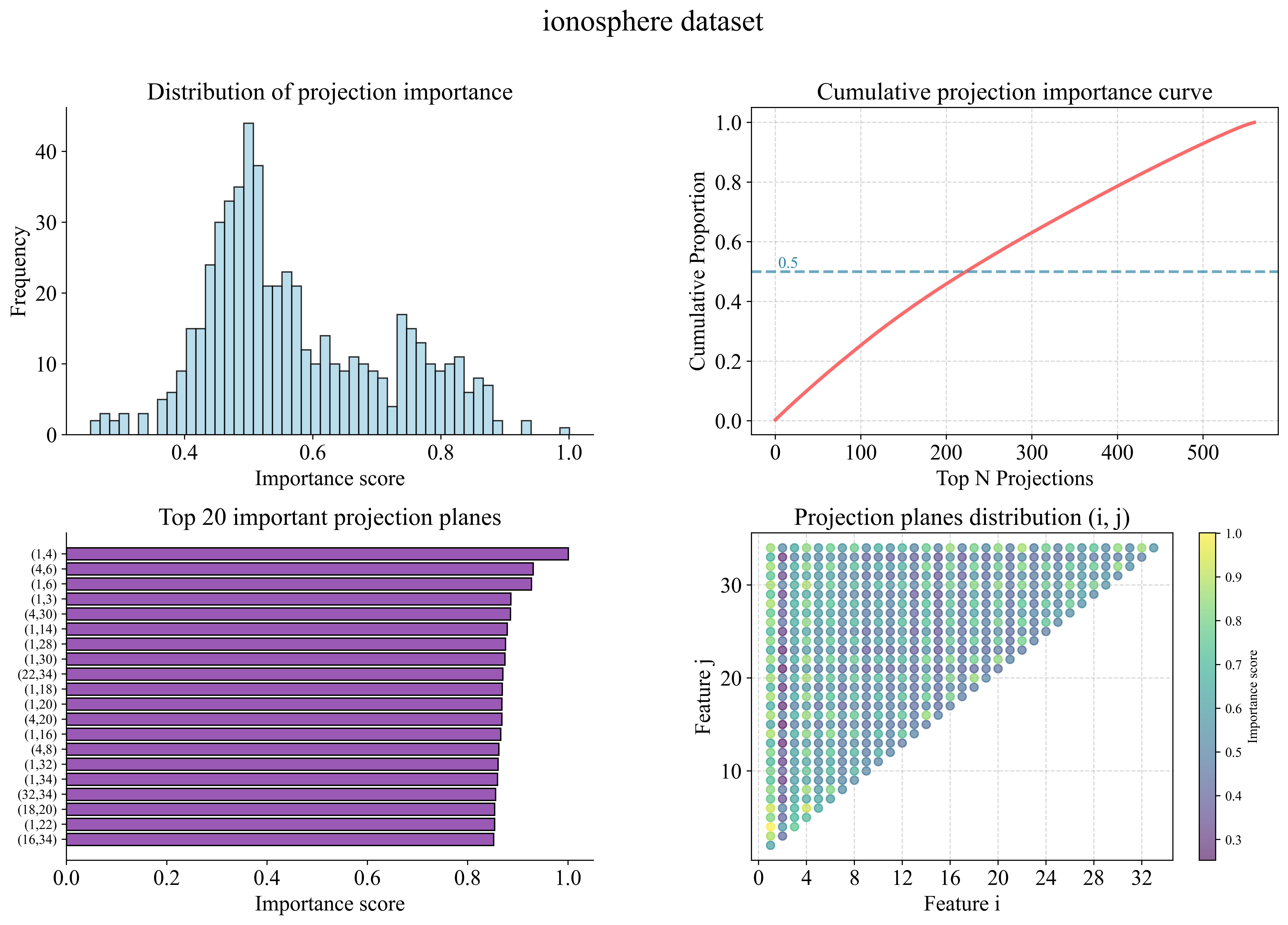}}
    	\end{minipage}
    	\vspace{3pt}
    	\begin{minipage}{0.45\linewidth}
    		\centerline{\includegraphics[width=\linewidth]{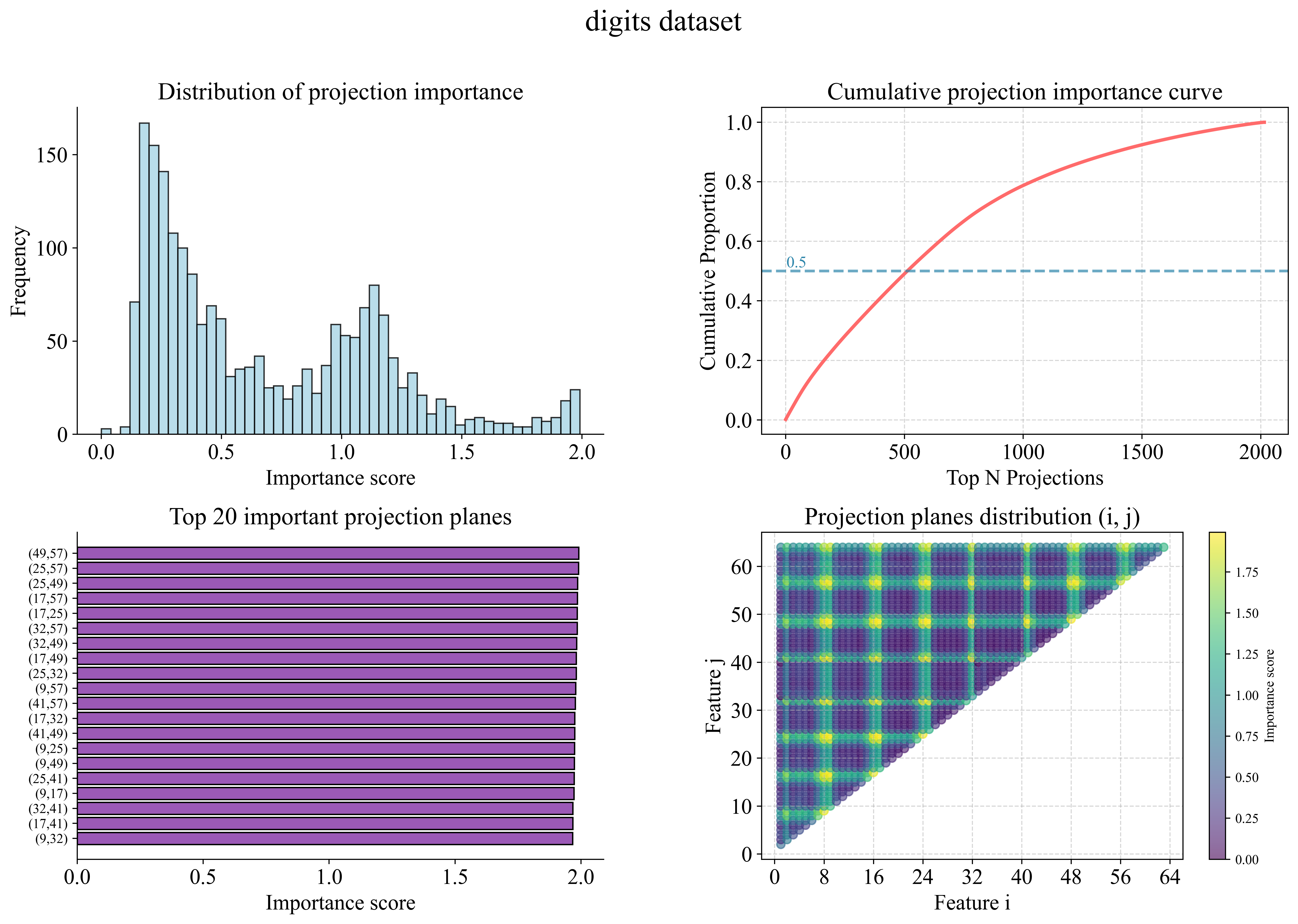}}
    	\end{minipage}
        \captionsetup{width=0.9\textwidth, font={footnotesize}}
        \caption{Distribution of importance scores for $2D$ projection planes for the ionosphere and digits datasets.}
        \label{scoresMiddel}
    \end{figure}

    Based on Equation~\ref{variance}, we computed the importance scores of $2D$ projection planes for each dataset. By analyzing the distribution of importance scores, cumulative trends, and the top-ranked planes by score, we aimed to investigate the general patterns governing the importance of $2D$ projection planes, thereby providing practical guidance for the plane selection strategy (as illustrated in Figures~\ref{scoresSynthetic}-\ref{scoresMiddel}.
    
    To analyze how magnitude discrepancies across features affect plane selection, we generate a $10D$ synthetic dataset. The first five features are configured with a mean of 100 and a variance of 100, whereas the latter five features follow a distribution with a mean of 1 and a variance of 1. As illustrated in Figure~\ref{scoresSynthetic}, the selected projection planes are completely dominated by the first five features with large magnitudes. In contrast, although the latter five features have relatively the same variance, their low magnitudes mean they contribute little to plane selection. Accordingly, normalization should be performed during the data preprocessing stage to ensure that all features have an equal opportunity to participate in projection‑plane selection.
    
    Experimental results on real‑world datasets indicate that the importance score distributions of $2D$ projection planes vary significantly across datasets, and most do not conform to uniform or normal distributions. However, a common characteristic across all datasets is a substantial gap between the highest and lowest importance scores, along with a cumulative curve that gradually plateaus. This suggests that only a subset of feature pairs contains critical information, while the remaining pairs are predominantly redundant or noisy. These findings underscore the necessity of selectively extracting $2D$ projection planes, which not only effectively reduce the computational complexity of distance calculation but also successfully filter out redundant and noisy planes.
    
    Furthermore, we observed a noteworthy phenomenon across all datasets: due to the high variance of certain features, their corresponding feature pairs tend to exhibit high, closely clustered importance scores. If the top $k$ $2D$ projection planes are selected solely based on these scores, the chosen planes will be highly concentrated on a few specific features (i.e., intersecting planes sharing the same features). If the selection is based solely on the importance of the projection planes, this approach is prone to selecting redundant planes while overlooking other feature pairs with slightly lower importance scores but significant information. Consequently, the final model would require more planes to achieve the desired performance level. To address this issue, we adopted a $2D$ plane selection strategy based on iterative MWM to select $k$ $2D$ projection planes. Although this strategy may still introduce some redundancy or noise, it largely prevents the selected planes from overfocusing on specific features. Consequently, the model can achieve better performance with a significantly smaller number of projection planes. The advantages of this deterministic selection strategy are clearly demonstrated in the subsequent performance comparisons.
               
    \subsection{Performance and stability analysis}
    \subsubsection{Clustering task}
    In both isotropic and anisotropic synthetic Gaussian data scenarios, the DBSCAN clustering model based on the View distance achieves competitive overall performance on the ARI and SIL metrics. Specifically, the View distance mostly yields optimal or suboptimal clustering results, outperforming traditional comparison distances such as the Chebyshev and higher-order Lp distances. This verifies the effectiveness and applicability of the proposed View distance for clustering tasks. Even if the View distance fails to achieve optimal performance under certain dimensionality, its clustering results remain consistently between Manhattan and Euclidean distances, without sharp performance degradation, demonstrating strong robustness.
        
    \begin{table}[H]
    	\centering
    	\small
    	\captionsetup{width=0.9\textwidth, font={footnotesize}}
    	\caption{Clustering performance comparison of different distance metrics under isotropic scenarios. Invalid indicates the number of invalid clustering results.}
    	\begin{tabular}{ccccccccccc}
    		\toprule
    		\multirow{2}{*}{Scenario} & \multirow{2}{*}{Dim} & \multirow{2}{*}{Metric} & \multirow{2}{*}{Stat} & \multicolumn{6}{c}{Distance Metric} \\
    		\cmidrule(lr){5-10}
    		& & & & $L_1$ & $L_2$ & $L_3$ & $L_4$ & $L_\infty$ & $L_v$ \\
    		\midrule
    		\multirow{20}{*}[-6ex]{Isotropy} & \multirow{9}{*}[5ex]{3} & \multirow{2}{*}{ARI} & mean & \textbf{0.7543} & 0.7474 & 0.7460 & 0.7422 & 0.7526 & \textbf{0.7550} \\
    		& & & std & \textbf{0.2375} & 0.2473 & 0.2578 & 0.2579 & 0.2477 & \textbf{0.2470} \\
    		\cmidrule(lr){3-10}
    		& & \multirow{2}{*}{SIL} & mean & \textbf{0.6260} & 0.5876 & 0.5916 & 0.5790 & 0.5716 & \textbf{0.5938} \\
    		& & & std & \textbf{0.1565} & 0.1633 & 0.1609 & 0.1585 & \textbf{0.1521} & 0.1615 \\
    		\cmidrule(lr){3-10}
    		& & Invalid & numbers & 2 & 3 & 4 & 4 & 3 & 2 \\
    		\cmidrule(lr){2-10}
    		& \multirow{9}{*}[5ex]{5} & \multirow{2}{*}{ARI} & mean & 0.8364 & \textbf{0.8399} & \textbf{0.8442} & 0.8364 & 0.8334 & 0.8365 \\
    		& & & std & 0.2268 & 0.2119 & \textbf{0.2104} & \textbf{0.2136} & 0.2153 & \textbf{0.2136} \\
    		\cmidrule(lr){3-10}
    		& & \multirow{2}{*}{SIL} & mean & \textbf{0.6293} & 0.5959 & 0.5792 & 0.5663 & 0.5358 & \textbf{0.6178} \\
    		& & & std & \textbf{0.088} & 0.1048 & 0.1002 & 0.1038 & 0.1081 & \textbf{0.0829} \\
    		\cmidrule(lr){3-10}
    		& & Invalid & numbers & 3 & 2 & 2 & 2 & 2 & 2 \\
    		\cmidrule(lr){2-10}
    		& \multirow{9}{*}[5ex]{7} & \multirow{2}{*}{ARI} & mean & 0.8568 & 0.8668 & \textbf{0.8788} & \textbf{0.8833} & 0.8642 & 0.8584 \\
    		& & & std & 0.1966 & 0.1889 & \textbf{0.181} & \textbf{0.1784} & 0.1918 & 0.2081 \\
    		\cmidrule(lr){3-10}
    		& & \multirow{2}{*}{SIL} & mean & \textbf{0.6013} & 0.5742 & 0.5474 & 0.5358 & 0.4948 & \textbf{0.5951} \\
    		& & & std & \textbf{0.0851} & 0.088 & 0.0924 & 0.0933 & 0.1013 & \textbf{0.0845} \\
    		\cmidrule(lr){3-10}
    		& & Invalid & numbers & 1 & 1 & 1 & 1 & 1 & 2 \\
    		\cmidrule(lr){2-10}
    		& \multirow{9}{*}[5ex]{9} & \multirow{2}{*}{ARI} & mean & 0.8826 & 0.884 & \textbf{0.8854} & 0.8819 & 0.878 & \textbf{0.8852} \\
    		& & & std & 0.1978 & 0.1944 & \textbf{0.1791} & \textbf{0.182} & 0.2 & 0.1948 \\
    		\cmidrule(lr){3-10}
    		& & \multirow{2}{*}{SIL} & mean & \textbf{0.5798} & 0.5408 & 0.5085 & 0.4884 & 0.4389 & \textbf{0.5669} \\
    		& & & std & \textbf{0.0573} & 0.0628 & 0.0698 & 0.0747 & 0.0817 & \textbf{0.06} \\
    		\cmidrule(lr){3-10}
    		& & Invalid & numbers & 2 & 2 & 1 & 1 & 2 & 2 \\
    		\bottomrule
    	\end{tabular}
    \end{table}
    
    In Section~\ref{sec5}, we prove that the View distance is metrically equivalent to the Manhattan and Euclidean distances, and its geometric shape lies between these two classical distance metrics. This theoretical conclusion reasonably explains the experimental phenomena described above. The View distance balances the respective advantages of Manhattan and Euclidean distance metrics, thus maintaining stable overall performance between the two classical distances and achieving adaptability to clustering data with different distributions. Overall, it provides a novel, more general alternative distance metric for clustering tasks involving data with unknown distributions.
    
    \begin{table}[H]
    	\centering
    	\small
    	\captionsetup{width=0.9\textwidth, font={footnotesize}}
    	\caption{Clustering performance comparison of different distance metrics under anisotropic scenarios.}
    	\begin{tabular}{ccccccccccc}
    		\toprule
    		\multirow{2}{*}{Scenario} & \multirow{2}{*}{Dim} & \multirow{2}{*}{Metric} & \multirow{2}{*}{Stat} & \multicolumn{6}{c}{Distance Metric} \\
    		\cmidrule(lr){5-10}
    		& & & & $L_1$ & $L_2$ & $L_3$ & $L_4$ & $L_\infty$ & $L_v$ \\
    		\midrule
    		\multirow{20}{*}[-6ex]{Anisotropy} & \multirow{9}{*}[5ex]{3} & \multirow{2}{*}{ARI} & mean & \textbf{0.7745} & 0.7692 & 0.7737 & 0.7742 & \textbf{0.7752} & 0.7697 \\
    		& & & std & \textbf{0.2291} & \textbf{0.2296} & 0.2299 & 0.2299 & 0.2303 & 0.2299 \\
    		\cmidrule(lr){3-10}
    		& & \multirow{2}{*}{SIL} & mean & \textbf{0.4415} & 0.4221 & 0.4254 & 0.4269 & 0.4368 & \textbf{0.4388} \\
    		& & & std & 0.2001 & 0.1657 & \textbf{0.1611} & \textbf{0.1616} & 0.1694 & 0.1876 \\
    		\cmidrule(lr){3-10}
    		& & Invalid & numbers & 0 & 0 & 0 & 1 & 0 & 0 \\
    		\cmidrule(lr){2-10}
    		& \multirow{9}{*}[5ex]{5} & \multirow{2}{*}{ARI} & mean & 0.8193 & 0.8198 & 0.8199 & \textbf{0.8201} & 0.8125 & \textbf{0.8240} \\
    		& & & std & 0.2373 & \textbf{0.2251} & 0.2252 & \textbf{0.2251} & 0.2398 & \textbf{0.2238} \\
    		\cmidrule(lr){3-10}
    		& & \multirow{2}{*}{SIL} & mean & \textbf{0.5277} & 0.4925 & 0.4814 & 0.4721 & 0.4683 & \textbf{0.5083} \\
    		& & & std & \textbf{0.1176} & 0.1404 & 0.1293 & 0.1346 & \textbf{0.1156} & 0.1216 \\
    		\cmidrule(lr){3-10}
    		& & Invalid & numbers & 2 & 2 & 2 & 2 & 2 & 1 \\
    		\cmidrule(lr){2-10}
    		& \multirow{9}{*}[5ex]{7} & \multirow{2}{*}{ARI} & mean & 0.7596 & \textbf{0.7925} & 0.7806 & 0.7687 & 0.7468 & \textbf{0.7847} \\
    		& & & std & 0.2805 & \textbf{0.2675} & 0.2690 & 0.2701 & 0.2900 & \textbf{0.2686} \\
    		\cmidrule(lr){3-10}
    		& & \multirow{2}{*}{SIL} & mean & \textbf{0.4888} & 0.4625 & 0.4386 & 0.4237 & 0.3964 & \textbf{0.4801} \\
    		& & & std & \textbf{0.0845} & 0.0907 & 0.0926 & 0.0909 & 0.0884 & \textbf{0.0807} \\
    		\cmidrule(lr){3-10}
    		& & Invalid & numbers & 6 & 6 & 5 & 5 & 8 & 6 \\
    		\cmidrule(lr){2-10}
    		& \multirow{9}{*}[5ex]{9} & \multirow{2}{*}{ARI} & mean & 0.7092 & \textbf{0.7624} & \textbf{0.7365} & 0.7151 & 0.6751 & 0.7220 \\
    		& & & std & 0.3304 & \textbf{0.2994} & \textbf{0.3075} & 0.3142 & 0.3318 & 0.3238 \\
    		\cmidrule(lr){3-10}
    		& & \multirow{2}{*}{SIL} & mean & \textbf{0.4676} & 0.4330 & 0.4073 & 0.3907 & 0.3486 & \textbf{0.4545} \\
    		& & & std & \textbf{0.0732} & 0.0795 & 0.0850 & 0.0877 & 0.0932 & \textbf{0.0771} \\
    		\cmidrule(lr){3-10}
    		& & Invalid & numbers & 13 & 9 & 10 & 11 & 14 & 12 \\
    		\bottomrule
    	\end{tabular}
    \end{table}
     
    \subsubsection{Classification task}
	As a distance-dependent classifier, KNN is adopted as the baseline model in this subsection. This setup enables intuitive quantitative comparison among the $L_{p}$ distance, the View distance, and the two proposed $2D$ projection plane selection strategies. We repeated the classification experiments under 100 different random seeds, and recorded the mean and standard deviation of classification accuracy and Macro-F1 score for performance evaluation. In addition, we performed a two-sample t-test on the KNN results based on the View distance and the Euclidean distance. The corresponding t-statistics and $p$-values were calculated to assess whether View distance yields a statistically significant performance improvement. The baseline comparison results are summarized in Table~\ref{full_projecion}.
	
	Experimental results show that the KNN algorithm based on View distance generally achieves optimal or sub-optimal performance across all 12 datasets. Compared to Euclidean distance, KNN based on View distance achieved higher average accuracy and Macro-F1 score on all datasets except the seeds and musk. All corresponding $p$-values are less than 0.05, and $t$-statistics exceed 2 except for darwin, which demonstrates statistically significant performance gains at the $95\%$ confidence level. Meanwhile, the KNN based on View distance generally yields lower standard deviations of evaluation metrics, proving that View distance can also enhance the stability of classification results. These results verify the practical application value of the View distance metric. Compared with other $L_p$ distance metrics, Manhattan distance achieves better performance than View distance on the sonar, ionosphere, and image datasets, with performance margins of $0.54\%$, $0.9\%$, and $0.16\%$, respectively. $L_{4}$ distance achieves better performance than View distance on the darwin dataset, with performance margins of $1.99\%$.
	
	Subsequently, we explored the performance of the random $2D$ projection selection strategy with four preset parameters $k=\{n/4, n/2, 3n/4, n\}$. As shown in Table~\ref{random_result}, excluding the ionosphere dataset, the classification performance and stability improve monotonically as $k$ increases. Specifically, the mean values of accuracy and Macro-F1 score rise while the standard deviations decline. In terms of performance convergence, selecting $n/2$ random projection planes retains over $93\%$ of the performance of using all projection planes, and selecting $n$ random planes retains more than $95\%$ of the performance (exceeding $99\%$ on most datasets). For the ionosphere dataset, the classification performance with $k=n/4$ outperforms that of the full-projection plane scheme.
	
	\begin{table}[H]
		\centering
		\small
		\captionsetup{width=0.9\textwidth, font={footnotesize}}
		\caption{Experimental results in terms of mean accuracy, mean Macro-F1 score, and average standard deviation are reported based on 100 repeated KNN evaluations using $L_p$ and View distances across 6 datasets. A two-sample $t$-test is further performed to statistically compare the performance of Euclidean distance and View distance.}
		\label{full_projecion}
		\setlength{\tabcolsep}{5pt}
		\renewcommand{\arraystretch}{1.3}
		\begin{tabular}{cccccccc|ccc}
			\toprule
			Dataset & Metric & Stat & $\text{KNN}_{L_{1}}$ & $\text{KNN}_{L_{2}}$ & $\text{KNN}_{L_{3}}$ & $\text{KNN}_{L_{4}}$ & $\text{KNN}_{L_{\infty}}$ & $\text{KNN}_{L_{v}}$ & p-value & t-stat\\
			\midrule
			\multirow{4}{*}{toxicity}
			& accuracy & mean & 0.6303 & 0.6098 & 0.6137 & 0.5693 & 0.5413 & \textbf{0.6331} & *** & 11.43 \\
			&          & std  & 0.0949 & \textbf{0.0940} & \textbf{0.0940} & 0.0969 & 0.1063 & 0.0949 & & \\
			& macro-f1 & mean & 0.5338 & 0.5109 & 0.5158 & 0.4708 & 0.4855 & \textbf{0.5376} & *** & 10.93 \\
			&          & std  & 0.1151 & 0.1137 & 0.1124 & \textbf{0.1076} & 0.1108 & 0.1154 & & \\
			\hline
			\multirow{4}{*}{darwin}
			& accuracy & mean & 0.6975 & 0.6903 & 0.7171 & \textbf{0.7185} & 0.6517 & 0.6986 & 0.1276 & 7.076 \\
			&          & std  & \textbf{0.0802} & 0.0877 & 0.0908 & 0.0953 & 0.1041 & 0.0805 & & \\
			& macro-f1 & mean & 0.6643 & 0.6633 & 0.7028 & \textbf{0.7100} & 0.6452 & 0.6656 & *** & 1.536 \\
			&          & std  & 0.1044 & 0.1066 & \textbf{0.1016} & \textbf{0.1016} & 0.1061 & 0.1045 & & \\
			\hline
			\multirow{4}{*}{wine}
			& accuracy & mean & 0.9644 & 0.9664 & 0.9543 & 0.9499 & 0.9208 & \textbf{0.9687} & 0.001 & 3.414 \\
			&          & std  & 0.0403 & 0.0394 & 0.0455 & 0.0469 & 0.0587 & \textbf{0.0384} & & \\
			& macro-f1 & mean & 0.9652 & 0.9674 & 0.9552 & 0.9509 & 0.9214 & \textbf{0.9693} & 0.0036 & 2.983 \\
			&          & std  & 0.0396 & 0.0385 & 0.0451 & 0.0465 & 0.0592 & \textbf{0.0378} & & \\
			\hline
			\multirow{4}{*}{seeds}
			& accuracy & mean & 0.9250 & \textbf{0.9309} & 0.9238 & 0.9224 & 0.9200 & 0.9264 & $***$ & -7.081 \\
			&          & std  & 0.0524 & \textbf{0.0509} & 0.0554 & 0.0559 & 0.0549 & 0.0516 & & \\
			& macro-f1 & mean & 0.9238 & \textbf{0.9301} & 0.9231 & 0.9217 & 0.9192 & 0.9254 & $***$ & -7.201 \\
			&          & std  & 0.0538 & \textbf{0.0518} & 0.0561 & 0.0565 & 0.0557 & 0.0527 & & \\
			\hline
			\multirow{4}{*}{ionosphere}
			& accuracy & mean & \textbf{0.8868} & 0.8496 & 0.8252 & 0.8288 & 0.8495 & 0.8778 & $***$ & 45.76 \\
			&          & std  & \textbf{0.0463} & 0.0492 & 0.0496 & 0.0494 & 0.0483 & 0.0473 & & \\
			& macro-f1 & mean & \textbf{0.8670} & 0.8170 & 0.7806 & 0.7855 & 0.8154 & 0.8554 & $***$ & 45.87 \\
			&          & std  & \textbf{0.0578} & 0.0659 & 0.0715 & 0.0709 & 0.0661 & 0.0599 & & \\
			\hline
			\multirow{4}{*}{digits}
			& accuracy & mean & 0.9770 & 0.9773 & 0.9724 & 0.9667 & 0.9361 & \textbf{0.9785} & $***$ & 6.694 \\
			&          & std  & 0.0102 & \textbf{0.0100} & 0.0111 & 0.0120 & 0.0170 & \textbf{0.0100} & & \\
			& macro-f1 & mean & 0.9768 & 0.9772 & 0.9723 & 0.9676 & 0.9356 & \textbf{0.9784} & $***$ & 6.403 \\
			&          & std  & 0.0104 & \textbf{0.0101} & 0.0112 & 0.0122 & 0.0172 & \textbf{0.0101} & & \\
			\bottomrule
		\end{tabular}
		
		\begin{tablenotes}  
		\footnotesize   
		\item[a] $*** \textit{P} < 0.001$.
		\end{tablenotes}
	\end{table}
	
\begin{figure}[H]
		\centering
		\begin{minipage}{0.45\linewidth}
			\centerline{\includegraphics[width=\linewidth]{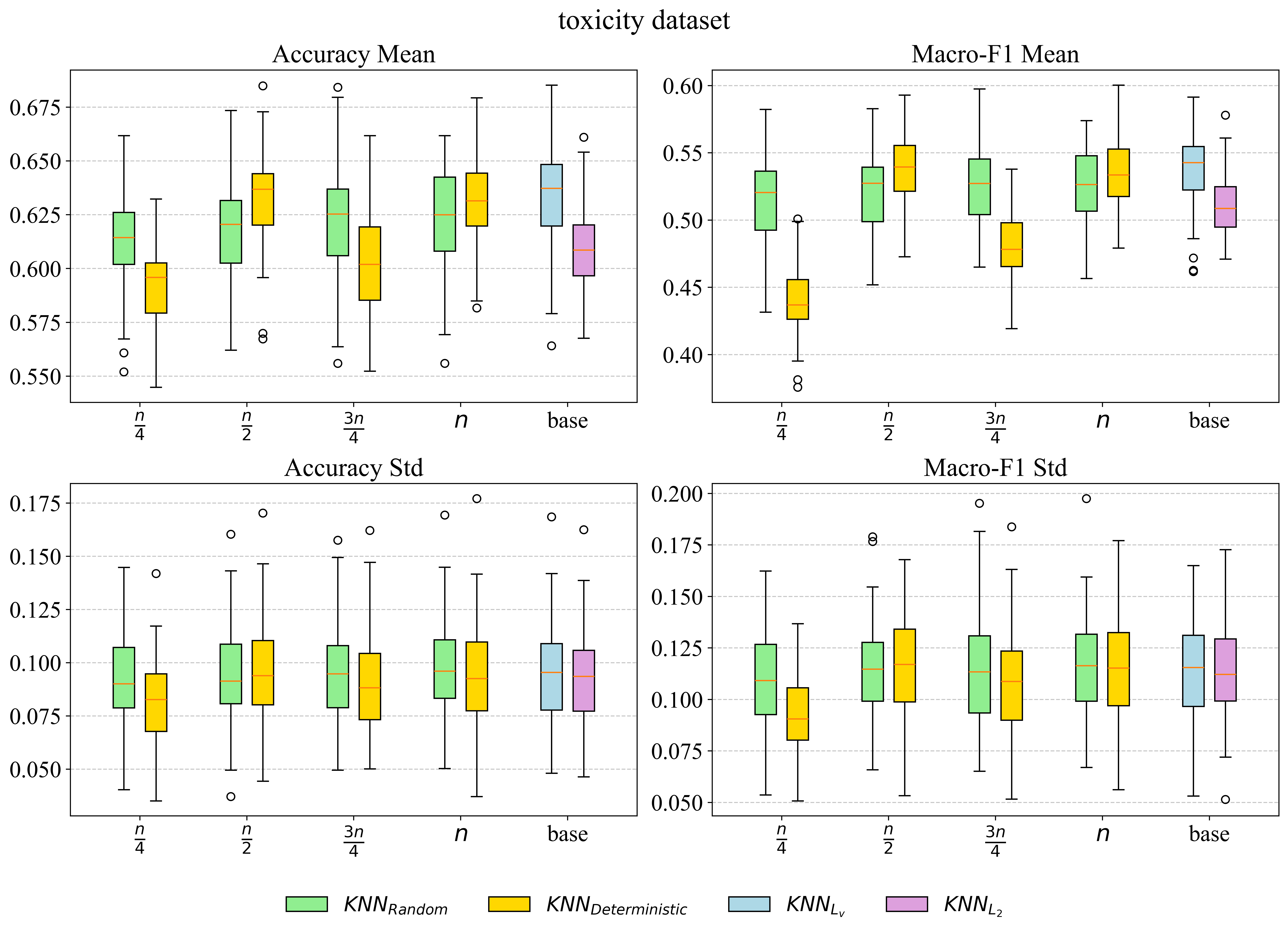}}
		\end{minipage}
		\vspace{3pt}
		\begin{minipage}{0.45\linewidth}
			\centerline{\includegraphics[width=\linewidth]{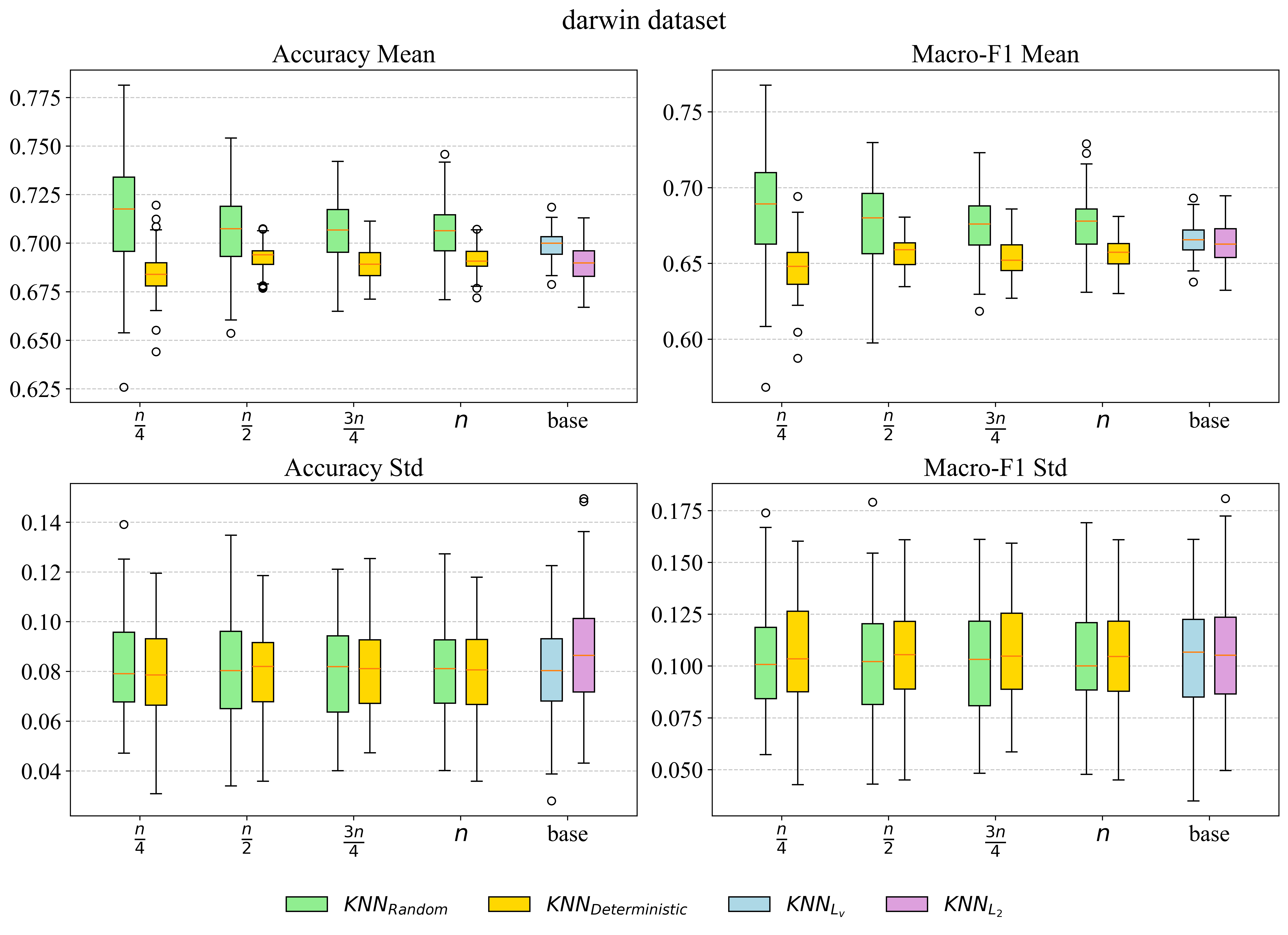}}
		\end{minipage}
		\vspace{3pt}
		\captionsetup{width=0.9\textwidth, font={footnotesize}}
		\caption{Box plots of KNN algorithm performance based on random and deterministic selection strategies for the toxicity and darwin datasets, with View and Euclidean distances as baselines.}
		\label{toxicity_boxplot}
	\end{figure}

\begin{figure}[H]
        \centering
    	\begin{minipage}{0.45\linewidth}
    		\centerline{\includegraphics[width=\linewidth]{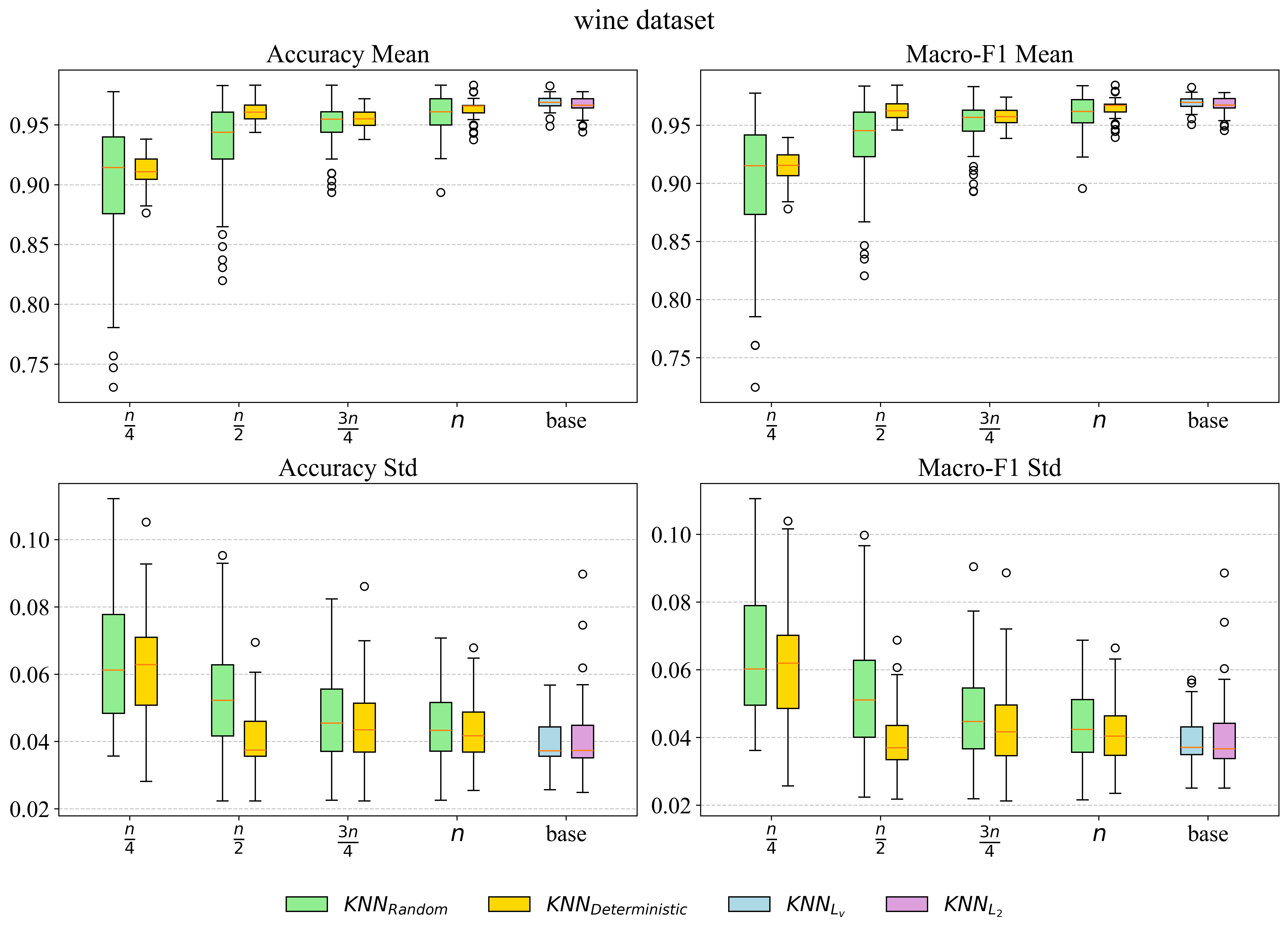}}
    	\end{minipage}
    	\vspace{3pt}
    	\begin{minipage}{0.45\linewidth}
    		\centerline{\includegraphics[width=\linewidth]{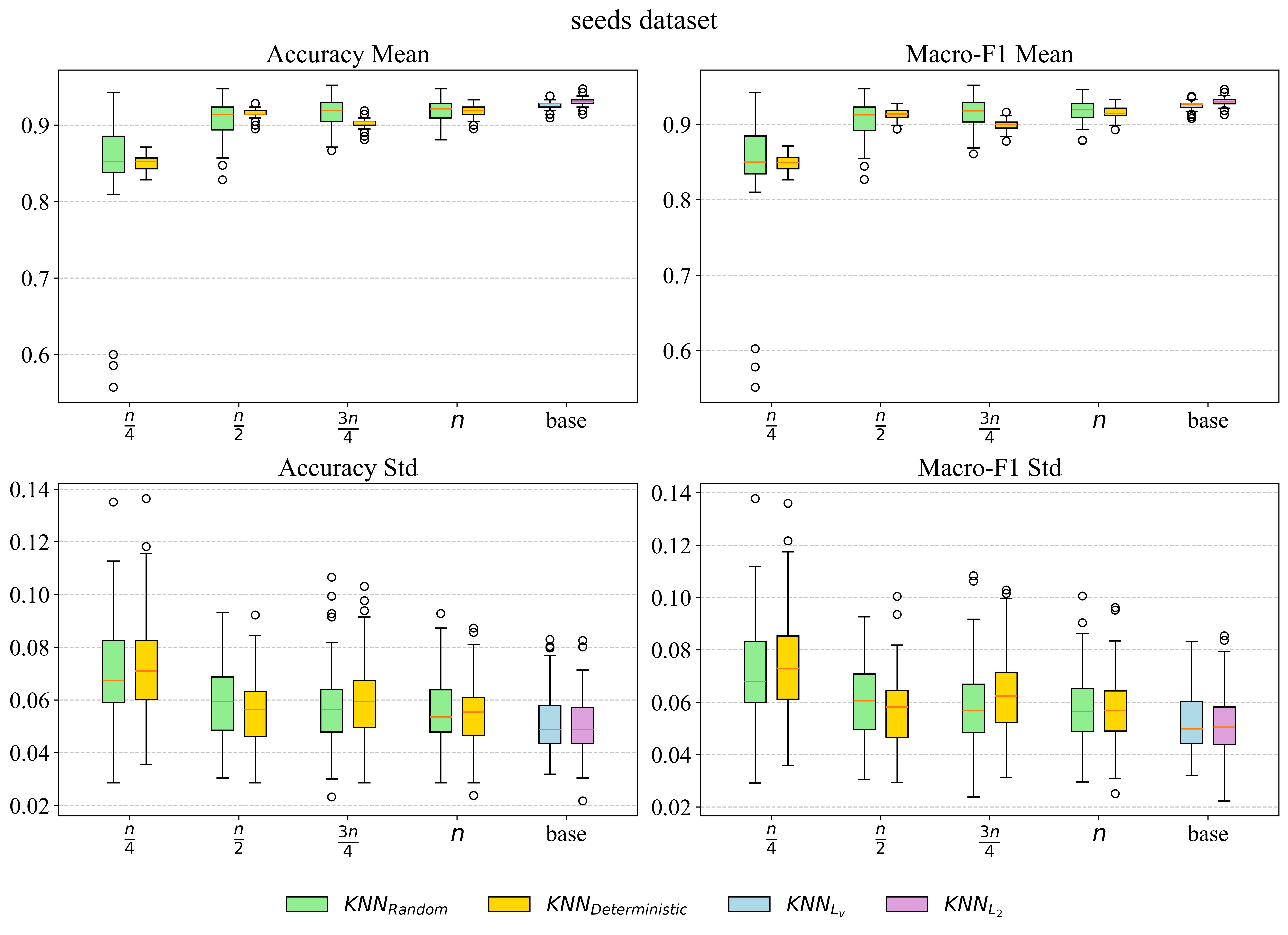}}
    	\end{minipage}
    	\vspace{3pt}
    	\captionsetup{width=0.9\textwidth, font={footnotesize}}
    	\caption{Box plots of KNN algorithm performance based on random and deterministic selection strategies for the wine and seeds datasets, with View and Euclidean distances as baselines.}
    	\label{wine_boxplot}
    \end{figure}

\begin{figure}[H]
    	\centering
    	\begin{minipage}{0.45\linewidth}
    		\centerline{\includegraphics[width=\linewidth]{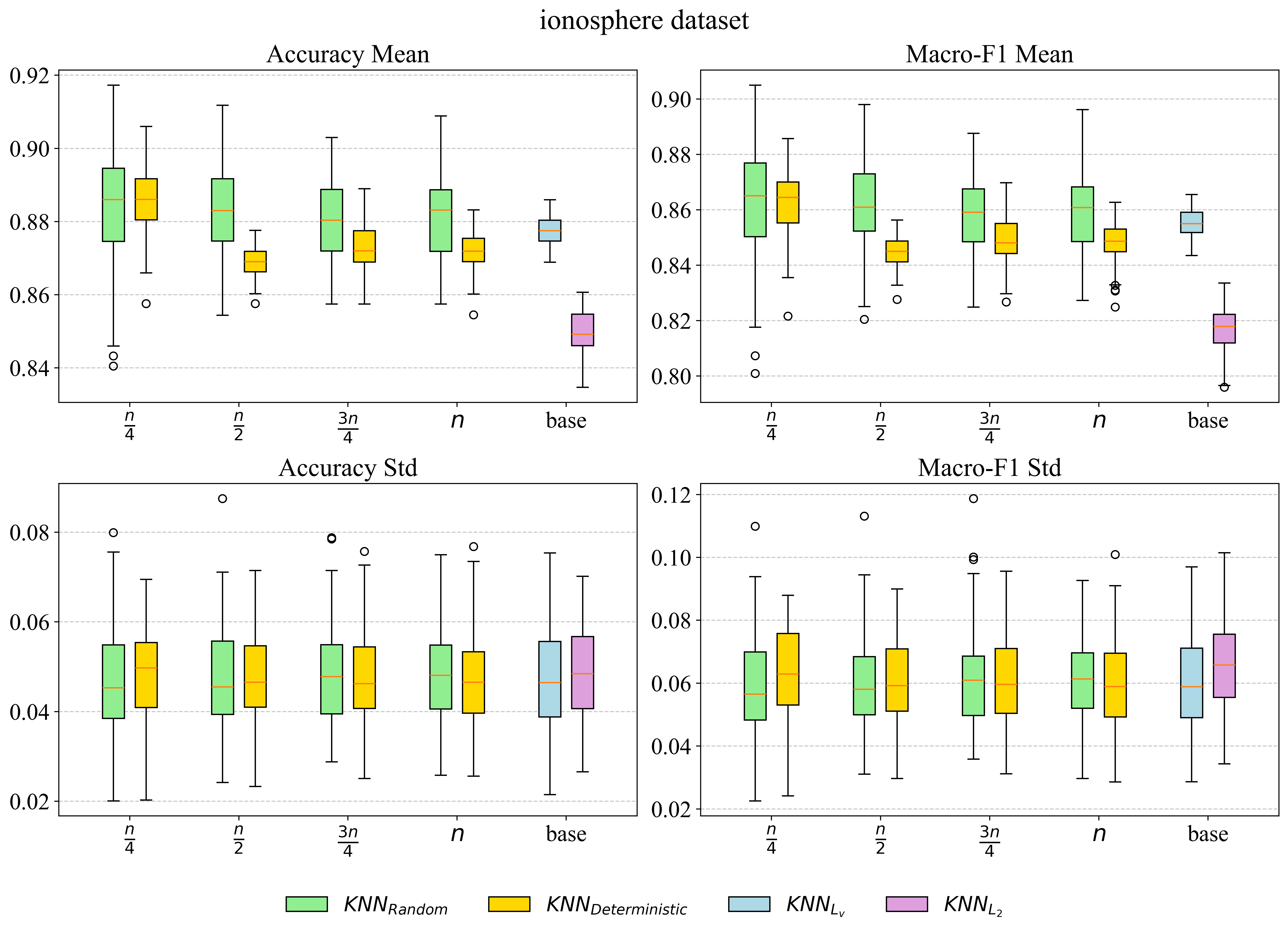}}
    	\end{minipage}
    	\vspace{3pt}
 	    \begin{minipage}{0.45\linewidth}
    		\centerline{\includegraphics[width=\linewidth]{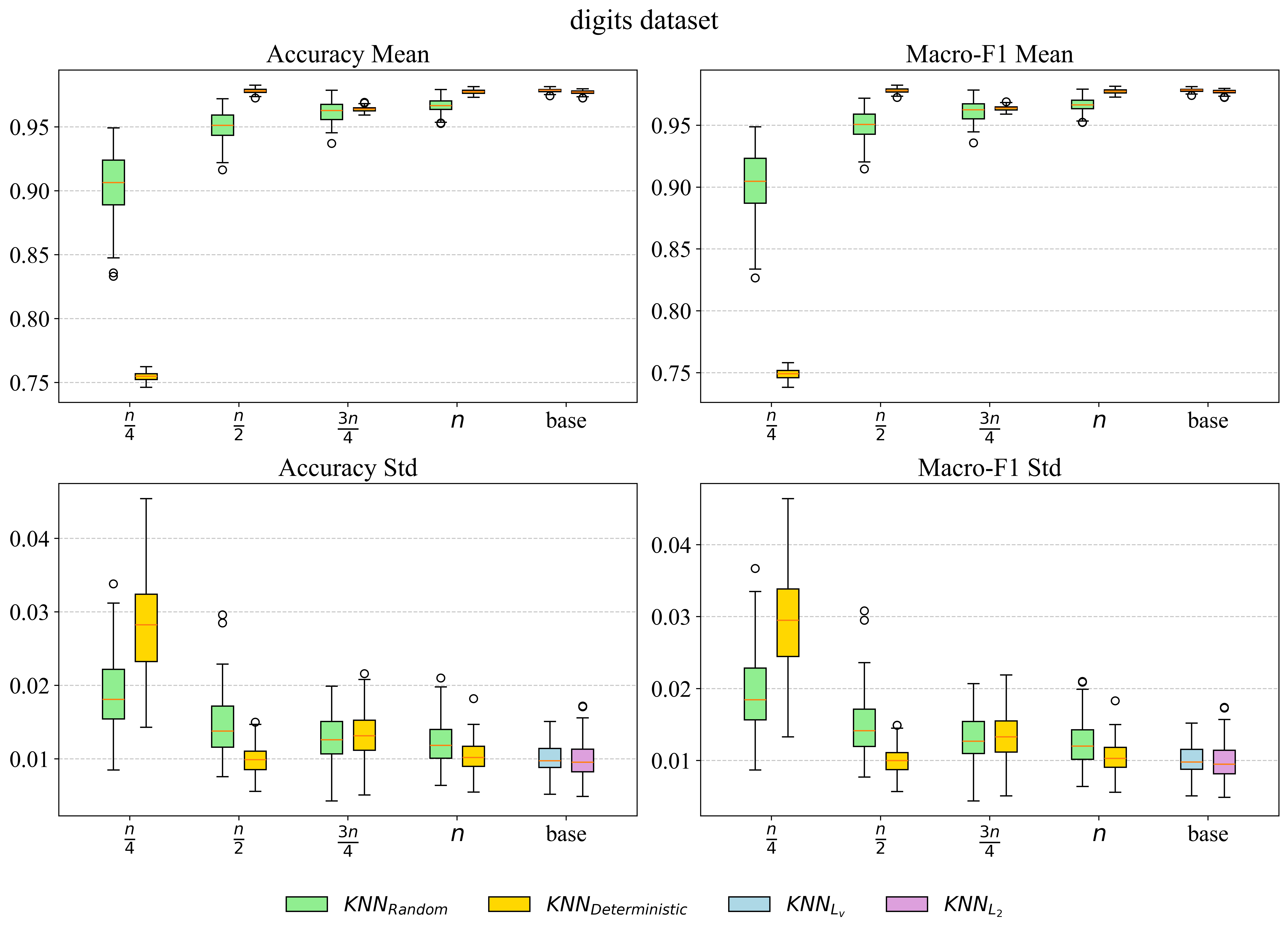}}
    	\end{minipage}
    	\vspace{3pt}
    	\captionsetup{width=0.9\textwidth, font={footnotesize}}
    	\caption{Box plots of KNN algorithm performance based on random and deterministic selection strategies for the ionosphere and digits datasets, with View and Euclidean distances as baselines.}
    	\label{ionosphere_boxplot}
    \end{figure}

	Finally, we evaluated the performance of the $2D$ projection selection strategy based on iterative MWM under the same four $k$ values. Table~\ref{iMWM_result} shows the quantitative results. We compared the deterministic selection strategy with the random projection strategy, alongside two KNN baselines (based on View distance and Euclidean distance). Figures~\ref{toxicity_boxplot}-\ref{ionosphere_boxplot} show the corresponding box plots for intuitive comparison. 
			 		    
    \begin{table}[H]
    	\centering
    	\small
    	\captionsetup{width=0.9\textwidth, font={footnotesize}}
    	\caption{Performance of KNN with random $2D$ projection plane selection strategy across different $k$ values $(n/4, n/2,$ $3n/4, n)$ over 100 runs: mean and mean standard deviation of accuracy and Macro-F1 score, and performance proportion relative to the full projection model.}
    	\label{random_result}
    	\setlength{\tabcolsep}{5pt}
    	\renewcommand{\arraystretch}{1.3}
    	\begin{tabular}{ccccccc|cccc}
    		\toprule
    		Dataset & Metric & Stat & $k_{1}=\dfrac{n}{4}$ & $k_{2}=\dfrac{n}{2}$ & $k_{3}=\dfrac{3n}{4}$ & $k_{4}=n$ & $prop_{1}$ & $prop_{2}$ & $prop_{3}$ & $prop_{4}$ \\
    		\midrule
    		\multirow{4}{*}{toxicity}
    		& accuracy & mean & 0.6145 & 0.6180 & 0.6213 & 0.6231 & 97.06\% & 97.61\% & 98.14\% & 98.42\% \\
    		&          & std  & 0.0926 & 0.0942 & 0.0954 & 0.0975 & & & & \\
    		& macro-f1 & mean & 0.5155 & 0.5205 & 0.5250 & 0.5269 & 95.89\% & 96.82\% & 97.66\% & 98.01\% \\
    		&          & std  & 0.1096 & 0.1135 & 0.1132 & 0.1164 & & & & \\
    		\hline
    		\multirow{4}{*}{darwin}
    		& accuracy & mean & 0.7137 & 0.7077 & 0.7066 & 0.7073 & 102.2\% & 101.3\% & 101.2\% & 101.3\% \\
    		&          & std  & 0.0826 & 0.0806 & 0.0804 & 0.0807 & & & & \\
    		& macro-f1 & mean & 0.6851 & 0.6773 & 0.6761 & 0.6770 & 102.9\% & 101.8\% & 101.6\% & 101.7\% \\
    		&          & std  & 0.1038 & 0.1029 & 0.1027 & 0.1029 & & & & \\
    		\hline
    		\multirow{4}{*}{wine}
    		& accuracy & mean & 0.9008 & 0.9356 & 0.9518 & 0.9595 & 92.99\% & 96.58\% & 98.26\% & 99.05\% \\
    		&          & std  & 0.0638 & 0.0525 & 0.0465 & 0.0434 & & & & \\
    		& macro-f1 & mean &0.9022 & 0.9371 & 0.9532 & 0.9607 & 93.08\% & 96.68\% & 98.34\% & 99.11\% \\
    		&          & std  & 0.0641 & 0.0520 & 0.0454 & 0.0424 & & & & \\
    		\hline
    		\multirow{4}{*}{seeds}
    		& accuracy & mean & 0.8566 & 0.9075 & 0.9169 & 0.9197 & 92.47\% & 97.96\% & 98.97\% & 99.28\% \\
    		&          & std  & 0.0695 & 0.0588 & 0.0568 & 0.056 & & & & \\
    		& macro-f1 & mean & 0.8547 & 0.9061 & 0.9155 & 0.9184 & 92.36\% & 97.91\% & 98.93\% & 99.24\% \\
    		&          & std  & 0.0708 & 0.0601 & 0.0584 & 0.0576 & & & & \\
    		\hline
    		\multirow{4}{*}{ionosphere}
    		& accuracy & mean & 0.8843 & 0.8828 & 0.8809 & 0.8809 & 100.7\% & 100.6\% & 100.4\% & 100.4\% \\
    		&          & std  & 0.0463 & 0.0476 & 0.0483 & 0.0481 & & & & \\
    		& macro-f1 & mean & 0.8634 & 0.8614 & 0.8589 & 0.859 & 100.9\% & 100.7\% & 100.4\% & 100.4\% \\
    		&          & std  & 0.0582 & 0.0600 & 0.0613 & 0.0608 & & & & \\
    		\hline
    		\multirow{4}{*}{digits}
    		& accuracy & mean & 0.9051 & 0.9505 & 0.9618 & 0.9667 & 92.5\% & 97.14\% & 98.29\% & 98.79\% \\
    		&          & std  & 0.0186 & 0.0145 & 0.0130 & 0.0124 & & & & \\
    		& macro-f1 & mean & 0.9034 & 0.9499 & 0.9614 & 0.9664 & 92.33\% & 97.09\% & 98.26\% & 98.77\% \\
    		
    		&          & std  & 0.0192 & 0.0148 & 0.0132 & 0.0125 & & & & \\
    		\bottomrule
    	\end{tabular}
    \end{table}
    
    Similar to the random selection strategy, the deterministic selection strategy yields rising classification performance and stability on all datasets except the ionosphere dataset as $k$ increases. Specifically, both the average accuracy and Macro-F1 score improve as $k$ grows, accompanied by a reduction in standard deviation.
    
    For datasets excluding the ionosphere, the deterministic selection strategy with $k=n/2$ retains more than $98\%$ of the full projection scheme's performance. Notably, it achieves peak performance at $k=n/2$ or $k=n$, and outperforms the full projection scheme on sonar, heart, and wdbc datasets. This indicates that eliminating redundant projection planes helps release classification potential.
    
    The box widths of the deterministic selection strategy are significantly narrower than those of the random selection strategy under all four $k$ settings. Narrower box widths indicate smaller fluctuations in accuracy and F1-score, indicating superior classification stability. This stability advantage is most prominent on the digit and image segmentation datasets. Moreover, the deterministic selection strategy achieves higher mean accuracy and F1-score than the random selection strategy when $k=n/2$ and $k=n$. The above findings confirm that the proposed deterministic selection strategy effectively mitigates the instability defect of the random selection strategy. The deterministic selection strategy performs poorly at $k=n/4$ across most datasets, suggesting that some feature pairs with similar importance scores but high discriminative power are omitted in small-scale selection.
        
    \begin{table}[H]
    	\centering
    	\small
    	\captionsetup{width=0.9\textwidth, font={footnotesize}}
    	\caption{Performance of KNN with $2D$ projection plane selection based on iterative MWM across different $k$ values $(n/4, n/2, 3n/4, n)$ over 100 runs: mean and mean standard deviation of accuracy and Macro-F1 score, and performance proportion relative to the full projection model.}
    	\label{iMWM_result}
    	\setlength{\tabcolsep}{5pt}
    	\renewcommand{\arraystretch}{1.3}
    	\begin{tabular}{ccccccc|cccc}
    		\toprule
    		Dataset & Metric & Stat & $k_{1}=\dfrac{n}{4}$ & $k_{2}=\dfrac{n}{2}$ & $k_{3}=\dfrac{3n}{4}$ & $k_{4}=n$ & $prop_{1}$ & $prop_{2}$ & $prop_{3}$ & $prop_{4}$ \\
    		\midrule
    		\multirow{4}{*}{toxicity}
    		& accuracy & mean & 0.5933 & 0.6333 & 0.6029 & 0.6322 & 93.71\% & 100.00\% & 95.23\% & 99.86\% \\
    		&          & std  & 0.0829 & 0.0951 & 0.0897 & 0.0943 & & & & \\
    		& macro-f1 & mean & 0.4398 & 0.5378 & 0.4809 & 0.5343 & 81.81\% & 100.00\% & 89.45\% & 99.39\% \\
    		&          & std  & 0.0929 & 0.1162 & 0.1082 & 0.1152 & & & & \\
    		\hline
    		\multirow{4}{*}{darwin}
    		& accuracy & mean & 0.6852 & 0.6922 & 0.6891 & 0.6915 & 98.08\% & 99.08\% & 98.64\% & 98.98\% \\
    		&          & std  & 0.0798 & 0.0798 & 0.0803 & 0.0796 & & & & \\
    		& macro-f1 & mean & 0.6477 & 0.6573 & 0.6530 & 0.6564 & 97.31\% & 98.75\% & 98.11\% & 98.62\% \\
    		&          & std  & 0.1065 & 0.1045 & 0.1057 & 0.1045 & & & & \\
    		\hline
    		\multirow{4}{*}{wine}
    		& accuracy & mean & 0.9120 & 0.9627 & 0.9558 & 0.9632 & 94.15\% & 99.38\% & 98.67\% & 99.43\% \\
    		&          & std  & 0.0616 & 0.0404 & 0.0451 & 0.0415 & & & & \\
    		& macro-f1 & mean & 0.9148 & 0.9641 & 0.9577 & 0.9648 & 94.38\%
    		& 99.46\% & 98.8\% & 99.54\% \\
    		&          & std  & 0.0605 & 0.0391 & 0.0435 & 0.0399 & & & & \\
    		\hline
    		\multirow{4}{*}{seeds}
    		& accuracy & mean & 0.8514 & 0.9162 & 0.9020 & 0.9175 & 91.9\% & 98.9\% & 97.37\% & 99.04\% \\
    		&          & std  & 0.0724 & 0.0552 & 0.0594 & 0.0550 & & & & \\
    		& macro-f1 & mean & 0.8490 & 0.9144 & 0.8992 & 0.9156 & 91.74\% & 98.81\% & 97.17\% & 98.94\% \\
    		&          & std  & 0.0745 & 0.0574 & 0.0623 & 0.0573 & & & & \\
    		\hline
    		\multirow{4}{*}{ionosphere}
    		& accuracy & mean & 0.8851 & 0.8699 & 0.8730 & 0.8726 & 100.8\% & 99.1\% & 99.45\% & 99.41\% \\
    		&          & std  & 0.0483 & 0.0472 & 0.0484 & 0.0468 & & & & \\
    		& macro-f1 & mean & 0.8620 & 0.8448 & 0.8489 & 0.8486 & 100.8\% & 98.76\% & 99.24\% & 99.21\% \\
    		&          & std  & 0.0633 & 0.0607 & 0.0617 & 0.0598 & & & & \\
    		\hline
    		\multirow{4}{*}{digits}
    		& accuracy & mean & 0.7547 & 0.9785 & 0.9640 & 0.9776 & 77.13\% & 100\% & 98.52\% & 99.91\% \\
    		&          & std  & 0.0281 & 0.0099 & 0.0130 & 0.0104 & & & & \\
    		& macro-f1 & mean & 0.7492 & 0.9783 & 0.9637 & 0.9775 & 76.57\% & 99.99\% & 98.5\% & 99.91\% \\
    		&          & std  & 0.0291 & 0.0100 & 0.0132 & 0.0105 & & & & \\
    		\bottomrule
    	\end{tabular}
    \end{table}
    
    In addition, classification performance degrades as $k$ rises from $n/2$ to $3n/4$. These observations further confirm that selecting feature pairs solely based on high importance scores cannot guarantee favorable overall classification accuracy. Both the importance and the diversity of features should be considered simultaneously.    
    
	\subsubsection{Ablation experiments}
    When the hyperparameter $k$ is set to $n/4$, the model using the iterative MWM selection strategy achieves suboptimal performance across most datasets. To deeply explore the intrinsic mechanisms underlying these performance differences, we conduct ablation experiments on 4 benchmark datasets: darwin, wine, seeds, and ionosphere, where wine and seeds yield poorer performance at $k=n/4$ than at $k=n/2$, while darwin and ionosphere obtain better results with $k=n/4$. According to the iterative MWM selection rule, selecting $k=n/4$ projection planes covers $n/2$ of the original features, whereas selecting $k=n/2$ planes fully incorporates all $n$ original features.

	The experimental procedures are implemented as follows. We sequentially select the top $n$ mutually exclusive projection planes based on their importance scores, where $n$ equals the dataset's feature dimensionality. Distance calculations are based on the chosen set of projection planes. To ensure the stability and reliability of the experiments, all experiments are repeated using 100 different random seeds, and the mean and standard deviation of classification accuracy and Macro-F1 scores are recorded. The experimental results demonstrate that the classification performance of wine and seeds improves as the number of selected projection planes increases, with the performance inflection point occurring at approximately $k=n/2$. In contrast, the classification metrics for darwin and ionosphere increase initially and then decline as the number of selected projection planes increases.
	
	\begin{figure}[H]
		\centering
		\includegraphics[width=0.7\textwidth]{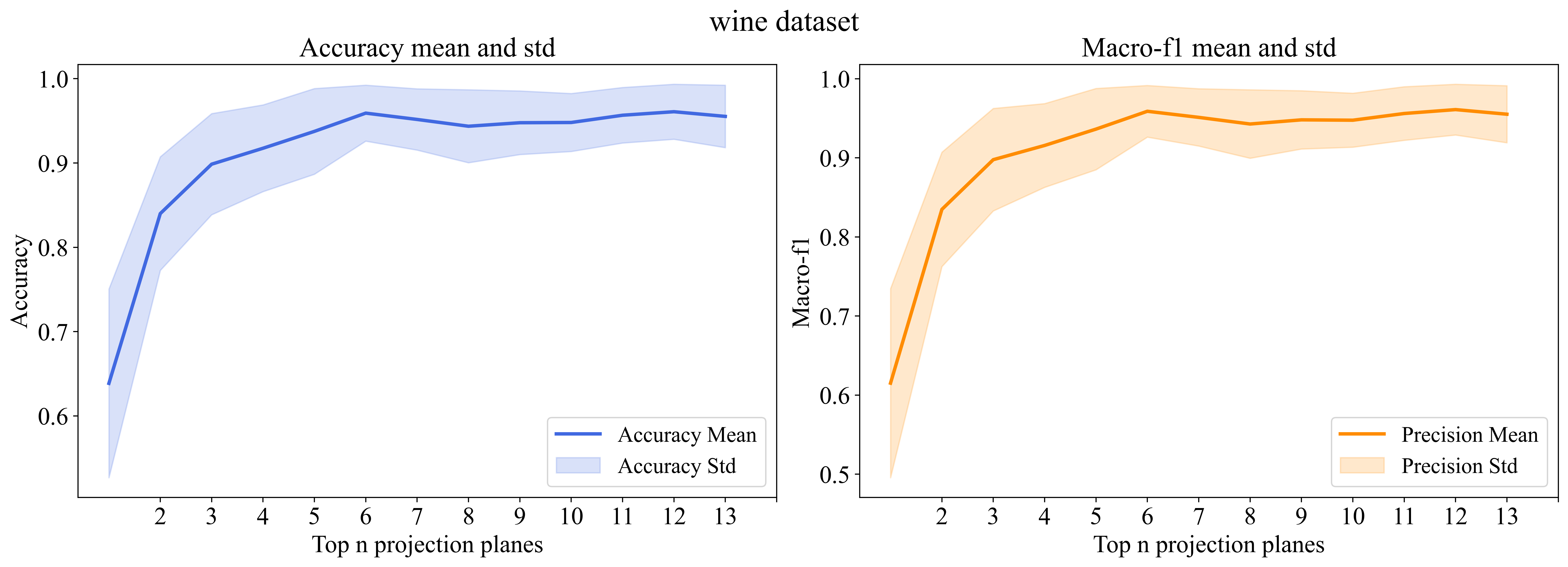}
		\captionsetup{width=0.9\textwidth, font={footnotesize}}
		\caption{Classification performance variations of the KNN algorithm based on top‑n projection planes with the highest importance scores on the wine dataset.}
		\label{wine_ablation_plot}
	\end{figure}
	
	\begin{figure}[H]
		\centering
		\includegraphics[width=0.7\textwidth]{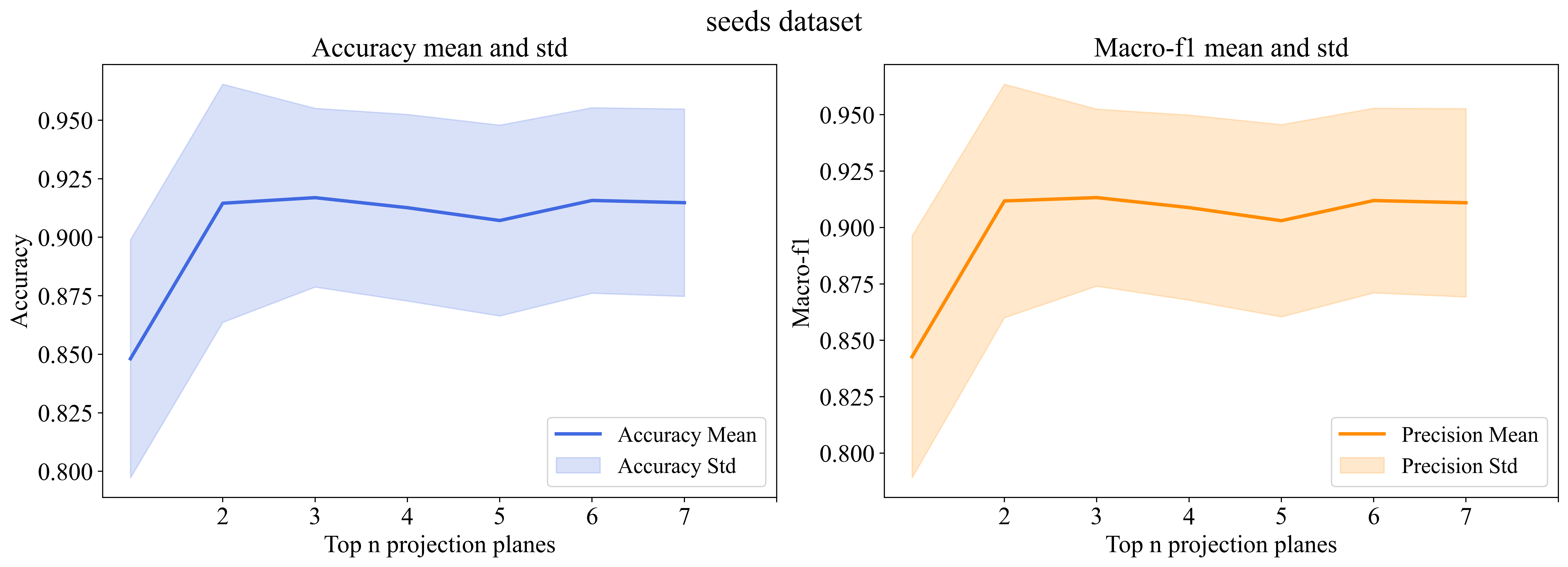}
		\captionsetup{width=0.9\textwidth, font={footnotesize}}
		\caption{Classification performance variations of the KNN algorithm based on top‑n projection planes with the highest importance scores on the seeds dataset.}
		\label{seeds_ablation_plot}
	\end{figure}
	
	The diverse experimental phenomena above can be explained by analyzing the importance score distributions of projection planes across different datasets. As illustrated in Figure~\ref{scoresLow}, the importance scores of all $2D$ projection planes in the seeds dataset range from 0.3 to 0.5 with slight differences and a uniform distribution. This indicates that the dataset exhibits strong linear separability and low redundancy in the original features, so additional discriminative information can hardly be extracted via a $2D$ projection scheme. In contrast, according to Figure~\ref{scoresMiddel}, the importance scores of $2D$ projection planes for the ionosphere dataset range from 0.3 to 1, with most values concentrated around 0.5 and a secondary peak around 0.8. This distribution indicates that the dataset contains many redundant or noisy features. As the number of selected projection planes $k$ increases, more projection planes corresponding to redundant features are incorporated. The discriminative gain from valid features is gradually offset, leading to degraded classification performance.
	
	In summary, the optimal number of selected projection planes is highly correlated with the dataset's redundancy and noise level. For datasets with low redundancy and few noisy features, the importance scores of projection planes are uniformly distributed, with a low proportion of low-quality redundant planes. In this case, plane selection should balance both importance and feature diversity. Setting $k=n/2$ can fully cover all original features and sufficiently exploit valid discriminative information, thus achieving promising performance. On the contrary, for datasets rich in redundant and noisy features, most projection plane importance scores fall in the low-value region. Selecting $k=n/2$ will introduce a large number of invalid noisy planes and degrade model performance. In comparison, setting $k=n/4$ effectively excludes low-quality redundant and noisy planes with low rankings. Therefore, the Darwin and Ionosphere datasets achieve optimal performance at $k=n/4$, confirming that superior classification and clustering results can sometimes be obtained by relying solely on a small set of high-quality projection planes. Therefore, the hyperparameter $k$ can be set by analyzing the distribution of importance scores across projection planes.
	
	\begin{figure}[H]
		\centering
		\includegraphics[width=0.7\textwidth]{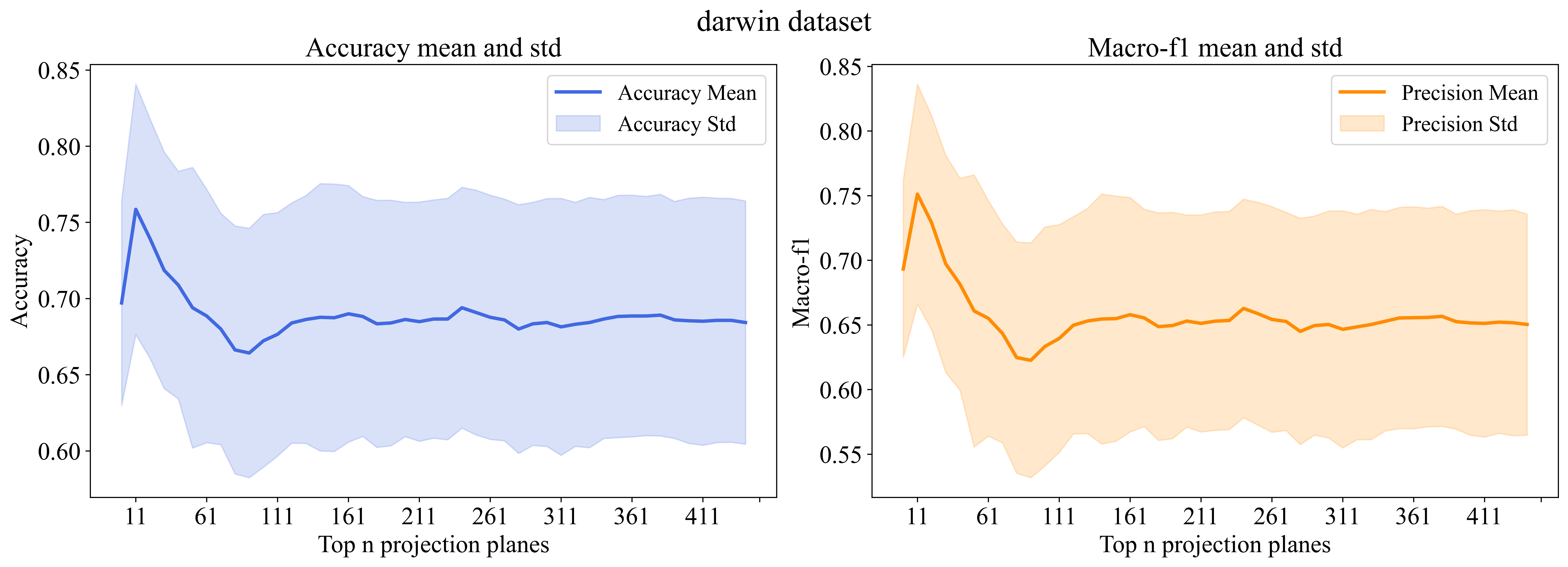}
		\captionsetup{width=0.9\textwidth, font={footnotesize}}
		\caption{Classification performance variations of the KNN algorithm based on top‑n projection planes with the highest importance scores on the darwin dataset.}
		\label{darwin_ablation_plot}
	\end{figure}
	
	\begin{figure}[H]
		\centering
		\includegraphics[width=0.7\textwidth]{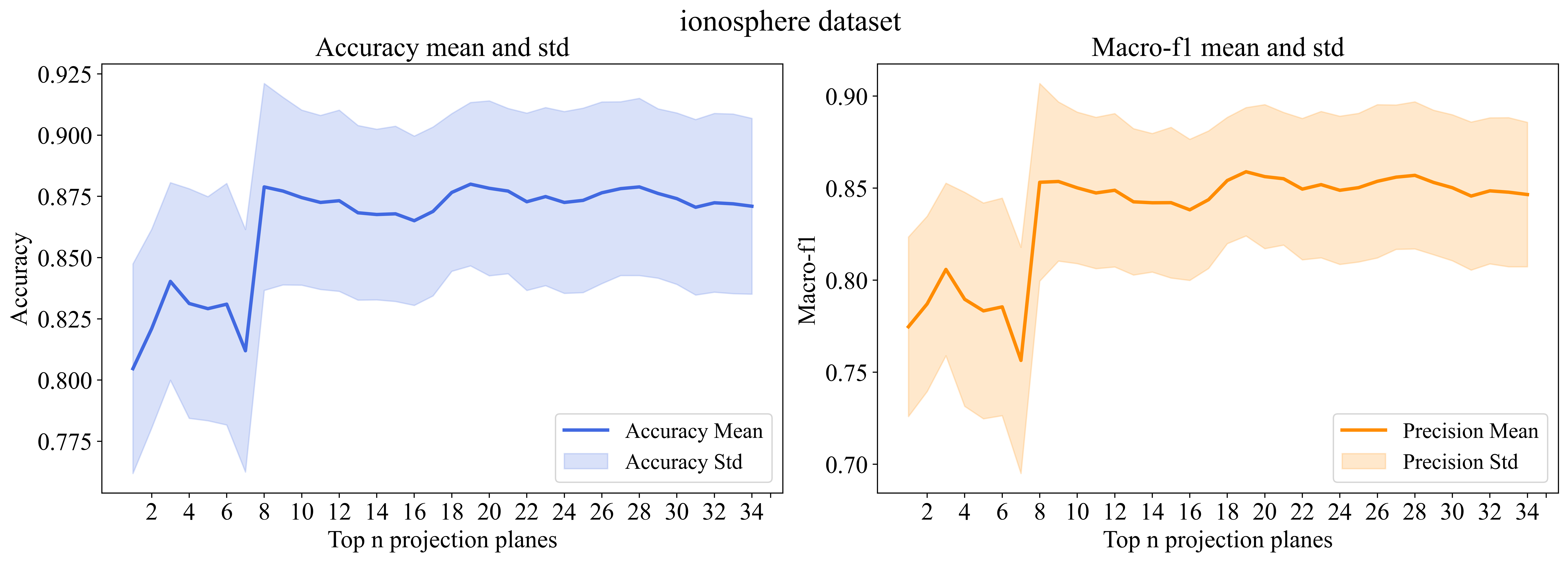}
		\captionsetup{width=0.9\textwidth, font={footnotesize}}
		\caption{Classification performance variations of the KNN algorithm based on top‑n projection planes with the highest importance scores on the ionosphere dataset.}
		\label{ionosphere_ablation_plot}
	\end{figure}
	
	\subsubsection{Discussion}       
	Based on 3 groups of comparative experiments, including baseline comparison, random selection strategy, and deterministic selection strategy, together with statistical tests of 100 repeated experiments and stability analysis via box plots, we draw the following conclusions. Compared with the Euclidean distance, the View distance generally achieves higher classification accuracy and greater model stability. In contrast to the random selection strategy, the proposed deterministic selection strategy can obtain higher accuracy and more stable classification performance with fewer projection planes.
	
	In terms of the selection mechanism of projection planes, the random selection strategy tends to select numerous redundant projection planes that share the same features, resulting in overlap of feature information. On the contrary, the proposed deterministic selection strategy enforces mutual exclusivity of dimensions across selected planes and avoids feature over-concentration. Although a small number of feature pairs with similar importance scores are omitted when $k=n/4$, the model performance can approach or even exceed that using all projection planes when $k=n/2$. This strategy balances computational cost and model performance well while mitigating the high computational complexity of the original View distance. For practical deployment, we recommend taking $k=n/2$ as the preferred configuration. The full projection scheme is adopted only for low-noise datasets with extremely high requirements for classification accuracy.
		    	    			
	\section{Limitations and future work}\label{sec8}
	Although applying the View distance has achieved some results, this study still has several limitations. First, all empirical validations in this work are conducted using DBSCAN clustering and KNN classification models. The generalizability of the proposed View distance and projection plane selection strategy to other distance‑sensitive machine learning models remains to be further verified. Second, all experiments are conducted on small‑scale datasets with balanced class distributions. The performance of the proposed distance metric in class‑imbalanced scenarios and on medium‑to‑large scale datasets still requires further investigation. Finally, this study only provides a practical method for selecting the number of projection planes based on the importance score distribution, rather than establishing a data‑adaptive criterion for determining the optimal number of planes. In future work, we will conduct supplementary experiments focusing on three key research directions: cross‑model generalizability, adaptability to complex datasets, and adaptive determination of the optimal $k$ value.

    \textbf{Acknowledgements}. \emph{The authors are very much indebted to the referees for their constructive comments and valuable suggestions}.

    {\bf Data Availability}  The codes generated during and analysed during the current study are available from the corresponding author on reasonable request, and we will publicly release the code used in this manuscript on GitHub.


				
	\bibliographystyle{ieeetr}
	\bibliography{reference}
	
	\appendix
	\section{Experimental results}
	\subsection{Numerical results}
	\begin{table}[H]
		\centering
		\small
		\captionsetup{width=0.9\textwidth, font={footnotesize}}
		\caption{The mean accuracy, mean Macro-F1 score, and the average standard deviation obtained by repeating 100 evaluations of the KNN algorithm based on $L_p$ distance and View distance over 8 low-to-medium dimensional datasets. Two-sample $t$-tests are conducted on the experimental results of Euclidean distance and View distance.}
		\setlength{\tabcolsep}{5pt}
		\renewcommand{\arraystretch}{1.3}
		\begin{tabular}{cccccccc|ccc}
			\toprule
			Dataset & Metric & Stat & $\text{KNN}_{L_{1}}$ & $\text{KNN}_{L_{2}}$ & $\text{KNN}_{L_{3}}$ & $\text{KNN}_{L_{4}}$ & $\text{KNN}_{L_{\infty}}$ & $\text{KNN}_{L_{v}}$ & p-value & t-stat\\
			\midrule
			\multirow{4}{*}{wine}
			& accuracy & mean & 0.9644 & 0.9664 & 0.9543 & 0.9499 & 0.9208 & \textbf{0.9687} & 0.001 & 3.414 \\
			&          & std  & 0.0403 & 0.0394 & 0.0455 & 0.0469 & 0.0587 & \textbf{0.0384} & & \\
			& macro-f1 & mean & 0.9652 & 0.9674 & 0.9552 & 0.9509 & 0.9214 & \textbf{0.9693} & 0.0036 & 2.983 \\
			&          & std  & 0.0396 & 0.0385 & 0.0451 & 0.0465 & 0.0592 & \textbf{0.0378} & & \\
			\hline
			\multirow{4}{*}{sonar}
			& accuracy & mean & \textbf{0.8370} & 0.8189 & 0.8085 & 0.7878 & 0.7414 & 0.8316 & $***$ \tnote{a} & 11.00 \\
			&          & std  & \textbf{0.0744} & 0.0757 & 0.0780 & 0.0827 & 0.0861 & 0.0762 & & \\
			& macro-f1 & mean & \textbf{0.8318} & 0.8124 & 0.8013 & 0.7789 & 0.7313 & 0.8261 & $***$ & 11.09 \\
			&          & std  & \textbf{0.0782} & 0.0803 & 0.0831 & 0.089 & 0.0918 & 0.08 & & \\
			\hline
			\multirow{4}{*}{seeds}
			& accuracy & mean & 0.9250 & \textbf{0.9309} & 0.9238 & 0.9224 & 0.9200 & 0.9264 & $***$ & -7.081 \\
			&          & std  & 0.0524 & \textbf{0.0509} & 0.0554 & 0.0559 & 0.0549 & 0.0516 & & \\
			& macro-f1 & mean & 0.9238 & \textbf{0.9301} & 0.9231 & 0.9217 & 0.9192 & 0.9254 & $***$ & -7.201 \\
			&          & std  & 0.0538 & \textbf{0.0518} & 0.0561 & 0.0565 & 0.0557 & 0.0527 & & \\
			\hline
			\multirow{4}{*}{heart}
			& accuracy & mean & 0.8283 & 0.8250 & 0.8188 & 0.8123 & 0.7778 & \textbf{0.832} & $***$ & 7.143 \\
			&          & std  & 0.0642 & 0.0654 & 0.0656 & 0.0678 & 0.0699 & \textbf{0.0641} & & \\
			& macro-f1 & mean & 0.8254 & 0.8221 & 0.8154 & 0.8092 & 0.7739 & \textbf{0.8292} & $***$ & 7.139 \\
			&          & std  & 0.0659 & 0.0668 & 0.0674 & 0.0695 & 0.0716 & \textbf{0.0656} & & \\
			\hline
			\multirow{4}{*}{ionosphere}
			& accuracy & mean & \textbf{0.8868} & 0.8496 & 0.8252 & 0.8288 & 0.8495 & 0.8778 & $***$ & 45.76 \\
			&          & std  & \textbf{0.0463} & 0.0492 & 0.0496 & 0.0494 & 0.0483 & 0.0473 & & \\
			& macro-f1 & mean & \textbf{0.8670} & 0.8170 & 0.7806 & 0.7855 & 0.8154 & 0.8554 & $***$ & 45.87 \\
			&          & std  & \textbf{0.0578} & 0.0659 & 0.0715 & 0.0709 & 0.0661 & 0.0599 & & \\
			\hline
			\multirow{4}{*}{wdbc}
			& accuracy & mean & 0.9674 & 0.9676 & 0.9660 & 0.9639 & 0.9460 & \textbf{0.9684} & 0.0123 & 2.550 \\
			&          & std  & 0.0214 & 0.0221 & 0.0217 & 0.0229 & 0.0275 & \textbf{0.0212} & & \\
			& macro-f1 & mean & 0.9646 & 0.9647 & 0.9630 & 0.9608 & 0.9414 & \textbf{0.9656} & 0.0099 & 2.629 \\
			&          & std  & 0.0235 & 0.0243 & 0.0238 & 0.0251 & 0.0302 & \textbf{0.0233} & & \\
			\hline
			\multirow{4}{*}{digits}
			& accuracy & mean & 0.9770 & 0.9773 & 0.9724 & 0.9667 & 0.9361 & \textbf{0.9785} & $***$ & 6.694 \\
			&          & std  & 0.0102 & \textbf{0.0100} & 0.0111 & 0.0120 & 0.0170 & \textbf{0.0100} & & \\
			& macro-f1 & mean & 0.9768 & 0.9772 & 0.9723 & 0.9676 & 0.9356 & \textbf{0.9784} & $***$ & 6.403 \\
			&          & std  & 0.0104 & \textbf{0.0101} & 0.0112 & 0.0122 & 0.0172 & \textbf{0.0101} & & \\
			\hline
			\multirow{4}{*}{image}
			& accuracy & mean & \textbf{0.9592} & 0.9435 & 0.9366 & 0.9332 & 0.9248 & 0.9576 & $***$ & 73.91 \\
			&          & std  & \textbf{0.0120} & 0.0137 & 0.0141 & 0.0148 & 0.0158 & 0.0121 & & \\
			& macro-f1 & mean & \textbf{0.9589} & 0.9427 & 0.9357 & 0.9322 & 0.9237 & 0.9573 & $***$ & 76.03 \\
			&          & std  & \textbf{0.0122} & 0.0141 & 0.0145 & 0.0152 & 0.0163 & 0.0123 & & \\
			\bottomrule
		\end{tabular}
			
		\begin{tablenotes}  
			\footnotesize   
			\item[a] $*** \textit{P} < 0.001$.
		\end{tablenotes}
	\end{table}
	
	\begin{table}[H]
		\centering
		\small
		\captionsetup{width=0.9\textwidth, font={footnotesize}}
		\caption{Performance of KNN with random $2D$ projection plane selection strategy across different $k$ values $(n/4, n/2,$ $3n/4, n)$ over 100 runs: mean and mean standard deviation of accuracy and Macro-F1 score, and performance proportion relative to the full projection model.}
		\setlength{\tabcolsep}{5pt}
		\renewcommand{\arraystretch}{1.3}
		\begin{tabular}{ccccccc|cccc}
			\toprule
			Dataset & Metric & Stat & $k_{1}=\dfrac{n}{4}$ & $k_{2}=\dfrac{n}{2}$ & $k_{3}=\dfrac{3n}{4}$ & $k_{4}=n$ & $prop_{1}$ & $prop_{2}$ & $prop_{3}$ & $prop_{4}$ \\
			\midrule
			\multirow{4}{*}{wine}
			& accuracy & mean & 0.9008 & 0.9356 & 0.9518 & 0.9595 & 92.99\% & 96.58\% & 98.26\% & 99.05\% \\
			&          & std  & 0.0638 & 0.0525 & 0.0465 & 0.0434 & & & & \\
			& macro-f1 & mean &0.9022 & 0.9371 & 0.9532 & 0.9607 & 93.08\% & 96.68\% & 98.34\% & 99.11\% \\
			&          & std  & 0.0641 & 0.0520 & 0.0454 & 0.0424 & & & & \\
			\hline
			\multirow{4}{*}{sonar}
			& accuracy & mean & 0.8131 & 0.8226 & 0.8293 & 0.8295 & 97.78\% & 98.92\% & 99.72\% & 99.75\% \\
			&          & std  & 0.0819 & 0.0775 & 0.0774 & 0.0796 & & & & \\
			& macro-f1 & mean & 0.8076 & 0.8173 & 0.8238 & 0.8241 & 97.76\% & 98.93\% & 99.72\% & 99.76\% \\
			&          & std  & 0.0859 & 0.0809 & 0.0815 & 0.0836 & & & & \\
			\hline
			\multirow{4}{*}{seeds}
			& accuracy & mean & 0.8566 & 0.9075 & 0.9169 & 0.9197 & 92.47\% & 97.96\% & 98.97\% & 99.28\% \\
			&          & std  & 0.0695 & 0.0588 & 0.0568 & 0.056 & & & & \\
			& macro-f1 & mean & 0.8547 & 0.9061 & 0.9155 & 0.9184 & 92.36\% & 97.91\% & 98.93\% & 99.24\% \\
			&          & std  & 0.0708 & 0.0601 & 0.0584 & 0.0576 & & & & \\
			\hline
			\multirow{4}{*}{heart}
			& accuracy & mean & 0.7467 & 0.7828 & 0.7986 & 0.8104 & 89.75\% & 94.09\% & 95.99\% & 97.4\% \\
			&          & std  & 0.0713 & 0.0707 & 0.0674 & 0.0656 & & & & \\
			& macro-f1 & mean & 0.7418 & 0.7784 & 0.7948 & 0.8071 & 89.46\% & 93.87\% & 95.85\% & 97.33\% \\
			&          & std  & 0.0733 & 0.0725 & 0.0693 & 0.0673 & & & & \\
			\hline
			\multirow{4}{*}{ionosphere}
			& accuracy & mean & 0.8843 & 0.8828 & 0.8809 & 0.8809 & 100.7\% & 100.6\% & 100.4\% & 100.4\% \\
			&          & std  & 0.0463 & 0.0476 & 0.0483 & 0.0481 & & & & \\
			& macro-f1 & mean & 0.8634 & 0.8614 & 0.8589 & 0.859 & 100.9\% & 100.7\% & 100.4\% & 100.4\% \\
			&          & std  & 0.0582 & 0.0600 & 0.0613 & 0.0608 & & & & \\
			\hline
			\multirow{4}{*}{wdbc}
			& accuracy & mean & 0.9513 & 0.9609 & 0.9629 & 0.9648 & 98.23\% & 99.23\% & 99.43\% & 99.63\% \\
			&          & std  & 0.0260 & 0.0235 & 0.0231 & 0.0223 & & & & \\
			& macro-f1 & mean & 0.9471 & 0.9576 & 0.9597 & 0.9618 & 98.08\% & 99.17\% & 99.39\% & 99.61\% \\
			&          & std  & 0.0285 & 0.0258 & 0.0253 & 0.0244 & & & & \\
			\hline
			\multirow{4}{*}{digits}
			& accuracy & mean & 0.9051 & 0.9505 & 0.9618 & 0.9667 & 92.5\% & 97.14\% & 98.29\% & 98.79\% \\
			&          & std  & 0.0186 & 0.0145 & 0.0130 & 0.0124 & & & & \\
			& macro-f1 & mean & 0.9034 & 0.9499 & 0.9614 & 0.9664 & 92.33\% & 97.09\% & 98.26\% & 98.77\% \\
			
			&          & std  & 0.0192 & 0.0148 & 0.0132 & 0.0125 & & & & \\
			\hline
			\multirow{4}{*}{image}
			& accuracy & mean & 0.8982 & 0.9290 & 0.9422 & 0.9497 & 93.8\% & 97.01\% & 98.39\% & 99.18\% \\
			&          & std  & 0.0174 & 0.0150 & 0.0141 & 0.0133 & & & & \\
			& macro-f1 & mean & 0.8967 & 0.9279 & 0.9415 & 0.9492 & 93.67\% & 96.93\% & 98.35\% & 99.15\% \\
			&          & std  & 0.0177 & 0.0154 & 0.0144 & 0.0136 & & & & \\
			\bottomrule
		\end{tabular}
	\end{table}
	
	\begin{table}[H]
		\centering
		\small
		\captionsetup{width=0.9\textwidth, font={footnotesize}}
		\caption{Performance of KNN with $2D$ projection plane selection based on iterative MWM across different $k$ values $(n/4, n/2, 3n/4, n)$ over 100 runs: mean and mean standard deviation of accuracy and Macro-F1 score, and performance proportion relative to the full projection model.}
		\setlength{\tabcolsep}{5pt}
		\renewcommand{\arraystretch}{1.3}
		\begin{tabular}{ccccccc|cccc}
			\toprule
			Dataset & Metric & Stat & $k_{1}=\dfrac{n}{4}$ & $k_{2}=\dfrac{n}{2}$ & $k_{3}=\dfrac{3n}{4}$ & $k_{4}=n$ & $prop_{1}$ & $prop_{2}$ & $prop_{3}$ & $prop_{4}$ \\
			\midrule
			\multirow{4}{*}{wine}
			& accuracy & mean & 0.9120 & 0.9627 & 0.9558 & 0.9632 & 94.15\% & 99.38\% & 98.67\% & 99.43\% \\
			&          & std  & 0.0616 & 0.0404 & 0.0451 & 0.0415 & & & & \\
			& macro-f1 & mean & 0.9148 & 0.9641 & 0.9577 & 0.9648 & 94.38\%
			& 99.46\% & 98.8\% & 99.54\% \\
			&          & std  & 0.0605 & 0.0391 & 0.0435 & 0.0399 & & & & \\
			\hline
			\multirow{4}{*}{sonar}
			& accuracy & mean & 0.7577 & 0.8319 & 0.8112 & 0.8285 & 91.11\% & 100\% & 97.55\% & 99.63\% \\
			&          & std  & 0.0861 & 0.0767 & 0.0753 & 0.0767 & & & & \\
			& macro-f1 & mean & 0.7496 & 0.8256 & 0.8033 & 0.8228 & 90.74\% & 99.94\% & 97.24\% & 99.6\% \\
			&          & std  & 0.0906 & 0.0817 & 0.0806 & 0.0809 & & & & \\
			\hline
			\multirow{4}{*}{seeds}
			& accuracy & mean & 0.8514 & 0.9162 & 0.9020 & 0.9175 & 91.9\% & 98.9\% & 97.37\% & 99.04\% \\
			&          & std  & 0.0724 & 0.0552 & 0.0594 & 0.0550 & & & & \\
			& macro-f1 & mean & 0.8490 & 0.9144 & 0.8992 & 0.9156 & 91.74\% & 98.81\% & 97.17\% & 98.94\% \\
			&          & std  & 0.0745 & 0.0574 & 0.0623 & 0.0573 & & & & \\
			\hline
			\multirow{4}{*}{heart}
			& accuracy & mean & 0.6764 & 0.8374 & 0.8207 & 0.8451 & 81.3\% & 100.6\% & 98.64\% & 101.6\% \\
			&          & std  & 0.0784 & 0.0626 & 0.0647 & 0.0614 & & & & \\
			& macro-f1 & mean & 0.6687 & 0.8353 & 0.8171 & 0.8431 & 80.64\% & 100.7\% & 98.54\% & 101.7\% \\
			&          & std  & 0.0813 & 0.0638 & 0.0668 & 0.0625 & & & & \\
			\hline
			\multirow{4}{*}{ionosphere}
			& accuracy & mean & 0.8851 & 0.8699 & 0.8730 & 0.8726 & 100.8\% & 99.1\% & 99.45\% & 99.41\% \\
			&          & std  & 0.0483 & 0.0472 & 0.0484 & 0.0468 & & & & \\
			& macro-f1 & mean & 0.8620 & 0.8448 & 0.8489 & 0.8486 & 100.8\% & 98.76\% & 99.24\% & 99.21\% \\
			&          & std  & 0.0633 & 0.0607 & 0.0617 & 0.0598 & & & & \\
			\hline
			\multirow{4}{*}{wdbc}
			& accuracy & mean & 0.9279 & 0.9692 & 0.9627 & 0.9692 & 95.82\% & 100.1\% & 99.41\% & 100.1\% \\
			&          & std  & 0.0326 & 0.0210 & 0.0227 & 0.0212 & & & & \\
			& macro-f1 & mean & 0.9218 & 0.9664 & 0.9595 & 0.9665 & 95.46\% & 100.1\% & 99.37\% & 100.1\% \\
			&          & std  & 0.0356 & 0.0231 & 0.0249 & 0.0233 & & & & \\
			\hline
			\multirow{4}{*}{digits}
			& accuracy & mean & 0.7547 & 0.9785 & 0.9640 & 0.9776 & 77.13\% & 100\% & 98.52\% & 99.91\% \\
			&          & std  & 0.0281 & 0.0099 & 0.0130 & 0.0104 & & & & \\
			& macro-f1 & mean & 0.7492 & 0.9783 & 0.9637 & 0.9775 & 76.57\% & 99.99\% & 98.5\% & 99.91\% \\
			&          & std  & 0.0291 & 0.0100 & 0.0132 & 0.0105 & & & & \\
			\hline
			\multirow{4}{*}{image}
			& accuracy & mean & 0.8439 & 0.9457 & 0.9346 & 0.9501 & 88.13\% & 98.76\% & 97.6\% & 99.22\% \\
			&          & std  & 0.0207 & 0.0136 & 0.0155 & 0.0143 & & & & \\
			& macro-f1 & mean & 0.8407 & 0.9453 & 0.9339 & 0.9497 & 87.82\% & 98.75\% & 97.56\% & 99.21\% \\
			&          & std  & 0.0215 & 0.0138 & 0.0158 & 0.0145 & & & & \\
			\bottomrule
		\end{tabular}
	\end{table}
	
	\begin{table}[H]
		\centering
		\small
		\captionsetup{width=0.9\textwidth, font={footnotesize}}
		\caption{The mean accuracy, mean Macro-F1 score, and the average standard deviation obtained by repeating 100 evaluations of the KNN algorithm based on $L_p$ distance and View distance over 4 high-dimensional datasets. Two‑sample $t$-tests are conducted on the experimental results of Euclidean distance and View distance.}
		\setlength{\tabcolsep}{5pt}
		\renewcommand{\arraystretch}{1.3}
		\begin{tabular}{cccccccc|ccc}
			\toprule
			Dataset & Metric & Stat & $\text{KNN}_{L_{1}}$ & $\text{KNN}_{L_{2}}$ & $\text{KNN}_{L_{3}}$ & $\text{KNN}_{L_{4}}$ & $\text{KNN}_{L_{\infty}}$ & $\text{KNN}_{L_{v}}$ & p-value & t-stat\\
			\midrule
			\multirow{4}{*}{toxicity}
			& accuracy & mean & 0.6303 & 0.6098 & 0.6137 & 0.5693 & 0.5413 & \textbf{0.6331} & *** & 11.43 \\
			&          & std  & 0.0949 & \textbf{0.0940} & \textbf{0.0940} & 0.0969 & 0.1063 & 0.0949 & & \\
			& macro-f1 & mean & 0.5338 & 0.5109 & 0.5158 & 0.4708 & 0.4855 & \textbf{0.5376} & *** & 10.93 \\
			&          & std  & 0.1151 & 0.1137 & 0.1124 & \textbf{0.1076} & 0.1108 & 0.1154 & & \\
			\hline
			\multirow{4}{*}{darwin}
			& accuracy & mean & 0.6975 & 0.6903 & 0.7171 & \textbf{0.7185} & 0.6517 & 0.6986 & 0.1276 & 7.076 \\
			&          & std  & \textbf{0.0802} & 0.0877 & 0.0908 & 0.0953 & 0.1041 & 0.0805 & & \\
			& macro-f1 & mean & 0.6643 & 0.6633 & 0.7028 & \textbf{0.7100} & 0.6452 & 0.6656 & *** & 1.536 \\
			&          & std  & 0.1044 & 0.1066 & \textbf{0.1016} & \textbf{0.1016} & 0.1061 & 0.1045 & & \\
			\hline
			\multirow{4}{*}{musk}
			& accuracy & mean & 0.8652 & \textbf{0.8849} & 0.8663 & 0.8487 & 0.7629 & 0.8767 & *** & -9.899 \\
			&          & std  & 0.0470 & \textbf{0.0436} & 0.0468 & 0.0486 & 0.0555 & 0.0463 & & \\
			& macro-f1 & mean & 0.8644 & \textbf{0.8842} & 0.8655 & 0.8479 & 0.7619 & 0.8759 & *** & -9.955 \\
			&          & std  & 0.0472 & \textbf{0.0437} & 0.0469 & 0.0487 & 0.0558 & 0.0464 & & \\
			\hline
			\multirow{4}{*}{drug}
			& accuracy & mean & \textbf{0.7808} & 0.7594 & 0.7597 & 0.7504 & 0.7286 & 0.7792 & *** & 19.19 \\
			&          & std  & 0.0527 & 0.0525 & 0.0511 & 0.0509 & \textbf{0.0490} & 0.0535 & & \\
			& macro-f1 & mean & \textbf{0.6957} & 0.6594 & 0.6441 & 0.6206 & 0.5843 & 0.6937 & *** & 24.20 \\
			&          & std  & \textbf{0.0732} & 0.0733 & 0.0761 & 0.0763 & 0.0737 & 0.0740 & & \\
			\bottomrule
		\end{tabular}
	\end{table}
	
	\begin{table}[H]
		\centering
		\small
		\captionsetup{width=0.9\textwidth, font={footnotesize}}
		\caption{Performance of KNN with $2D$ projection plane selection based on iterative MWM across different $k$ values $(n/4, n/2, 3n/4, n)$ over 100 runs: mean and mean standard deviation of accuracy and Macro-F1 score, and performance proportion relative to the full projection model.}
		\setlength{\tabcolsep}{5pt}
		\renewcommand{\arraystretch}{1.3}
		\begin{tabular}{ccccccc|cccc}
			\toprule
			Dataset & Metric & Stat & $k_{1}=\dfrac{n}{4}$ & $k_{2}=\dfrac{n}{2}$ & $k_{3}=\dfrac{3n}{4}$ & $k_{4}=n$ & $prop_{1}$ & $prop_{2}$ & $prop_{3}$ & $prop_{4}$ \\
			\midrule
			\multirow{4}{*}{toxicity}
			& accuracy & mean & 0.5933 & 0.6333 & 0.6029 & 0.6322 & 93.71\% & 100.00\% & 95.23\% & 99.86\% \\
			&          & std  & 0.0829 & 0.0951 & 0.0897 & 0.0943 & & & & \\
			& macro-f1 & mean & 0.4398 & 0.5378 & 0.4809 & 0.5343 & 81.81\% & 100.00\% & 89.45\% & 99.39\% \\
			&          & std  & 0.0929 & 0.1162 & 0.1082 & 0.1152 & & & & \\
			\hline
			\multirow{4}{*}{darwin}
			& accuracy & mean & 0.6852 & 0.6922 & 0.6891 & 0.6915 & 98.08\% & 99.08\% & 98.64\% & 98.98\% \\
			&          & std  & 0.0798 & 0.0798 & 0.0803 & 0.0796 & & & & \\
			& macro-f1 & mean & 0.6477 & 0.6573 & 0.6530 & 0.6564 & 97.31\% & 98.75\% & 98.11\% & 98.62\% \\
			&          & std  & 0.1065 & 0.1045 & 0.1057 & 0.1045 & & & & \\
			\hline
			\multirow{4}{*}{musk}
			& accuracy & mean & 0.8574 & 0.8635 & 0.8810 & 0.8661 & 97.80\% & 98.49\% & 100.5\% & 98.79\% \\
			&          & std  & 0.0461 & 0.0479 & 0.0434 & 0.0472 & & & & \\
			& macro-f1 & mean & 0.8570 & 0.8627 & 0.8805 & 0.8654 & 97.84\% & 98.49\% & 100.5\% & 98.80\% \\
			&          & std  & 0.0462 & 0.0481 & 0.0435 & 0.0472 & & & & \\
			\hline
			\multirow{4}{*}{drug}
			& accuracy & mean & 0.7670 & 0.7774 & 0.7721 & 0.7765 & 98.43\% & 99.77\% & 99.09\% & 99.65\% \\
			&          & std  & 0.0496 & 0.0523 & 0.0512 & 0.0536 & & & & \\
			& macro-f1 & mean & 0.6612 & 0.6888 & 0.6795 & 0.6908 & 95.31\% & 99.29\% & 97.95\% & 99.58\% \\
			&          & std  & 0.0723 & 0.0728 & 0.0726 & 0.0733 & & & & \\
			\bottomrule
		\end{tabular}
	\end{table}
	
	\begin{table}[H]
		\centering
		\small
		\captionsetup{width=0.9\textwidth, font={footnotesize}}
		\caption{Performance of KNN with random $2D$ projection plane selection strategy across different $k$ values $(n/4, n/2,$ $3n/4, n)$ over 100 runs: mean and mean standard deviation of accuracy and Macro-F1 score, and performance proportion relative to the full projection model.}
		\setlength{\tabcolsep}{5pt}
		\renewcommand{\arraystretch}{1.3}
		\begin{tabular}{ccccccc|cccc}
			\toprule
			Dataset & Metric & Stat & $k_{1}=\dfrac{n}{4}$ & $k_{2}=\dfrac{n}{2}$ & $k_{3}=\dfrac{3n}{4}$ & $k_{4}=n$ & $prop_{1}$ & $prop_{2}$ & $prop_{3}$ & $prop_{4}$ \\
			\midrule
			\multirow{4}{*}{toxicity}
			& accuracy & mean & 0.6145 & 0.6180 & 0.6213 & 0.6231 & 97.06\% & 97.61\% & 98.14\% & 98.42\% \\
			&          & std  & 0.0926 & 0.0942 & 0.0954 & 0.0975 & & & & \\
			& macro-f1 & mean & 0.5155 & 0.5205 & 0.5250 & 0.5269 & 95.89\% & 96.82\% & 97.66\% & 98.01\% \\
			&          & std  & 0.1096 & 0.1135 & 0.1132 & 0.1164 & & & & \\
			\hline
			\multirow{4}{*}{darwin}
			& accuracy & mean & 0.7137 & 0.7077 & 0.7066 & 0.7073 & 102.2\% & 101.3\% & 101.2\% & 101.3\% \\
			&          & std  & 0.0826 & 0.0806 & 0.0804 & 0.0807 & & & & \\
			& macro-f1 & mean & 0.6851 & 0.6773 & 0.6761 & 0.6770 & 102.9\% & 101.8\% & 101.6\% & 101.7\% \\
			&          & std  & 0.1038 & 0.1029 & 0.1027 & 0.1029 & & & & \\
			\hline
			\multirow{4}{*}{musk}
			& accuracy & mean & 0.8519 & 0.8643 & 0.8671 & 0.8688 & 97.17\% & 98.59\% & 98.90\% & 99.10\% \\
			&          & std  & 0.0455 & 0.0460 & 0.0456 & 0.0457 & & & & \\
			& macro-f1 & mean & 0.8510 & 0.8634 & 0.8662 & 0.8679 & 97.16\% & 98.57\% & 98.89\% & 99.09\% \\
			&          & std  & 0.0456 & 0.0461 & 0.0458 & 0.0459 & & & & \\
			\hline
			\multirow{4}{*}{drug}
			& accuracy & mean & 0.7757 & 0.7738 & 0.7779 & 0.7759 & 99.55\% & 99.31\% & 99.83\% & 99.58\% \\
			&          & std  & 0.0481 & 0.0510 & 0.0501 & 0.0522 & & & & \\
			& macro-f1 & mean & 0.6757 & 0.6749 & 0.6824 & 0.6820 & 97.41\% & 97.29\% & 98.37\% & 98.31\% \\
			&          & std  & 0.0703 & 0.0738 & 0.0719 & 0.0741 & & & & \\
			\bottomrule
		\end{tabular}
	\end{table}
	
	\subsection{Figures}
	\subsubsection{$2D$ projection planes importance scores}
	\begin{figure}[H]
		\centering
		\begin{minipage}{0.45\linewidth}
			\centerline{\includegraphics[width=\linewidth]{wine_pairs_plot.png}}
		\end{minipage}
		\vspace{3pt}
		\begin{minipage}{0.45\linewidth}
			\centerline{\includegraphics[width=\linewidth]{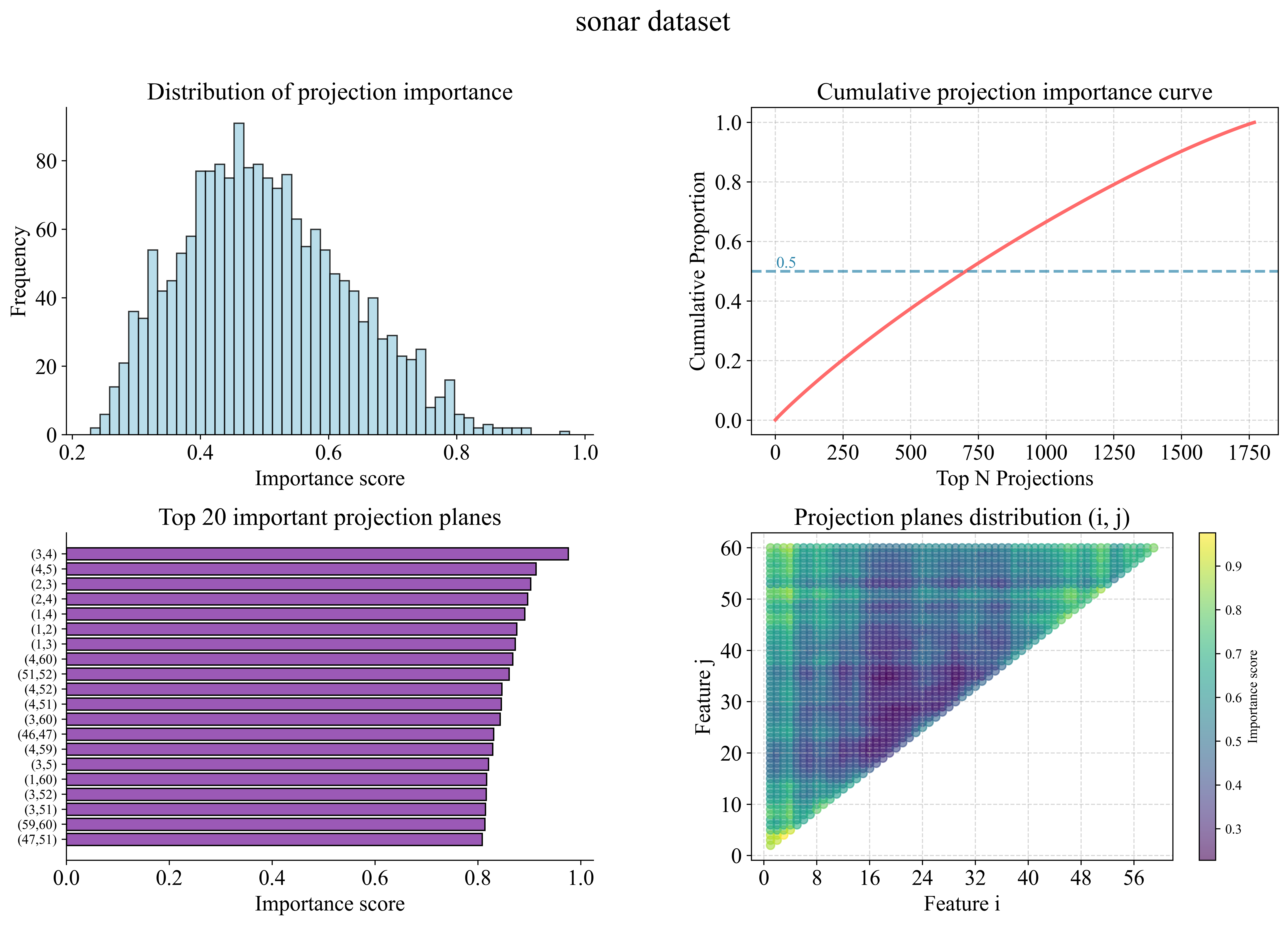}}
		\end{minipage}
		\captionsetup{width=0.9\textwidth, font={footnotesize}}
		\caption{Distribution of importance scores for $2D$ projection planes for the wine and sonar datasets: histogram, cumulative trend, top-20 ranked planes, and their spatial distribution.}
	\end{figure}
	
	\begin{figure}[H]
		\centering
		\vspace{3pt}
		\begin{minipage}{0.45\linewidth}
			\centerline{\includegraphics[width=\linewidth]{seeds_pairs_plot.png}}
		\end{minipage}
		\vspace{3pt}
		\begin{minipage}{0.45\linewidth}
			\centerline{\includegraphics[width=\linewidth]{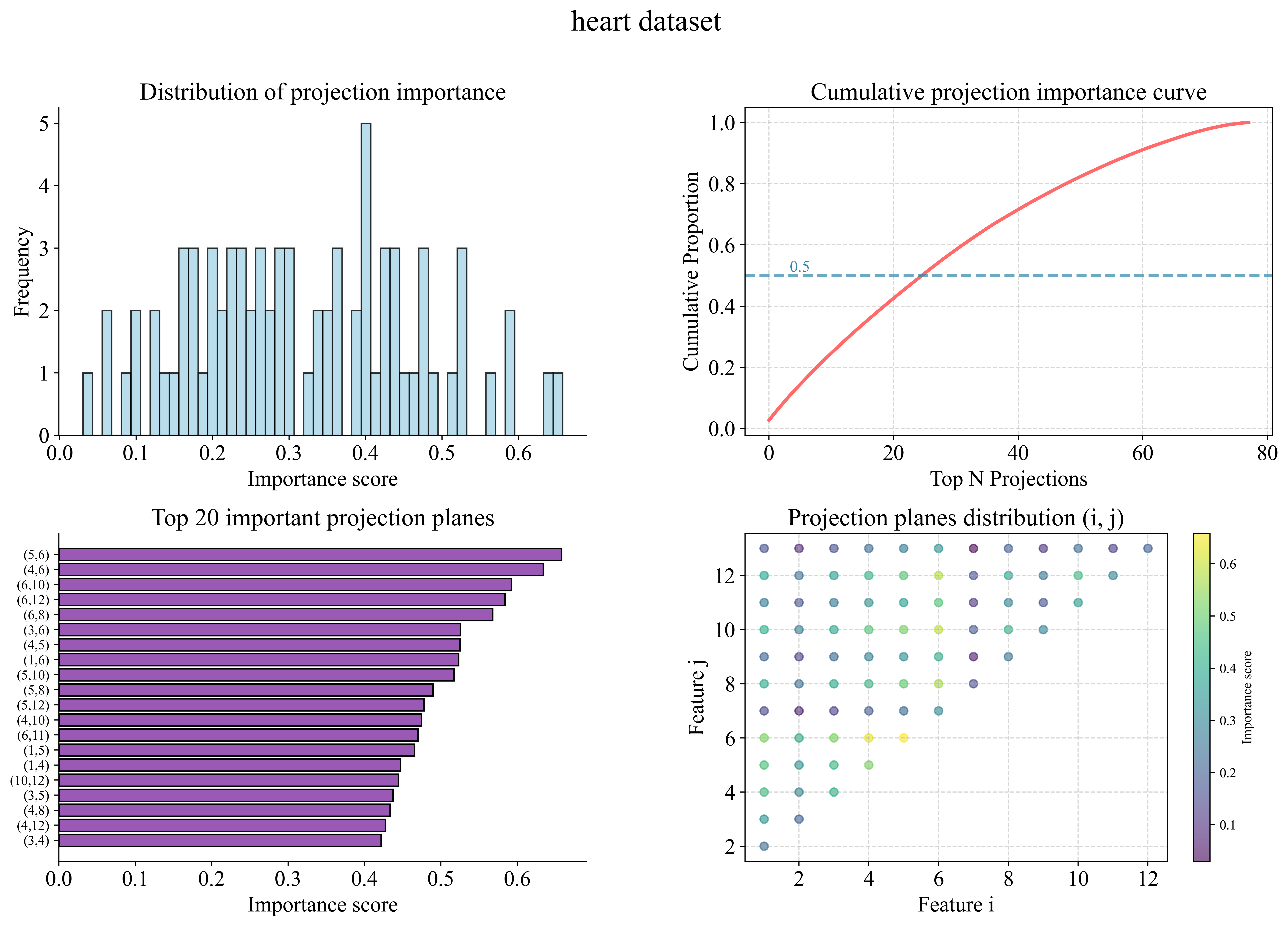}}
		\end{minipage}
		\captionsetup{width=0.9\textwidth, font={footnotesize}}
		\caption{Distribution of importance scores for $2D$ projection planes for the seeds and heart datasets.}
	\end{figure}
	
	\begin{figure}[H]
		\centering
		\vspace{3pt}
		\begin{minipage}{0.45\linewidth}
			\centerline{\includegraphics[width=\linewidth]{ionosphere_pairs_plot.png}}
		\end{minipage}
		\vspace{3pt}
		\begin{minipage}{0.45\linewidth}
			\centerline{\includegraphics[width=\linewidth]{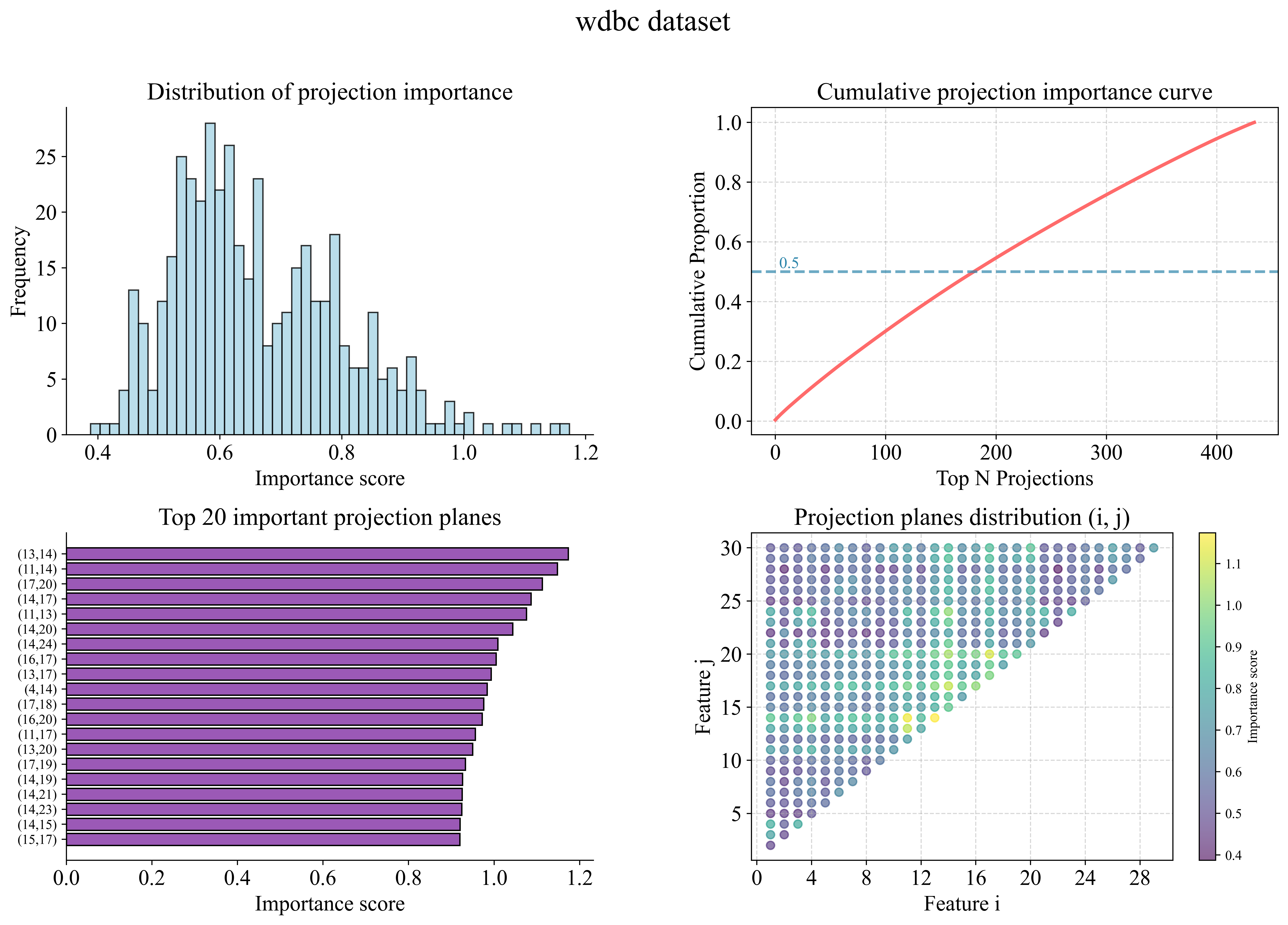}}
		\end{minipage}
		\captionsetup{width=0.9\textwidth, font={footnotesize}}
		\caption{Distribution of importance scores for $2D$ projection planes for the ionosphere and wdbc datasets.}
	\end{figure}
	
	\begin{figure}[H]
		\centering
		\vspace{3pt}
		\begin{minipage}{0.45\linewidth}
			\centerline{\includegraphics[width=\linewidth]{digits_pairs_plot.png}}
		\end{minipage}
		\vspace{3pt}
		\begin{minipage}{0.45\linewidth}
			\centerline{\includegraphics[width=\linewidth]{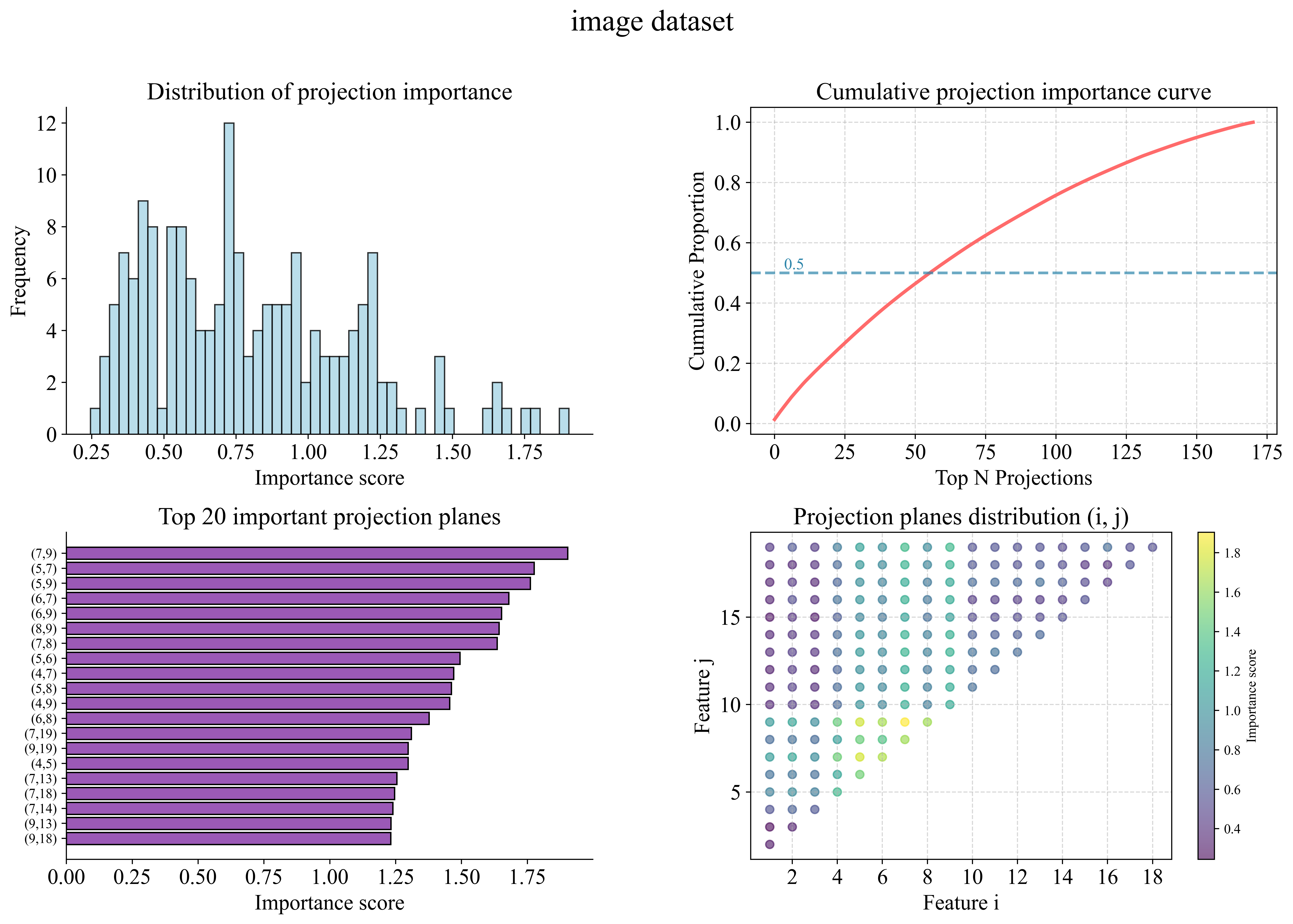}}
		\end{minipage}
		\captionsetup{width=0.9\textwidth, font={footnotesize}}
		\caption{Distribution of importance scores for $2D$ projection planes for the digits and image datasets.}
	\end{figure}
	
	\begin{figure}[H]
		\centering
		\begin{minipage}{0.45\linewidth}
			\centerline{\includegraphics[width=\linewidth]{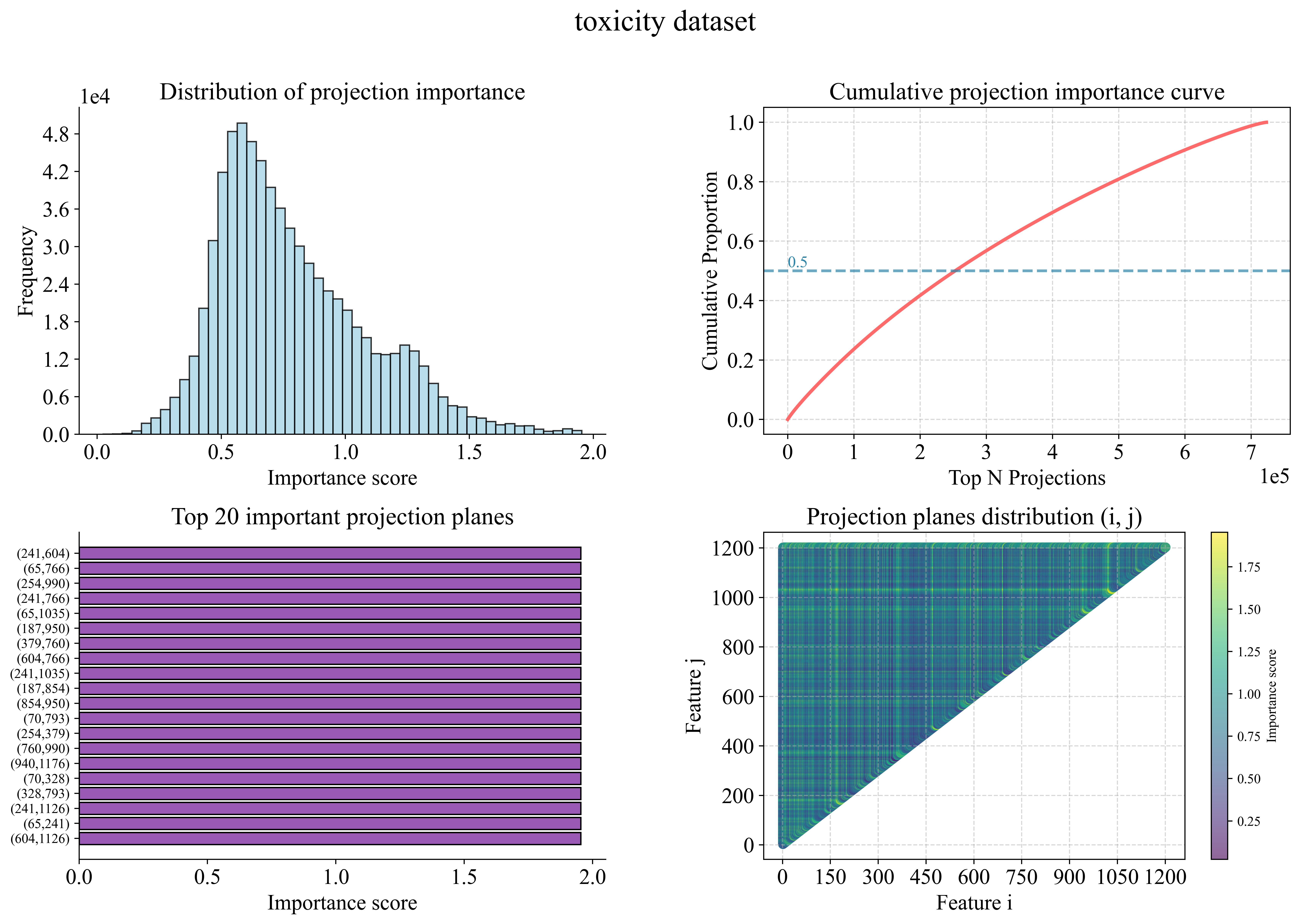}}
		\end{minipage}
		\vspace{3pt}
		\begin{minipage}{0.45\linewidth}
			\centerline{\includegraphics[width=\linewidth]{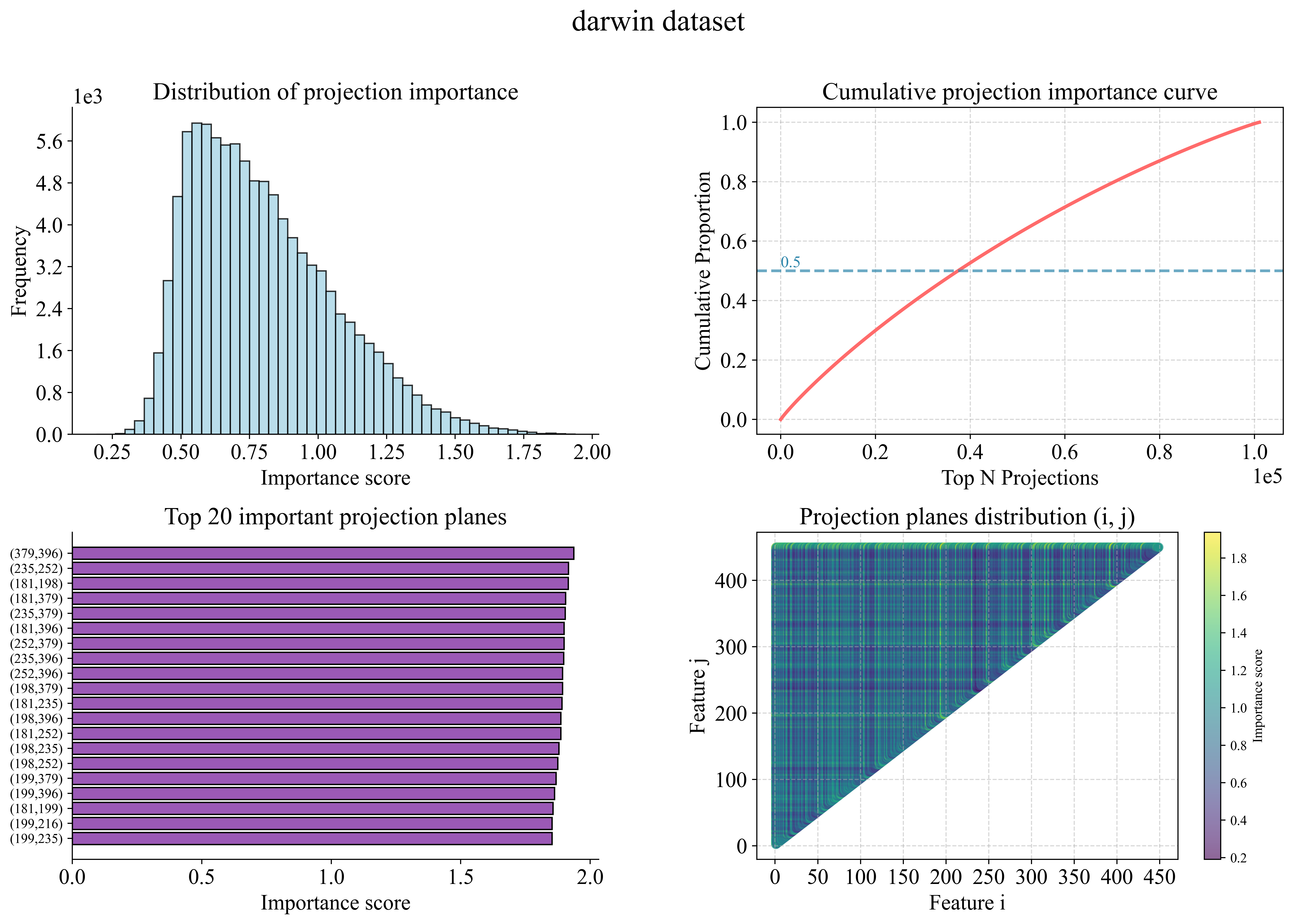}}
		\end{minipage}
		\captionsetup{width=0.9\textwidth, font={footnotesize}}
		\caption{Distribution of importance scores for $2D$ projection planes for the toxicity and darwin datasets: histogram, cumulative trend, top-20 ranked planes, and their spatial distribution.}
	\end{figure}
	
	\begin{figure}[H]
		\centering
		\begin{minipage}{0.45\linewidth}
			\centerline{\includegraphics[width=\linewidth]{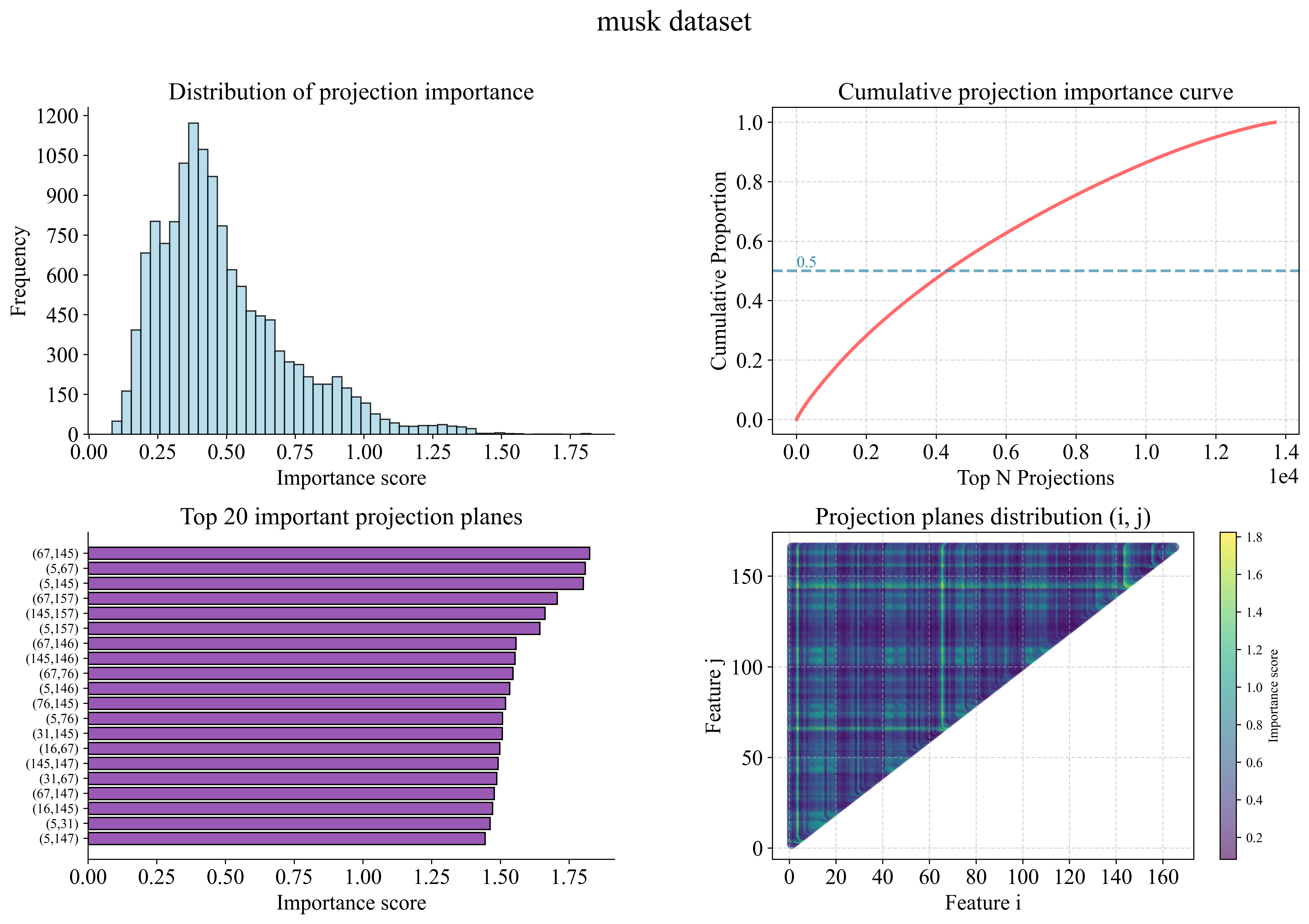}}
		\end{minipage}
		\vspace{3pt}
		\begin{minipage}{0.45\linewidth}
			\centerline{\includegraphics[width=\linewidth]{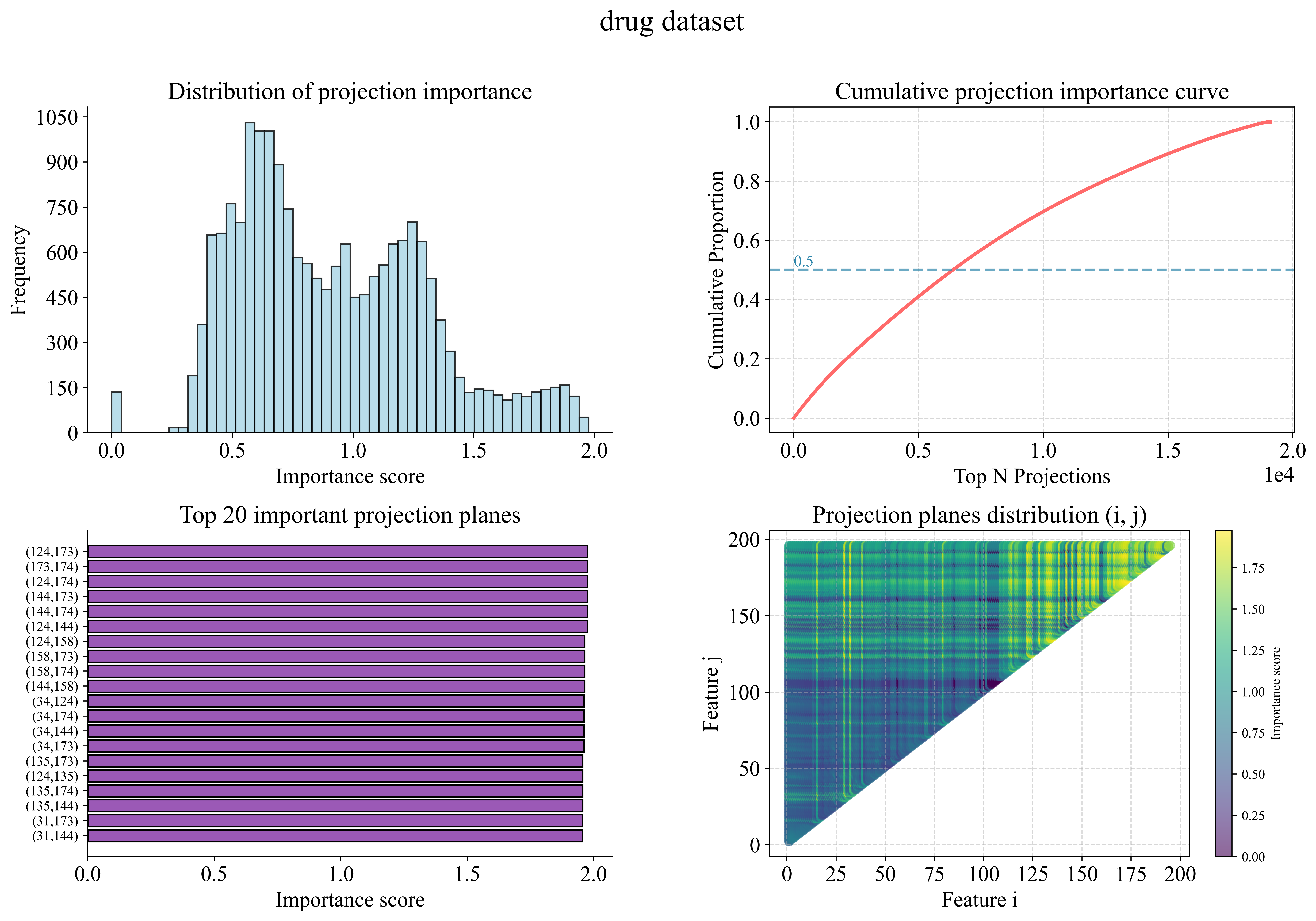}}
		\end{minipage}
		\captionsetup{width=0.9\textwidth, font={footnotesize}}
		\caption{Distribution of importance scores for $2D$ projection planes for the musk and drug datasets: histogram, cumulative trend, top-20 ranked planes, and their spatial distribution.}
	\end{figure}
	
	\subsubsection{Boxplot}
	\begin{figure}[H]
		\centering
		\begin{minipage}{0.45\linewidth}
			\centerline{\includegraphics[width=\linewidth]{wine_boxplot_result_comp.png}}
		\end{minipage}
		\vspace{3pt}
		\begin{minipage}{0.45\linewidth}
			\centerline{\includegraphics[width=\linewidth]{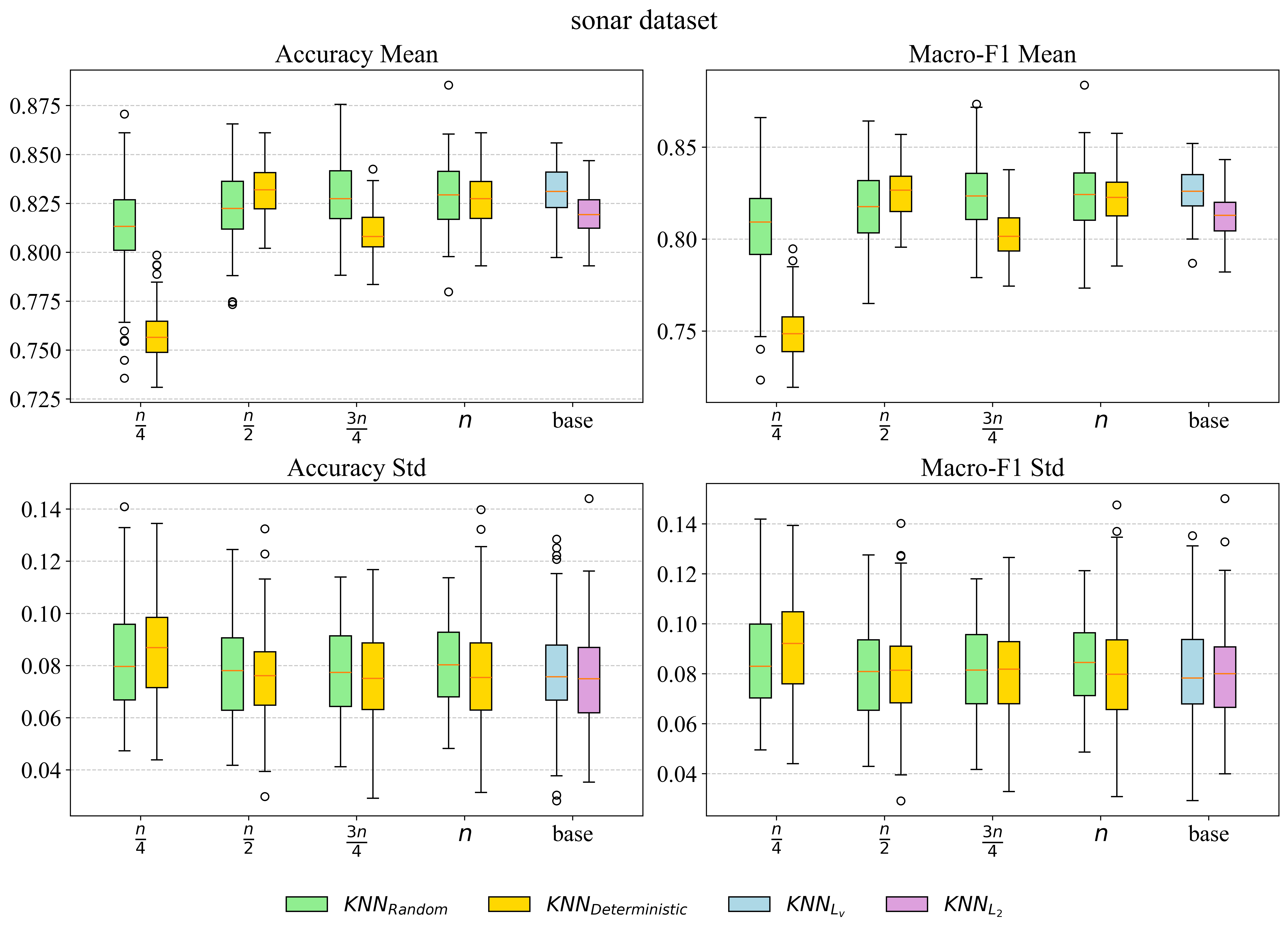}}
		\end{minipage}
		\vspace{3pt}
		\captionsetup{width=0.9\textwidth, font={footnotesize}}
		\caption{Box plots of KNN algorithm performance based on random and deterministic selection strategies for the wine and sonar datasets, with View and Euclidean distances as baselines.}
	\end{figure}
	
	\begin{figure}[H]
		\centering
		\begin{minipage}{0.45\linewidth}
			\centerline{\includegraphics[width=\linewidth]{seeds_boxplot_result_comp.png}}
		\end{minipage}
		\vspace{3pt}
		\begin{minipage}{0.45\linewidth}
			\centerline{\includegraphics[width=\linewidth]{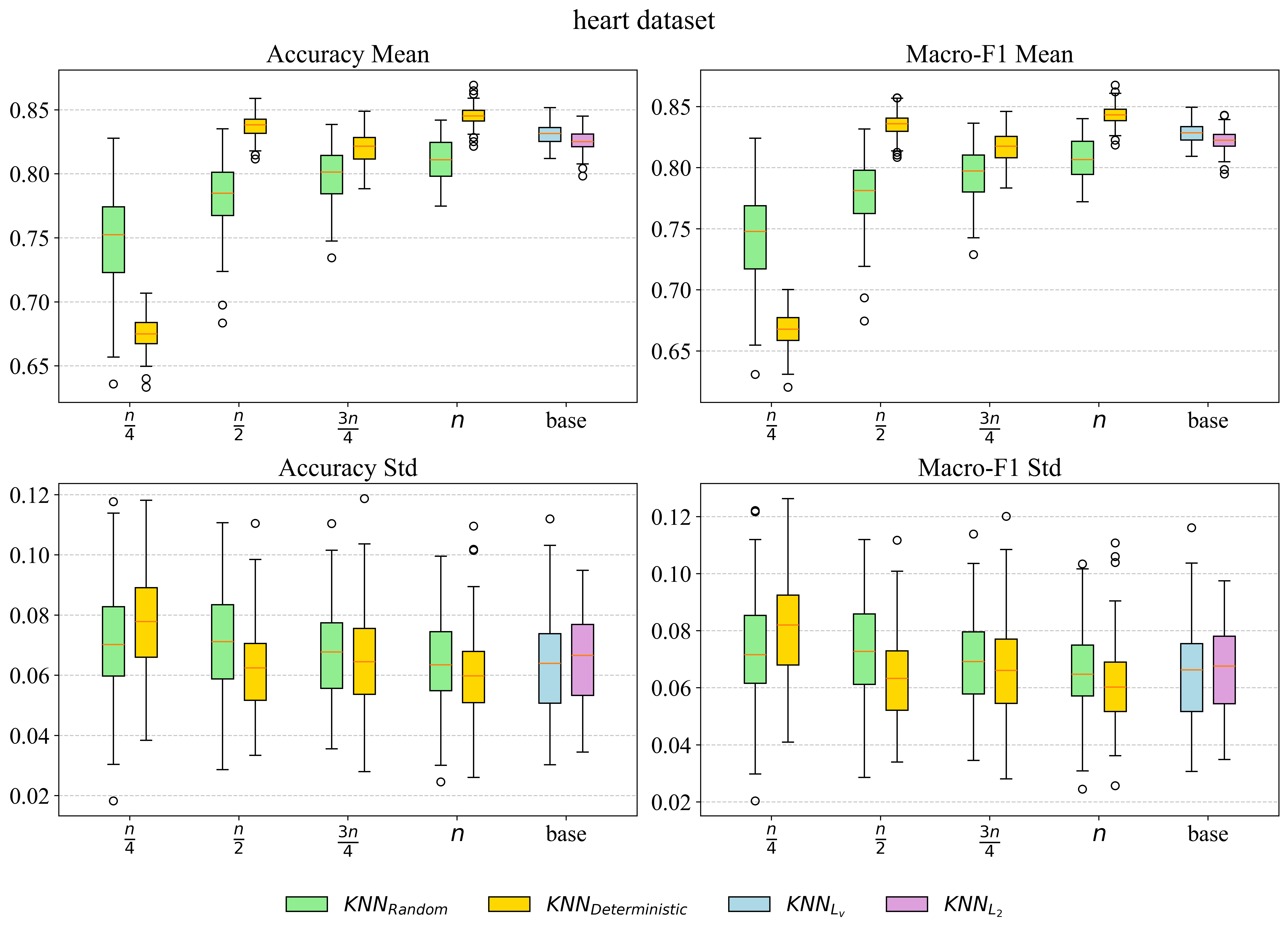}}
		\end{minipage}
		\vspace{3pt}
		\captionsetup{width=0.9\textwidth, font={footnotesize}}
		\caption{Box plots of KNN algorithm performance based on random and deterministic selection strategies for the seeds and heart datasets, with View and Euclidean distances as baselines.}
	\end{figure} 
	
	\begin{figure}[H]
		\centering
		\begin{minipage}{0.45\linewidth}
			\centerline{\includegraphics[width=\linewidth]{ionosphere_boxplot_result_comp.png}}
		\end{minipage}
		\vspace{3pt}
		\begin{minipage}{0.45\linewidth}
			\centerline{\includegraphics[width=\linewidth]{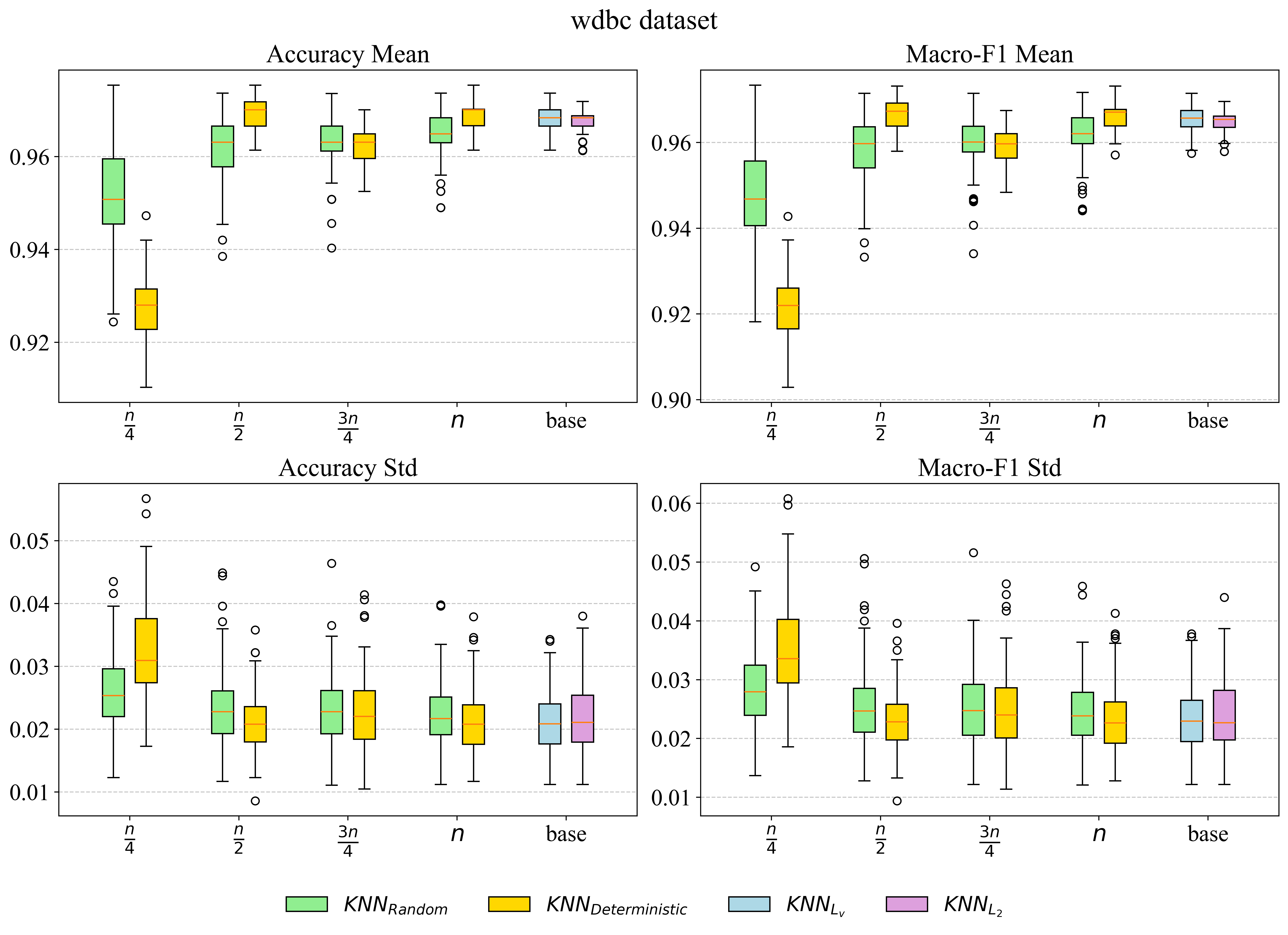}}
		\end{minipage}
		\vspace{3pt}
		\captionsetup{width=0.9\textwidth, font={footnotesize}}
		\caption{Box plots of KNN algorithm performance based on random and deterministic selection strategies for the ionosphere and wdbc datasets, with View and Euclidean distances as baselines.}
	\end{figure}
	
	\begin{figure}[H]
		\centering
		\begin{minipage}{0.45\linewidth}
			\centerline{\includegraphics[width=\linewidth]{digits_boxplot_result_comp.png}}
		\end{minipage}
		\vspace{3pt}
		\begin{minipage}{0.45\linewidth}
			\centerline{\includegraphics[width=\linewidth]{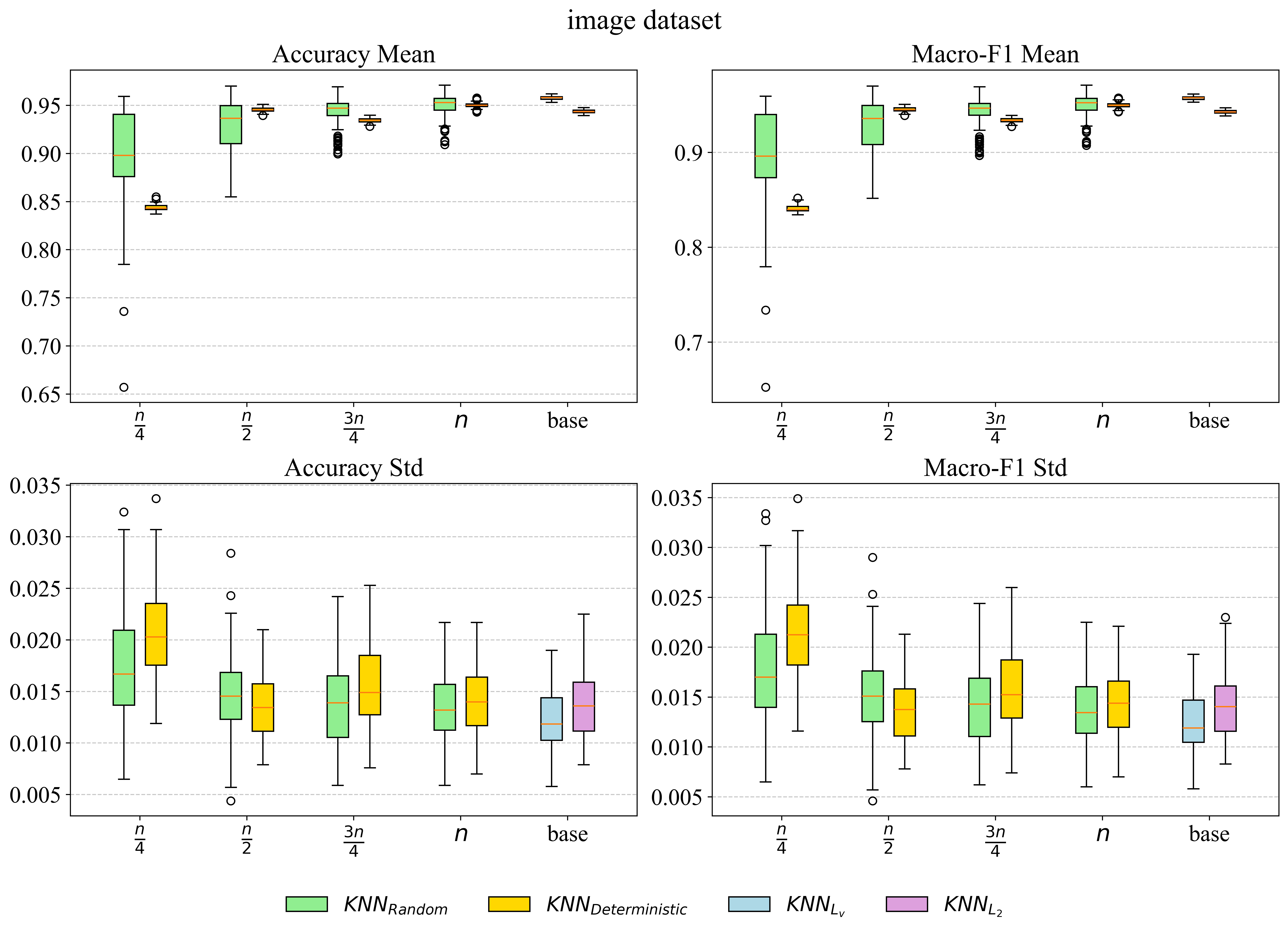}}
		\end{minipage}
		\vspace{3pt}
		\captionsetup{width=0.9\textwidth, font={footnotesize}}
		\caption{Box plots of KNN algorithm performance based on random and deterministic selection strategies for the digits and image datasets, with View and Euclidean distances as baselines.}
	\end{figure} 
	
	\begin{figure}[H]
		\centering
		\begin{minipage}{0.45\linewidth}
			\centerline{\includegraphics[width=\linewidth]{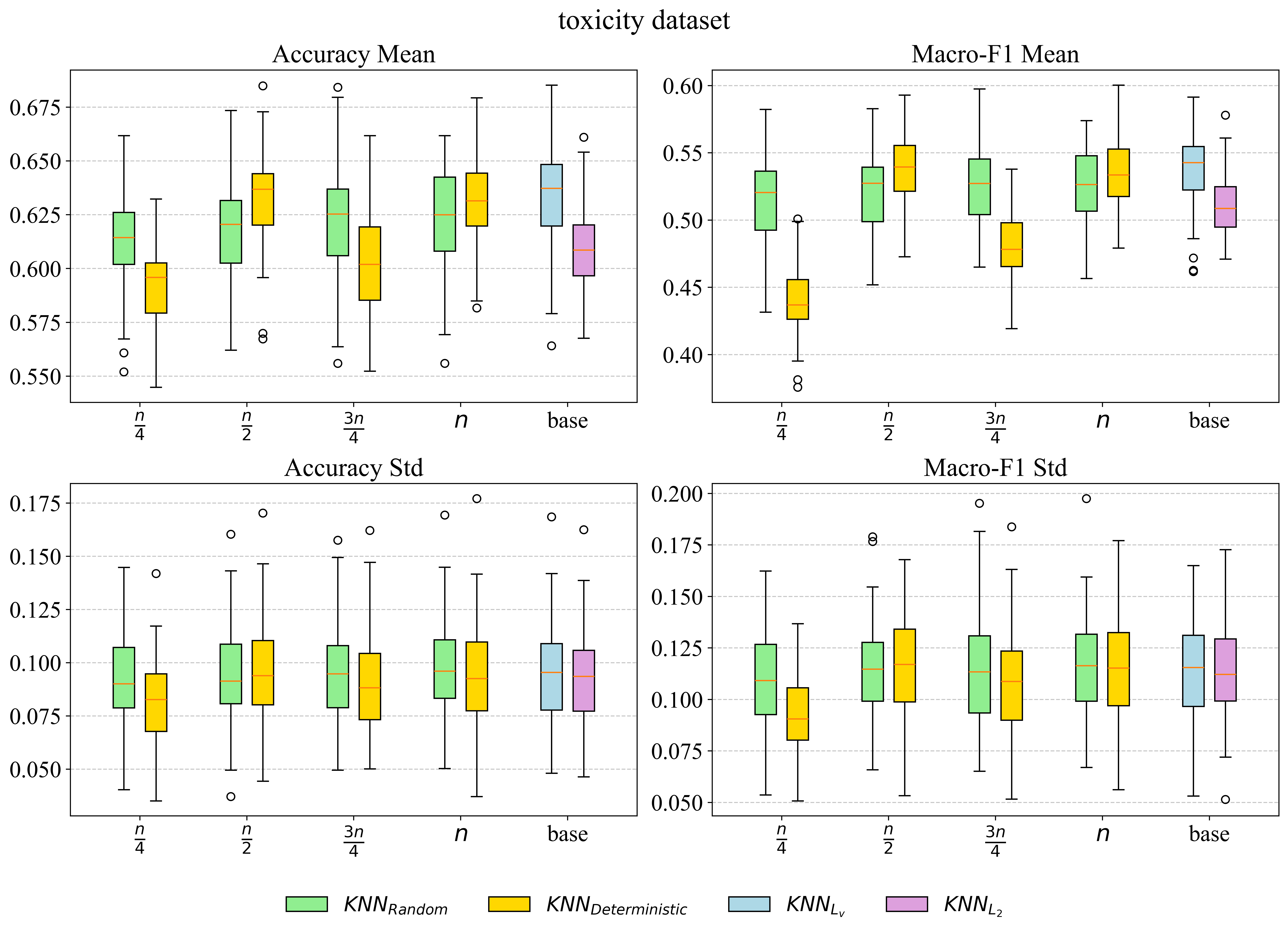}}
		\end{minipage}
		\vspace{3pt}
		\begin{minipage}{0.45\linewidth}
			\centerline{\includegraphics[width=\linewidth]{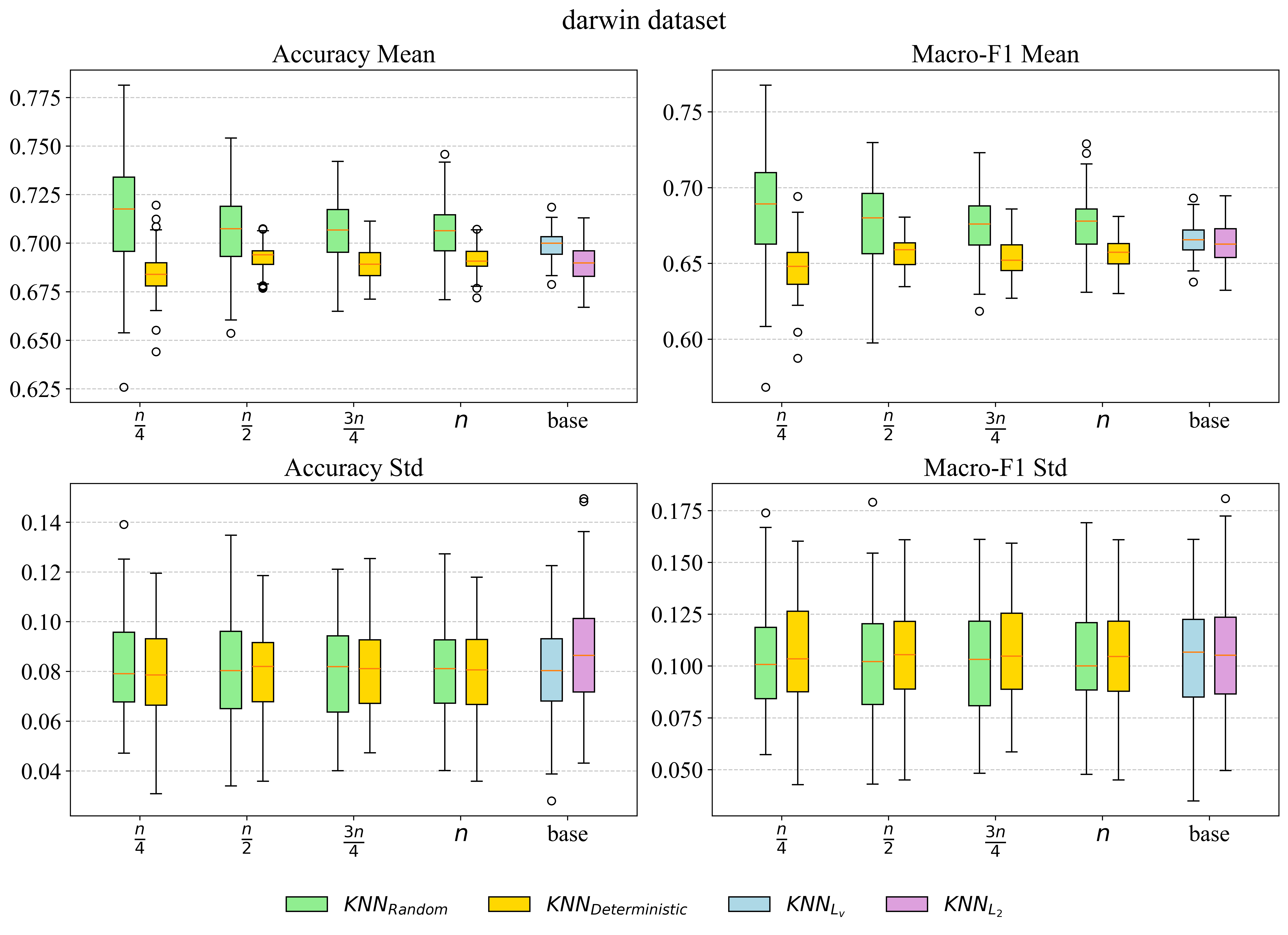}}
		\end{minipage}
		\vspace{3pt}
		\captionsetup{width=0.9\textwidth, font={footnotesize}}
		\caption{Box plots of KNN algorithm performance based on random and deterministic selection strategies for the toxicity and darwin datasets, with View and Euclidean distances as baselines.}
	\end{figure} 
	
		\begin{figure}[H]
		\centering
		\begin{minipage}{0.45\linewidth}
			\centerline{\includegraphics[width=\linewidth]{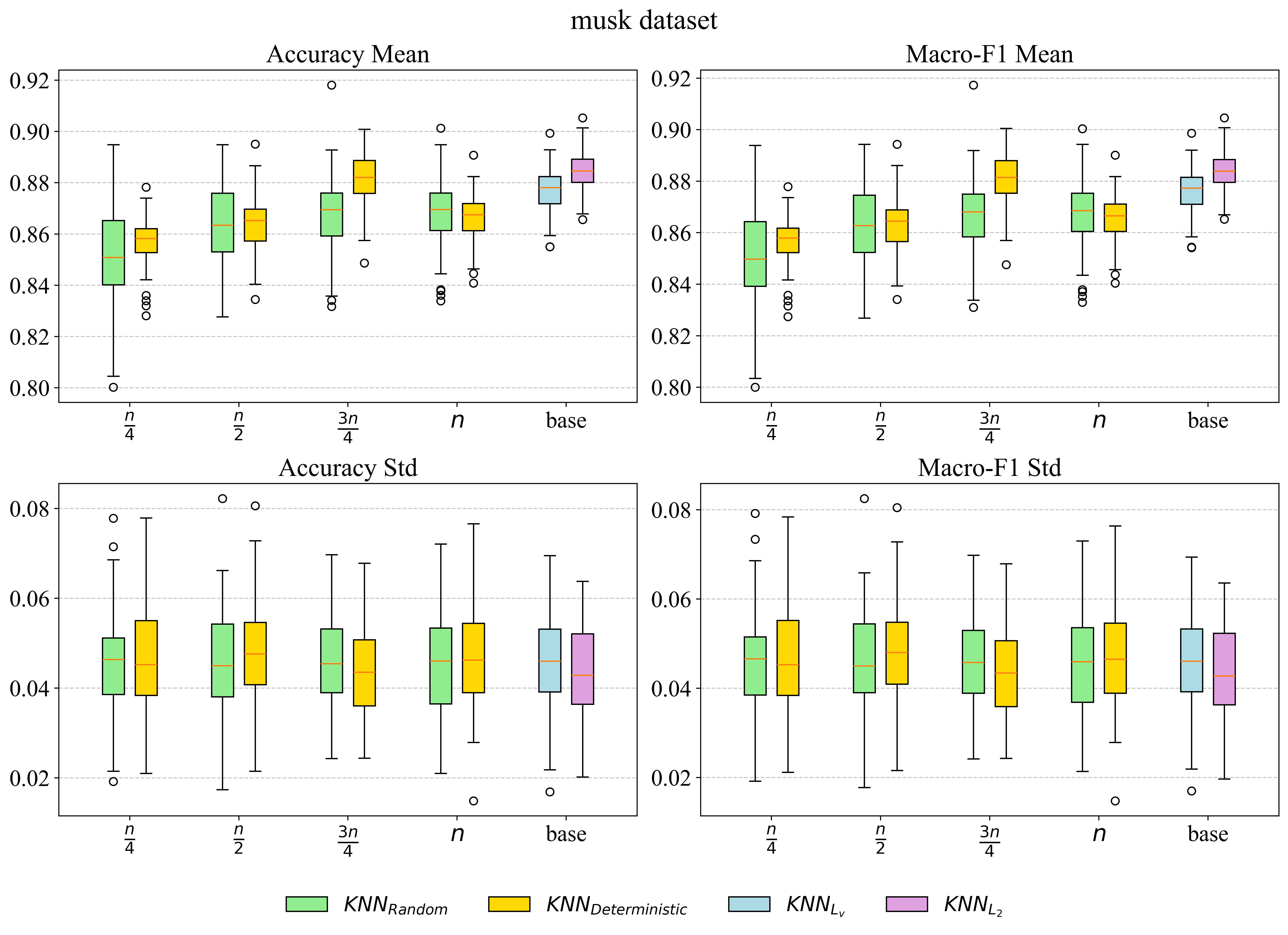}}
		\end{minipage}
		\vspace{3pt}
		\begin{minipage}{0.45\linewidth}
			\centerline{\includegraphics[width=\linewidth]{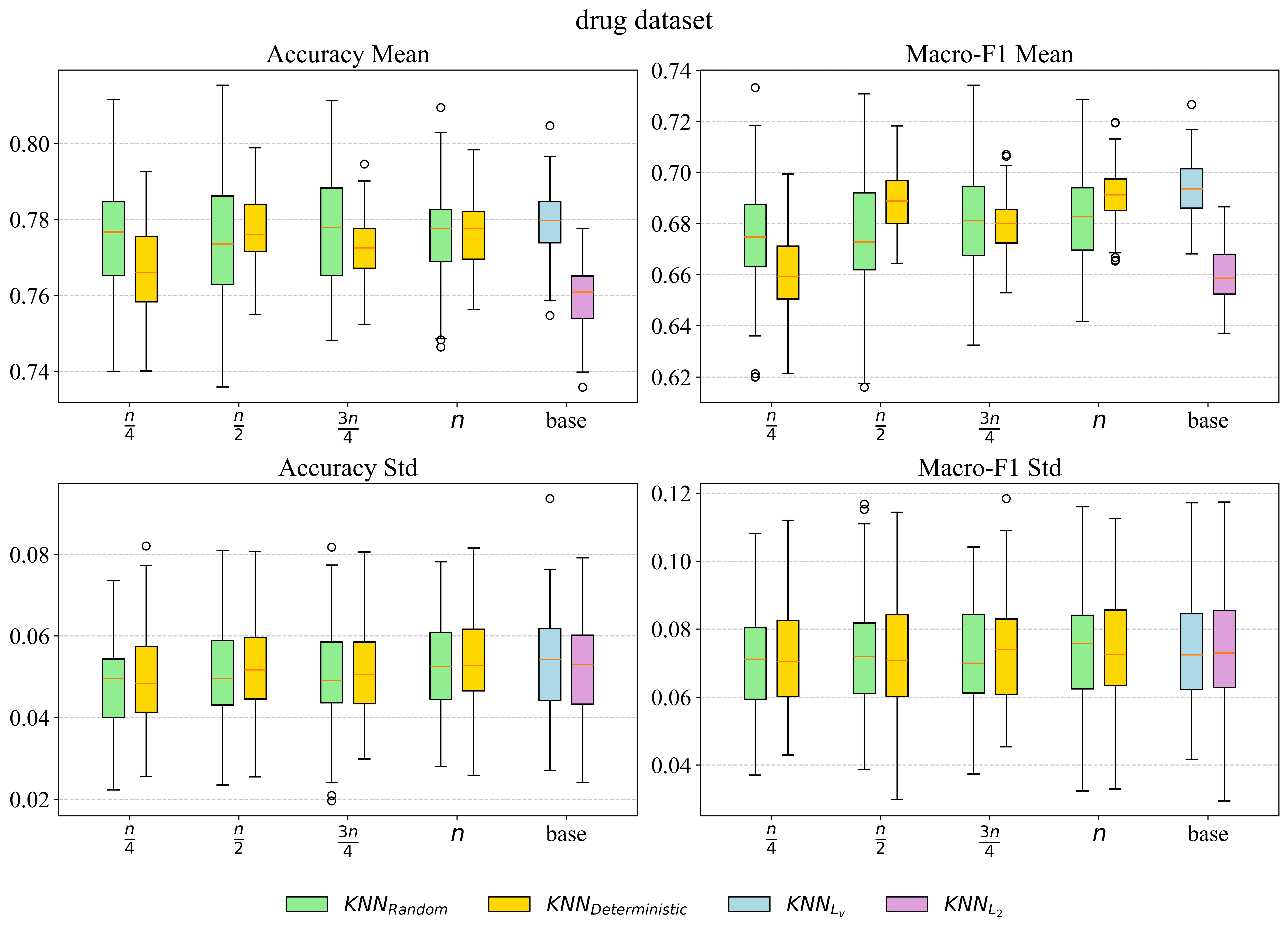}}
		\end{minipage}
		\vspace{3pt}
		\captionsetup{width=0.9\textwidth, font={footnotesize}}
		\caption{Box plots of KNN algorithm performance based on random and deterministic selection strategies for the musk and drug datasets, with View and Euclidean distances as baselines.}
	\end{figure} 
	
\end{document}